\documentclass[pdflatex,sn-mathphys-num]{sn-jnl}

\usepackage[T1]{fontenc}%
\usepackage{graphicx}%
\usepackage{multirow}%
\usepackage{amsmath,amssymb,amsfonts}%
\usepackage{amsthm}%
\usepackage[title]{appendix}%
\usepackage{xcolor}%
\usepackage{textcomp}%
\usepackage{manyfoot}%
\usepackage{booktabs}%
\usepackage{array}%
\usepackage{tabularx}%
\usepackage{siunitx}%
\usepackage[section]{placeins}%
\usepackage{algorithm}%
\usepackage{algorithmicx}%
\usepackage{algpseudocode}%
\usepackage{listings}%
\usepackage{fvextra}%
\makeatletter
\@ifundefined{microtypesetup}{%
  \newcommand{\promptmicrooff}{}%
  \newcommand{\promptmicroon}{}%
}{%
  \newcommand{\promptmicrooff}{\microtypesetup{protrusion=false,expansion=false}}%
  \newcommand{\promptmicroon}{\microtypesetup{protrusion=true,expansion=true}}%
}
\makeatother
\lstdefinestyle{annotationprompt}{
  basicstyle=\ttfamily\footnotesize,
  breaklines=true,
  breakatwhitespace=true,
  columns=fullflexible,
  keepspaces=true,
  showstringspaces=false,
  frame=single,
  rulecolor=\color{black!35},
  xleftmargin=0.4em,
  xrightmargin=0.4em,
  aboveskip=3pt,
  belowskip=3pt
}

\newcommand{\concept}[1]{%
  \ifthenelse{\equal{#1}{ambivalent\_attachment}}{\textsc{Ambivalent Attachment}}{%
    \ifthenelse{\equal{#1}{emotional\_dependency}}{\textsc{Emotional Dependency}}{%
      \ifthenelse{\equal{#1}{idealization}}{\textsc{Idealization}}{%
        \ifthenelse{\equal{#1}{identity\_fragmentation}}{\textsc{Identity Fragmentation}}{%
          \ifthenelse{\equal{#1}{internal\_projection}}{\textsc{Internal Projection}}{%
            \ifthenelse{\equal{#1}{melancholia}}{\textsc{Melancholia}}{%
              \ifthenelse{\equal{#1}{romantic\_obsession}}{\textsc{Romantic Obsession}}{%
                \ifthenelse{\equal{#1}{self\_destructive\_idealization}}{\textsc{Self-Destructive Idealization}}{%
                  \ifthenelse{\equal{#1}{spiritual\_narcissism}}{\textsc{Spiritual Narcissism}}{%
                    \textsc{#1}%
                  }%
                }%
              }%
            }%
          }%
        }%
      }%
    }%
  }%
}
\hypersetup{hypertexnames=false}
\begin{document}

\title[Eigenmood Space]{Eigenmood Space: Uncertainty-Aware Spectral Graph Analysis of Psychological Patterns in Classical Persian Poetry}

\author*[1]{\fnm{Kourosh} \sur{Shahnazari}}\equalcont{They contributed equally in this work.}
\author*[1]{\fnm{Seyed Moein} \sur{Ayyoubzadeh}}\equalcont{They contributed equally in this work.}
\author[2]{\fnm{Mohammadali} \sur{Keshtparvar}}

\affil[1]{\orgname{Sharif University of Technology}, \orgaddress{\city{Tehran}, \country{Iran}}}
\affil[2]{\orgname{Amirkabir University of Technology}, \orgaddress{\city{Tehran}, \country{Iran}}}

\abstract{Classical Persian poetry is a historically sustained archive in which affective life is expressed through metaphor, intertextual convention, and rhetorical indirection. These properties make close reading indispensable while limiting reproducible comparison at scale. We present an uncertainty-aware computational framework for poet-level psychological analysis based on large-scale automatic multi-label annotation. Each verse is associated with a set of psychological concepts, per-label confidence scores, and an abstention flag that signals insufficient evidence. We aggregate confidence-weighted evidence into a Poet $\times$ Concept matrix, interpret each poet as a probability distribution over concepts, and quantify poetic individuality as divergence from a corpus baseline using Jensen--Shannon divergence and Kullback--Leibler divergence. To capture relational structure beyond marginals, we build a confidence-weighted co-occurrence graph over concepts and define an Eigenmood embedding through Laplacian spectral decomposition. On a corpus of 61{,}573 verses across 10 poets, 22.2\% of verses are abstained, underscoring the analytical importance of uncertainty. We further report sensitivity analysis under confidence thresholding, selection-bias diagnostics that treat abstention as a category, and a distant-to-close workflow that retrieves verse-level exemplars along Eigenmood axes. The resulting framework supports scalable, auditable digital-humanities analysis while preserving interpretive caution by propagating uncertainty from verse-level evidence to poet-level inference.}


\maketitle

\section{Introduction}

Persian classical poetry is among the most continuous and socially embedded literary traditions in Eurasian intellectual history. It has served as a vehicle for ethical reflection, mystical reasoning, political allusion, and everyday affect regulation across nearly a millennium of production and reception. The tradition remains active in pedagogy, public memory, and cultural practice, making it a uniquely durable archive for studying how affective meaning is stylized and transmitted over time.

The psychological intensity of Persian poetry has long invited interpretive scholarship, yet its affective expression is rarely literal. Poetic language encodes emotion through metaphor, symbolic displacement, and genre conventions that resist reduction to simple sentiment polarities. A single couplet can simultaneously instantiate longing and renunciation, humility and grandiosity, attachment and withdrawal, or grief and ecstatic transformation. These layered states are aesthetic resources rather than noise, which is why close reading remains indispensable. The methodological difficulty arises when research questions demand corpus-level evidence: prevalence, co-occurrence, comparative baselines, and cross-poet individuality over tens of thousands of verses.

Recent transformer-based language models enable large-scale semantic annotation even for culturally dense corpora, including multi-label concept tagging in which a verse may activate several psychological constructs. However, deploying automatic annotation in the humanities raises two coupled epistemic risks. First, literary ambiguity can cause overconfident labeling unless uncertainty is explicitly modeled. Second, applying contemporary psychological constructs to pre-modern texts can invite anachronism if labels are treated as literal diagnoses. We therefore adopt an instrumental stance: concept labels are operational categories that approximate recurring rhetorical-affective patterns in verse, used for comparative analysis rather than clinical inference. Abstention and confidence are treated as first-class variables that preserve non-commitment and graded evidence.

We introduce a framework that transforms verse-level annotations into poet-level inferential objects with explicit uncertainty propagation. The first object is a Poet $\times$ Concept matrix that aggregates confidence-weighted concept mass over non-abstained verses. From it, each poet is represented as a probability distribution over concepts and compared to a corpus-wide baseline. Individuality is quantified as divergence from baseline, providing an operational and reproducible measure of distributional distinctiveness. Because affective meaning is also relational, we complement marginal distributions with a confidence-weighted co-occurrence graph over concepts and derive an Eigenmood space through spectral decomposition of the graph Laplacian. Eigenmood coordinates quantify the direction of deviation in a coupling-informed latent space and support a distant-to-close workflow via verse retrieval along spectral axes.

Relative to common affective computing pipelines based on fixed lexica or generic embedding spaces, our contribution is to make uncertainty explicit and structurally consequential at every stage: abstention determines what enters aggregation, confidence determines how much it counts, divergence determines magnitude of individuality, and spectral geometry determines direction of individuality based on empirically observed coupling. Relative to classic digital-humanities distant reading, the framework is designed to remain auditable and actionable for close reading by providing both poet-level summaries and verse-level retrieval scores \cite{meymandi2024opportunities,raji2023corpus}.

In summary, the paper contributes an uncertainty-aware formalization for poet-level psychological profiling, including (i) confidence- and abstention-aware aggregation into a Poet $\times$ Concept matrix, (ii) divergence-from-baseline individuality with robustness and selection-bias diagnostics, (iii) a co-occurrence graph and an Eigenmood embedding that complements divergence by encoding direction of deviation, and (iv) an auditable distant-to-close workflow that retrieves verse exemplars aligned with spectral axes while exposing the underlying evidence. The remainder of the paper is organized as follows: Section~2 reviews related work; Section~3 describes the corpus and annotation provenance; Section~4 presents the uncertainty-aware methodology; Section~5 reports quantitative results and robustness analyses; Section~6 presents poet-level psychological profiles and historical interpretation; Section~7 provides interpretive case studies; Section~8 reports validation and human evaluation; Section~9 analyzes abstention-induced selection bias; Section~10 discusses implications and limitations; and Section~11 concludes.

\section{Related Work}

Computational Persian digital humanities has been enabled by corpus-scale resources and by task formulations that respect poetic structure. Large, temporally broad corpora with metadata and rhetorical annotation support reproducible study of stylistic change, periodization, and authorial signatures \cite{raji2023corpus}. Poet identification and stylometry have benefited from hybrid modeling that combines transformer encoders with metrical and stylometric features, and from evaluation regimes that explicitly trade coverage for confidence \cite{shahnazari2025parsi}. Other Persian-focused strands address poem classification and chronological semantics \cite{rahgozar2020automatic,ruma2022deep}, formal and prosodic analysis \cite{shahrestani2025prosody}, and author authenticity questions in Sufi corpora \cite{ishikawa2025statistical}. Recent work further discusses opportunities and risks of applying large language models to Persian DH and argues for workflow transparency and interpretive caution \cite{meymandi2024opportunities}. At a mesoscale, similarity and influence networks have been used to map schools and bridging figures in Persian poetic tradition, offering graph-theoretic structure between close reading and macro literary history \cite{shahnazari2025nazm}. Taken together, this literature establishes the feasibility of large-scale computational analysis while underscoring the methodological need to respect poetic ambiguity, meter, and genre conventions.

Poetry NLP has also advanced through cross-lingual work that develops poetry-specific resources, pretrained models, and task suites. Large annotated corpora for Arabic poetry and specialized pretrained models demonstrate that domain-specific pretraining can improve meter, rhyme, and thematic tasks relative to general-purpose language models \cite{alonazi2025diwan,qarah2024arapoembert}. Neural approaches to classical Arabic prosody and diacritization similarly highlight the importance of formal constraints in poetic text \cite{mutawa2025determining,abandah2020classifying}. Beyond Arabic, computational genre recognition for Uzbek poetry and computational tooling for Hindi poetics provide further examples of how poetic structure can be operationalized without collapsing interpretive nuance \cite{mengliev2024computational,naaz2022design}. Surveys of poetry NLP emphasize that figurative language, rhetorical density, and non-literal reference distinguish poetry from everyday sentiment domains and require tailored evaluation criteria \cite{sisto2024understanding}.

Affective computing and computational psychology provide a second foundation, but the translation from everyday affect to literary affect is non-trivial. Surveys of affective computing emphasize that emotion models, datasets, and evaluation regimes vary widely, and that model outputs require careful calibration when used in downstream inference \cite{wang2022systematic,alsaadawi2024systematic,afzal2023comprehensive}. In poetry-specific settings, reviews of emotion detection in verse highlight the risk of category mismatch and the need for interpretability when affect is encoded through trope rather than literal statement \cite{mehta2025artificial}. For Persian specifically, sentiment analysis has a growing ecosystem but remains challenged by domain shift, metaphor, and resource limitations \cite{rajabi2021survey}. DH systems increasingly explore interactive visualization of affective spaces and emotion distributions in literary works, supporting exploratory workflows that connect distant reading to close reading \cite{wang2025emotionlens,jnicke2017visual,suissa2021text}.

A third thread concerns multi-label prediction, label interaction, and uncertainty estimation. Modern multi-label classifiers model label correlation through attention, graph convolution, feature selection, and contrastive learning objectives \cite{liu2021multi,zeng2023multi,miri2022ensemble,huang2025contrastive,zhang2025hierarchical,wang2021novel}. For humanities corpora, multi-label framing is particularly natural because a verse can simultaneously activate multiple rhetorical-affective configurations; however, evaluation is subtle because prevalence is skewed and labels are not mutually exclusive. Recent work highlights that confidence in multi-label settings is itself a modeling object: poorly calibrated measures can systematically bias downstream aggregation by over-counting uncertain labels \cite{maltoudoglou2022well}. Our framework builds on these insights by treating confidence and abstention as first-class variables in aggregation, divergence computation, and selection-bias analysis.

Finally, debates about computational literary studies emphasize that interpretive value depends on connecting scales rather than replacing close reading with summary statistics. Foundational arguments for abstract models and network reasoning motivate graph-based representations as interpretive instruments \cite{moretti2005graphs,moretti2024network}, while visualization research in DH foregrounds the role of interface and perceptual design in supporting scholarly inference \cite{jnicke2017visual}. Minimal-computing approaches likewise argue for transparent methods that can be inspected and critiqued by domain experts \cite{rizvi2022minimal}. At the same time, methodological critiques warn that computational analyses can become rhetorically persuasive without providing adequate validation or explanatory clarity \cite{da2019computational}, motivating hybrid approaches that integrate prediction with explanation and evaluation \cite{hofman2021integrating}. Our work aligns with these concerns by offering a scalable yet epistemically modest framework: we model poets as probability distributions over an explicit ontology, quantify individuality as divergence from a baseline, derive a coupling-informed spectral geometry (Eigenmood) from concept co-occurrence, and propagate uncertainty through confidence and abstention. The central novelty is not the use of networks per se but the integration of uncertainty-aware aggregation, selection-bias diagnostics, and spectral retrieval as a method for uncertainty-aware psychological profiling in a classical poetic tradition.

\section{Data and Annotation Provenance}

\subsection{Corpus source and composition}

The analyzed corpus comprises 61{,}573 verses across 10 poets (Athir, Eraghi, Hafez, Jahan, Khaghani, Khayyam, Parvin, Saadi, Shahriar, Vahshi). The poetic text was collected from the Ganjoor digital corpus, which is widely used in Persian digital humanities as a large-scale source for classical poetry. We use poet identity either from a record-level field when available or inferred from filename, mirroring established corpus curation practices in computational Persian studies \cite{raji2023corpus,shahnazari2025parsi}. In the present dataset, each poet is stored as a separate JSONL file (e.g., \texttt{POET\_labels.jsonl}), and poet identity is inferred as the filename prefix when a record-level \texttt{poet} field is absent.

\subsection{Text normalization and deduplication}

Preprocessing aims to reduce non-linguistic variation without erasing interpretive signals. We apply Unicode normalization (NFKC), whitespace normalization, and optional diacritic removal for duplicate detection. In this corpus, diacritic marks occur 21{,}993 times, and deduplication after normalization affects 206 verses (0.33\% of the corpus), concentrated almost entirely in the Athir subcorpus (153 duplicates). A recomputation of divergence measures under per-poet deduplication yields negligible changes (Spearman rank correlation of poet individuality remains 1.000), suggesting that duplicates do not drive the reported poet-level patterns.

\subsection{Annotation workflow and uncertainty signals}

Verses were automatically annotated using Gemini 2.5 Flash as a multi-label psychological concept tagger. For each verse $v$, the model returns a label set $L_v$, concept-specific confidence scores $p_{v,c}\in[0,1]$, and a rationale string per predicted label. The prompting protocol constrained the output to a JSON schema with fields \texttt{labels}, \texttt{confidences}, \texttt{rationale}, \texttt{abstain}, and \texttt{notes}, and explicitly enumerated the nine-label ontology to reduce label drift. The model was instructed to assign multiple labels when supported by textual evidence, to express graded confidence, and to abstain when the verse could not be linked to the ontology with sufficient reliability. The workflow includes an abstention mechanism: the model outputs an explicit abstention flag $a_v\in\{0,1\}$ when it cannot identify a reliable psychological signal for the given ontology. Abstention is treated as non-commitment rather than negative evidence.
An abridged schematic of the required output structure is shown in Listing~\ref{lst:schema}.

\begin{lstlisting}[basicstyle=\ttfamily\footnotesize,frame=single,caption={Annotation output schema (abridged).},label={lst:schema}]
{
  "input_verse": "<PERSIAN VERSE>",
  "labels": ["melancholia", "romantic_obsession"],
  "confidences": {"melancholia": 0.72, "romantic_obsession": 0.61},
  "rationale": {"melancholia": "...", "romantic_obsession": "..."},
  "abstain": false,
  "notes": ""
}
\end{lstlisting}

The workflow parses and validates model outputs, logging failures and retrying generation when the response is not valid JSON. In 305 verses (2.23\% of abstained verses) the system exhausted five retries and recorded a failure note; in 92 verses (0.15\% of the corpus) the \texttt{notes} field contains an invalid-JSON trace from the model, providing an auditable provenance trail of extraction errors. Across all non-abstained label instances, only one confidence value and one rationale string are missing, indicating near-complete structured outputs. Abstention is recorded as a first-class field and is retained for coverage diagnostics and selection-bias analysis. In the dataset, 22.2\% of verses are abstained (13{,}678/61{,}573). The most common abstention note is ``no clear psychological signal'' (10{,}192 occurrences; 74.5\% of abstained verses), which captures epistemic non-commitment rather than negative evidence.

\subsection{Label ontology}

The concept inventory contains nine labels: \concept{ambivalent\_attachment}, \concept{emotional\_dependency}, \concept{idealization}, \concept{identity\_fragmentation}, \concept{internal\_projection}, \concept{melancholia}, \concept{romantic\_obsession}, \concept{self\_destructive\_idealization}, and \concept{spiritual\_narcissism}. The ontology is not intended to diagnose authors or historical subjects. It operationalizes recurring affective and interpersonal patterns in poetic discourse, enabling comparative distributional analysis under explicit uncertainty controls.

\subsection{Construct validity of the psychological ontology}

Construct validity is a central concern when importing modern psychological language into a pre-modern poetic archive. Our ontology is therefore framed as a set of operational constructs that approximate recurrent rhetorical-affective configurations rather than latent clinical states. In Persian lyric, affect is not merely expressed but staged through conventionalized figures: the beloved may be human or divine, separation (hijr) and union (vasl) are formalized narrative positions, and selfhood is often distributed across masks such as the lover, the ascetic, and the libertine. The role of the ontology is to make these recurrent configurations comparable at scale while keeping interpretive claims modest.

Each label is grounded in a family of tropes recognizable in Persian poetics. \concept{melancholia} captures lament registers of grief and loss that cluster around motifs of sorrow, separation, and temporal decay; \concept{romantic\_obsession} captures compulsive fixation and repetitive return to the beloved as an object of thought; \concept{emotional\_dependency} captures a posture of need and reliance in which the speaker's stability is rhetorically tied to the beloved's presence or recognition. \concept{ambivalent\_attachment} marks oscillatory discourse that alternates between approach and withdrawal, blessing and reproach, or devotion and refusal, often within the same couplet. \concept{self\_destructive\_idealization} operationalizes devotional intensities in which self-erasure, sacrifice, or harm is framed as the cost of an idealized beloved or transcendent aim. \concept{internal\_projection} corresponds to inward localization of the beloved or the divine as an interior presence, an established move in mystical lyric. \concept{identity\_fragmentation} captures explicit self-contradiction, paradoxical self-description, and unstable speaker position that are prominent in skeptical or epistemically unsettled registers. \concept{spiritual\_narcissism} captures rhetorical self-ascription of spiritual distinction or exceptional proximity to transcendence, which may function as irony, critique, or self-fashioning rather than literal self-aggrandizement. These mappings justify the ontology as a bridge between rhetorical convention and comparative quantification while preserving a non-clinical stance.

The near-absence of \concept{idealization} provides an instructive diagnostic. Across 71{,}638 label assignments, \concept{idealization} occurs only 3 times ($<0.005\%$), yielding baseline share below $10^{-4}$ and disappearing under stricter confidence thresholding ($p_{v,c}\ge 0.7$ retains only 2 instances). This sparsity makes the construct statistically unstable: any small number of annotation errors would dominate its estimated prevalence and could induce spurious spectral structure. We therefore treat \concept{idealization} as a candidate for ontology revision, either by merging it with \concept{self\_destructive\_idealization} or by redefining it to distinguish ``pure'' idealization from sacrificial or obsessional variants. In the present analysis, \concept{idealization} is retained for distributional completeness but excluded from the spectral graph to avoid sparsity-driven eigenmodes.

\section{Methodology}

\subsection{LLM Annotation Protocol and Prompt Transparency}

\noindent\textbf{Model description.}
Verse-level psychological annotations were generated using Gemini~2.5 Flash via the official API. Each verse (bayt) was processed independently under a fixed instruction template, with no cross-verse context and no poet-specific metadata provided to the model. The same prompt was applied uniformly across all poets, with no manual prompt adjustments per author and no few-shot exemplars. To support auditability at scale, the system enforced a strict JSON-only response format and rejected any generation that did not parse as valid JSON.

\noindent\textbf{Decoding parameters.}
Decoding was configured to reduce annotation variance under repeated runs: temperature was set to $0.2$ with $\texttt{top\_p}=1.0$, and the maximum output length was set to accommodate a complete structured JSON object including Persian rationales. Deterministic decoding was preferred to stabilize concept assignment under a multi-label ontology, and a retry mechanism was invoked only when the returned text was malformed JSON. In such cases, the system re-issued the same prompt up to five times and logged failures when schema-conformant JSON could not be obtained.

\noindent\textbf{Annotation constraints.}
The task was formulated as multi-label classification over nine predefined psychological constructs. For each verse $v$, the model output includes a label set $L_v$, an explicit abstention flag, and per-label confidence scores $p_{v,c}\in[0,1]$ represented as numeric values; values are rounded to two decimals at reporting time. For every assigned label, the model provided a brief rationale in Persian, and the response was required to contain only the JSON object (no surrounding commentary). When abstaining, the model returned an empty label set with empty confidence and rationale maps and a short explanatory note.

\noindent\textbf{Justification.}
Abstention is critical in metaphorically dense classical Persian poetry because figurative indirection and intertextual convention can make ontology assignment underdetermined even when affective intensity is high; an explicit non-commitment signal prevents downstream aggregation from converting ambiguity into spurious evidence. Confidence scores are propagated into poet-level aggregation and divergence to represent graded evidential strength rather than uniform label votes, enabling robustness checks under confidence thresholding and selective prediction analyses. Finally, strict structured output enforcement improves reproducibility and auditability by making every verse-level decision machine-parseable, traceable to a fixed schema, and directly comparable across poets under an identical prompt template (Appendix~\ref{app:full_prompt}).

\subsection{Uncertainty-aware Poet \texorpdfstring{$\times$}{x} Concept aggregation}

Let $V_i$ denote the set of verses for poet $i$, and let $a_v$ be the abstention indicator for verse $v$. For each concept $c\in\mathcal{C}$ we define confidence-weighted concept mass
\begin{equation}
X_{i,c}=\sum_{v\in V_i}(1-a_v)\,p_{v,c}\,\mathbf{1}[c\in L_v],
\label{eq:poet_concept_mass}
\end{equation}
yielding a Poet $\times$ Concept matrix $\mathbf{X}\in\mathbb{R}^{N\times|\mathcal{C}|}$. We interpret each poet as a probability distribution over concepts by row normalization with additive smoothing $\varepsilon=10^{-9}$:
\begin{equation}
P_i(c)=\frac{X_{i,c}+\varepsilon}{\sum_{c'\in\mathcal{C}}(X_{i,c'}+\varepsilon)}.
\label{eq:poet_dist}
\end{equation}
The global baseline distribution is
\begin{equation}
P_0(c)=\frac{\sum_i(X_{i,c}+\varepsilon)}{\sum_i\sum_{c'\in\mathcal{C}}(X_{i,c'}+\varepsilon)}.
\label{eq:global_dist}
\end{equation}
This normalization removes corpus-size confounds and supports commensurate poet comparisons.

\subsection{Divergence from baseline}

We quantify individuality as divergence from baseline. KL divergence is
\begin{equation}
D_{KL}(P_i\|P_0)=\sum_{c\in\mathcal{C}} P_i(c)\log\frac{P_i(c)}{P_0(c)},
\label{eq:kl}
\end{equation}
and Jensen--Shannon divergence is
\begin{equation}
D_{JS}(P_i,P_0)=\tfrac{1}{2}D_{KL}(P_i\|M)+\tfrac{1}{2}D_{KL}(P_0\|M), \quad M=\tfrac{1}{2}(P_i+P_0).
\label{eq:js}
\end{equation}
All logarithms use the natural base (units in nats). Divergence is not aesthetic valuation; it is distributional distance from the corpus baseline.

\subsection{Confidence-weighted co-occurrence graph}

To model relational affective structure, we define a concept co-occurrence graph with adjacency matrix $\mathbf{W}=[W_{cd}]$:
\begin{equation}
W_{cd}=\sum_{v}(1-a_v)\,\mathbf{1}[c\in L_v]\,\mathbf{1}[d\in L_v]\cdot\frac{p_{v,c}+p_{v,d}}{2}, \quad c\neq d.
\label{eq:edge_weight}
\end{equation}
Let $D_{cc}=\sum_d W_{cd}$ and define the (unnormalized) Laplacian $\mathbf{L}=\mathbf{D}-\mathbf{W}$.
We use the unnormalized form because $\mathbf{W}$ is constructed as an absolute co-activation mass (expected co-occurrence strength), and unnormalized spectral modes therefore emphasize contrasts supported by high-weight coupling. Normalized Laplacians reweight modes by node degree and can change the resulting geometry; we treat normalization as a sensitivity dimension and report its effects in the Results.

\subsection{Eigen decomposition and Eigenmood embedding}

Let $(\lambda_k,u_k)$ be eigenpairs satisfying
\begin{equation}
\mathbf{L}u_k=\lambda_k u_k, \qquad 0=\lambda_0\le \lambda_1\le\cdots,
\label{eq:eigen}
\end{equation}
with orthonormal eigenvectors. We define Eigenmood coordinates by projecting the baseline-centered distribution onto the first non-trivial modes:
\begin{equation}
z_i^{(k)}=\sum_{c\in\mathcal{C}} (P_i(c)-P_0(c))\,u_k(c).
\label{eq:embedding}
\end{equation}
Eigenmood is distinct from a direct low-rank factorization (e.g., PCA/SVD) of the Poet $\times$ Concept matrix. Factorizations of $\mathbf{X}$ or $\mathbf{P}$ produce axes of variance in marginal concept prevalence, but they do not encode how concepts co-activate within verses. By learning the basis on the concept co-occurrence graph, Laplacian eigenvectors represent relational constraints among concepts and yield modes that support both poet comparison and verse retrieval through graph-coupled directions of deviation.
Rare concepts can behave as weakly connected nodes that dominate early modes through sparsity. For spectral interpretation we therefore restrict to concepts with baseline share at least $10^{-3}$ (0.1\% of global confidence-weighted mass), excluding only \concept{idealization}; distributional and divergence analyses continue to use the full concept set.

\subsection{Robustness and selection-bias diagnostics}

We assess robustness under confidence thresholding by restricting aggregation to label instances with $p_{v,c}\ge\tau$ for $\tau\in\{0.5,0.7\}$ while retaining confidence weighting. We also model selection bias by treating abstention as a category. Specifically, we define an augmented concept set $\mathcal{C}'=\mathcal{C}\cup\{\textsc{Abstain}\}$ where abstained verses contribute unit mass to \textsc{Abstain}. We recompute poet distributions and divergence in $\mathcal{C}'$ and compare rank stability to the baseline ranking.

Because metaphor and polysemy are constitutive in classical Persian verse, reliable interpretation requires explicit validation rather than implicit trust in model outputs. We therefore integrate a controlled two-annotator evaluation workflow (Section~7) that audits label correctness, abstention appropriateness, and the calibration of confidence scores under the nine-label ontology. The two-annotator validation was conducted by domain experts in psychology. All reported agreement, precision, abstention, and calibration statistics are computed directly from the completed validation sheet.

\section{Results}

\subsection{Corpus summary and uncertainty regime}

\begin{figure}[ht]
\centering
\includegraphics[width=0.95\linewidth]{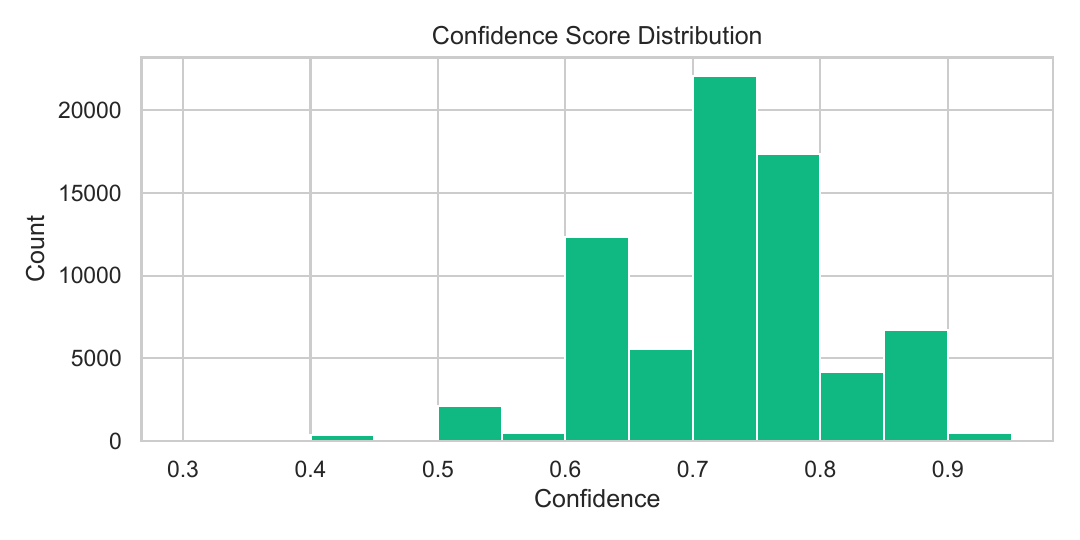}
\caption{Distribution of confidence scores across all label assignments.}
\label{fig:confidence}
\end{figure}

The corpus contains 61{,}573 verses, of which 47{,}895 are non-abstained and 13{,}678 are abstained (abstain rate 0.222). Across non-abstained verses there are 71{,}638 label assignments, corresponding to 1.496 labels per annotated verse. Confidence values range from 0.30 to 0.95 with mean 0.704 (Figure~\ref{fig:confidence}). These statistics indicate that uncertainty is an empirically substantial property rather than a marginal edge case.

Section~7 specifies a two-annotator evaluation on a stratified 500-verse sample designed to quantify label reliability, abstention appropriateness, and confidence calibration under the nine-label ontology. In the \textit{Two-Annotator Validation} reported here, macro agreement is $\bar{\kappa}=0.818$ (Table~\ref{tab:human_kappa}). Against the two-annotator adjudicated reference, macro precision/recall/$F_1$ are $\mathrm{Prec}_{\mathrm{macro}}=0.800$, $\mathrm{Rec}_{\mathrm{macro}}=0.792$, and $F_{1,\mathrm{macro}}=0.794$ (Table~\ref{tab:human_prf1}). Abstention decisions are judged appropriate in 0.856 of verses, and temperature-scaled confidence achieves $\mathrm{ECE}=0.0346$ with a monotonic coverage--risk trade-off (Section~7.5).

\subsection{Global concept distribution}

\begin{figure}[ht]
\centering
\includegraphics[width=0.98\linewidth]{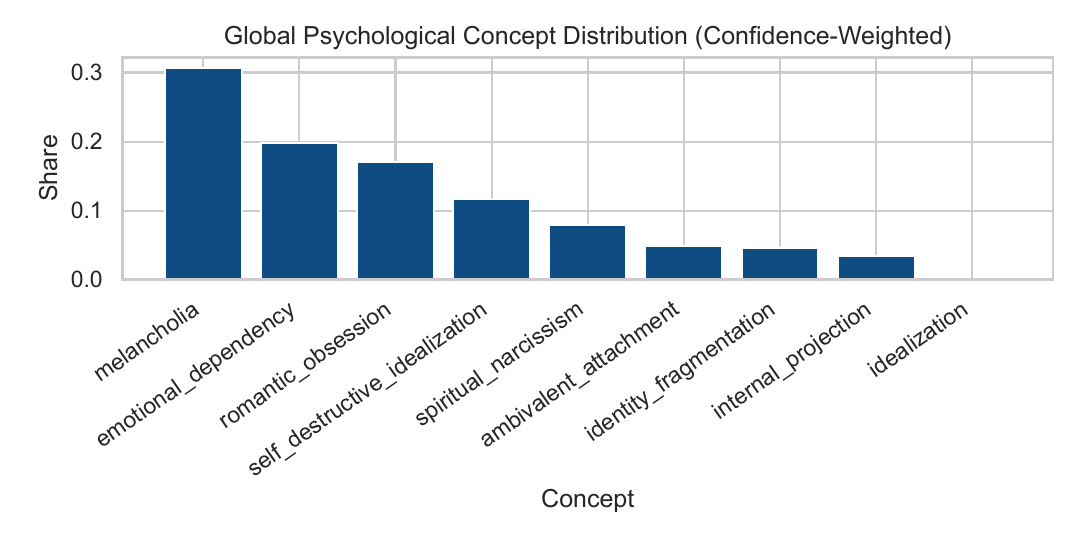}
\caption{Global psychological concept distribution (confidence-weighted).}
\label{fig:label_freq}
\end{figure}

The global baseline distribution $P_0(c)$ is concentrated (Table~\ref{tab:global}). \concept{melancholia} accounts for 0.307 of total confidence-weighted mass, followed by \concept{emotional\_dependency} (0.198) and \concept{romantic\_obsession} (0.170). Figure~\ref{fig:label_freq} visualizes this distribution.

\begin{table}[ht]
\centering
\caption{Global confidence-weighted concept distribution.}
\label{tab:global}
\small
\begin{tabular}{lrr}
\toprule
Concept & Weighted mass & Share ($P_0$) \\
\midrule
\concept{melancholia} & 15462.2 & 0.307 \\
\concept{emotional\_dependency} & 9962.4 & 0.198 \\
\concept{romantic\_obsession} & 8564.9 & 0.170 \\
\concept{self\_destructive\_idealization} & 5900.7 & 0.117 \\
\concept{spiritual\_narcissism} & 3955.4 & 0.078 \\
\concept{ambivalent\_attachment} & 2483.0 & 0.049 \\
\concept{identity\_fragmentation} & 2341.7 & 0.046 \\
\concept{internal\_projection} & 1737.4 & 0.034 \\
\concept{idealization} & 2.1 & 0.000 \\
\bottomrule
\end{tabular}
\end{table}

Notably, \concept{idealization} is effectively absent (baseline share $<10^{-4}$). This pattern suggests an ontology coverage imbalance: idealizing rhetoric in this corpus may be subsumed by more specific constructs such as \concept{self\_destructive\_idealization} and \concept{romantic\_obsession}, or it may appear primarily in verses where the model abstains. We therefore treat \concept{idealization} as a diagnostic signal for ontology refinement and exclude it from the spectral graph to avoid sparsity-driven eigenmodes, while retaining it in distributional and divergence computations for completeness.

Confidence thresholding changes the evidence regime by focusing on high-certainty label instances. At the corpus level, the global distribution is stable under $\tau=0.5$ but shifts under $\tau=0.7$, where \concept{romantic\_obsession} and \concept{emotional\_dependency} become more prevalent while \concept{ambivalent\_attachment} and \concept{internal\_projection} decrease (Table~\ref{tab:global_thresholds}).

\begin{table}[ht]
\centering
\caption{Global concept shares under confidence thresholding (non-abstained verses only).}
\label{tab:global_thresholds}
\small
\begin{tabular}{lrrr}
\toprule
Concept & Base & $\tau=0.5$ & $\tau=0.7$ \\
\midrule
\concept{melancholia} & 0.307 & 0.306 & 0.302 \\
\concept{emotional\_dependency} & 0.198 & 0.198 & 0.206 \\
\concept{romantic\_obsession} & 0.170 & 0.170 & 0.202 \\
\concept{self\_destructive\_idealization} & 0.117 & 0.117 & 0.106 \\
\concept{spiritual\_narcissism} & 0.078 & 0.078 & 0.081 \\
\concept{ambivalent\_attachment} & 0.049 & 0.049 & 0.036 \\
\concept{identity\_fragmentation} & 0.046 & 0.046 & 0.041 \\
\concept{internal\_projection} & 0.034 & 0.035 & 0.026 \\
\bottomrule
\end{tabular}
\end{table}

\subsection{Poet-level profiles and individuality}

\begin{figure}[ht]
\centering
\includegraphics[width=0.98\linewidth]{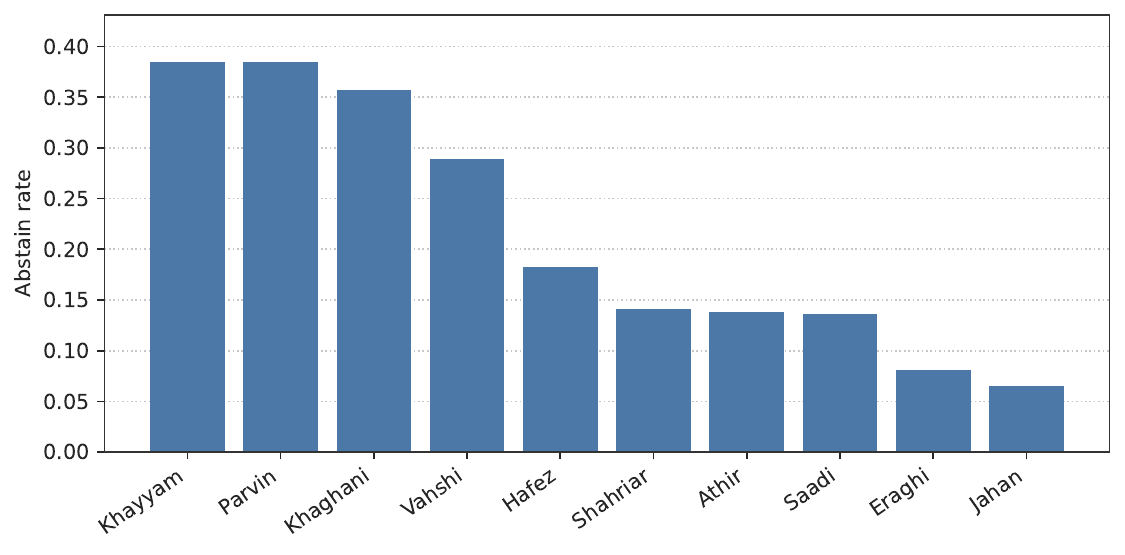}
\caption{Abstention rate by poet.}
\label{fig:abstention}
\end{figure}

\begin{figure}[ht]
\centering
\includegraphics[width=0.98\linewidth]{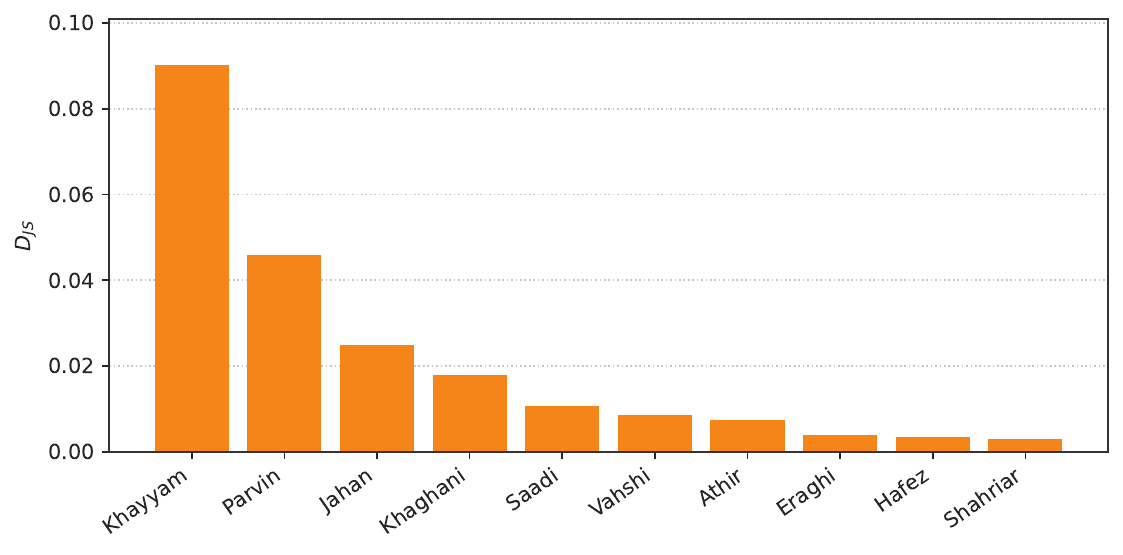}
\caption{Poet individuality measured as Jensen--Shannon divergence from baseline.}
\label{fig:divergence}
\end{figure}

\begin{figure}[ht]
\centering
\includegraphics[width=0.82\linewidth]{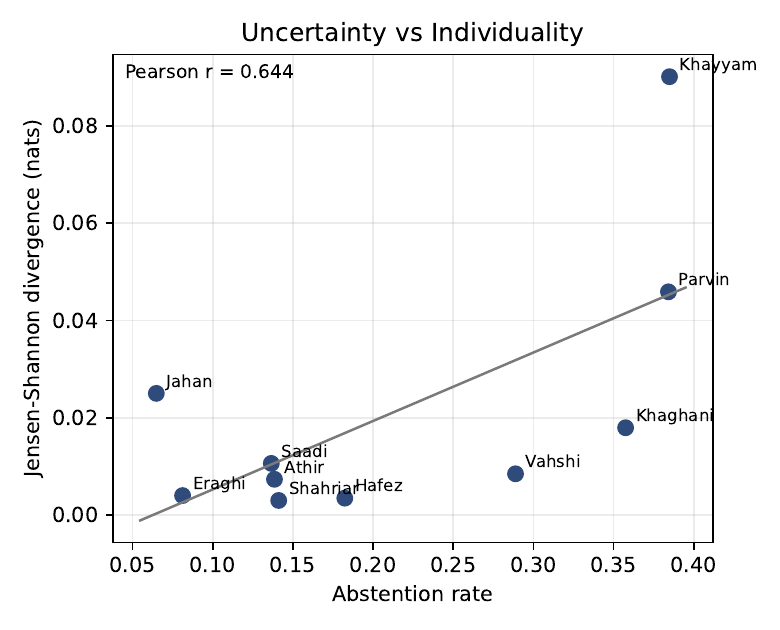}
\caption{Association between abstention rate and poet individuality ($D_{JS}$).}
\label{fig:abstain_jsd}
\end{figure}

Poet-level abstention rates vary widely (Figure~\ref{fig:abstention}), ranging from 0.065 (\textsc{Jahan}) to about 0.385 (\textsc{Khayyam}, \textsc{Parvin}). Individuality, measured by $D_{JS}(P_i,P_0)$, is likewise heterogeneous (Figure~\ref{fig:divergence}). \textsc{Khayyam} is most divergent ($D_{JS}=0.0901$), followed by \textsc{Parvin} ($0.0459$), while \textsc{Hafez} and \textsc{Shahriar} are close to baseline ($0.0035$ and $0.0030$ respectively). Across poets, abstention rate correlates moderately with divergence ($r=0.644$, $t(8)=2.38$, $p=0.044$, 95\% CI $[0.025,\,0.906]$; Figure~\ref{fig:abstain_jsd}); given the small sample of $N=10$ poets, this association should be interpreted cautiously. By contrast, verse count exhibits weak association with divergence ($r=-0.195$), indicating that individuality is not a trivial artifact of subcorpus size. A simpler cosine distance to baseline yields an almost identical individuality ranking (Spearman $\rho=0.976$), suggesting that the main comparative conclusions are not dependent on a particular divergence functional.

\begin{table}[ht]
\centering
\caption{Poet-level summary: verses, abstention, mean confidence, divergence, and dominant concepts (shares from $P_i$).}
\label{tab:poets}
\small
{\renewcommand{\arraystretch}{1.15}%
\setlength{\tabcolsep}{4pt}%
\begin{tabular}{@{}lrrrrp{4.6cm}@{}}
\toprule
Poet & Verses & Abstain & Avg.\ conf.\ & $D_{JS}$ & Top concepts \\
\midrule
Khaghani & 17292 & 0.357 & 0.701 & 0.0179 & \concept{melancholia} (0.316)\newline \concept{spiritual\_narcissism} (0.164) \\
Jahan & 12299 & 0.065 & 0.710 & 0.0250 & \concept{melancholia} (0.319)\newline \concept{emotional\_dependency} (0.305) \\
Saadi & 6864 & 0.137 & 0.708 & 0.0106 & \concept{romantic\_obsession} (0.239)\newline \concept{melancholia} (0.238) \\
Vahshi & 5727 & 0.289 & 0.698 & 0.0085 & \concept{melancholia} (0.327)\newline \concept{self\_destructive\_idealization} (0.152) \\
Parvin & 5573 & 0.384 & 0.695 & 0.0459 & \concept{melancholia} (0.370)\newline \concept{self\_destructive\_idealization} (0.190) \\
Hafez & 5221 & 0.182 & 0.693 & 0.0035 & \concept{melancholia} (0.294)\newline \concept{romantic\_obsession} (0.188) \\
Eraghi & 4668 & 0.081 & 0.711 & 0.0040 & \concept{melancholia} (0.277)\newline \concept{emotional\_dependency} (0.241) \\
Shahriar & 1970 & 0.141 & 0.693 & 0.0030 & \concept{melancholia} (0.348)\newline \concept{romantic\_obsession} (0.175) \\
Athir & 1603 & 0.138 & 0.701 & 0.0073 & \concept{melancholia} (0.290)\newline \concept{romantic\_obsession} (0.222) \\
Khayyam & 356 & 0.385 & 0.700 & 0.0901 & \concept{melancholia} (0.538)\newline \concept{identity\_fragmentation} (0.147) \\
\bottomrule
\end{tabular}
}
\end{table}

\begin{figure}[ht]
\centering
\includegraphics[width=0.98\linewidth]{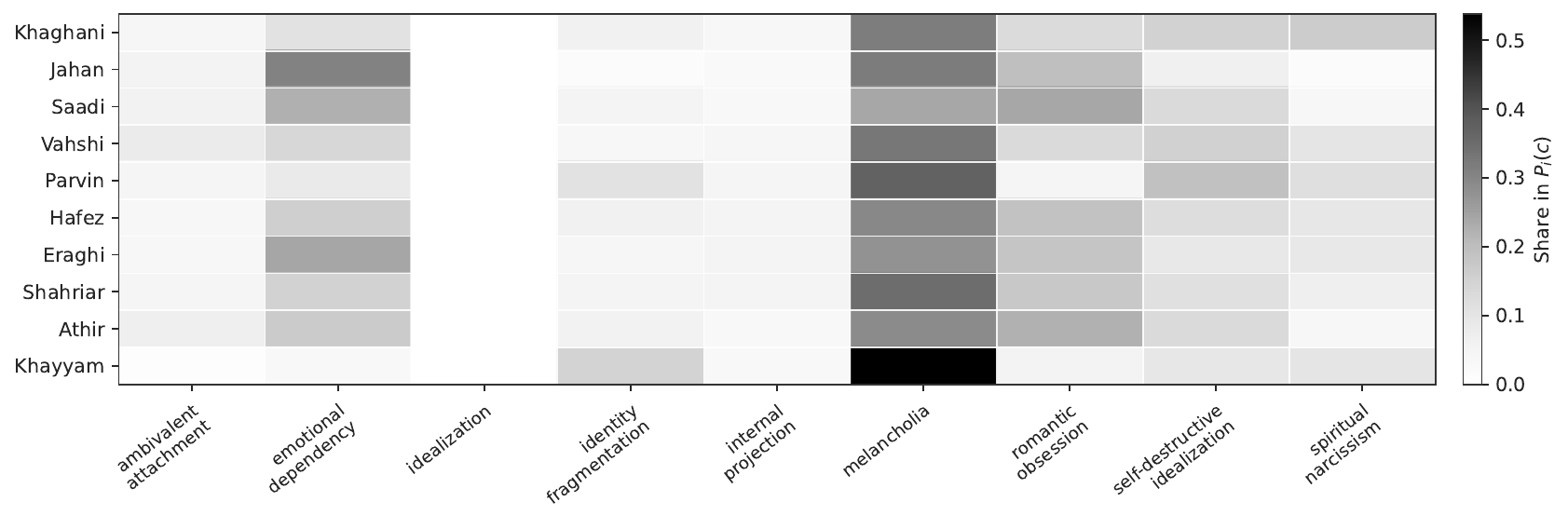}
\caption{Poet $\times$ Concept profiles as confidence-weighted distributions $P_i(c)$ (darker indicates higher share).}
\label{fig:poet_heatmap}
\end{figure}

Figure~\ref{fig:poet_heatmap} provides a compact view of the full Poet $\times$ Concept profiles $P_i(c)$, making visible both shared baseline structure and the localized over- and under-representations that drive divergence.

To attribute individuality to specific concepts, we compute concept lifts $\Delta_i(c)=P_i(c)-P_0(c)$. Table~\ref{tab:lifts} reports the largest positive and negative lifts for the three most distinctive poets. Deviations are concentrated in a small subset of concepts: \textsc{Khayyam} is characterized by elevated \concept{melancholia} and \concept{identity\_fragmentation} and reduced \concept{emotional\_dependency} and \concept{romantic\_obsession}, whereas \textsc{Jahan} shows the opposite pattern with elevated \concept{emotional\_dependency} and reduced \concept{spiritual\_narcissism}.

\begin{table}[ht]
\centering
\caption{Concept lifts for the three most distinctive poets, defined as $\Delta_i(c)=P_i(c)-P_0(c)$ (positive indicates over-representation relative to baseline).}
\label{tab:lifts}
\small
{\renewcommand{\arraystretch}{1.15}%
\setlength{\tabcolsep}{3pt}%
\begin{tabular}{@{}l p{5.15cm} p{5.15cm}@{}}
\toprule
Poet & Largest positive lifts & Largest negative lifts \\
\midrule
Khayyam &
\begin{tabular}[t]{@{}p{3.8cm} S[table-format=1.3]@{}}
\concept{melancholia} & 0.232 \\
\concept{identity\_fragmentation} & 0.101 \\
\concept{spiritual\_narcissism} & 0.022 \\
\end{tabular}
&
\begin{tabular}[t]{@{}p{3.8cm} S[table-format=-1.3]@{}}
\concept{emotional\_dependency} & -0.169 \\
\concept{romantic\_obsession} & -0.116 \\
\concept{ambivalent\_attachment} & -0.040 \\
\end{tabular}
\\
Parvin &
\begin{tabular}[t]{@{}p{3.8cm} S[table-format=1.3]@{}}
\concept{self\_destructive\_idealization} & 0.073 \\
\concept{melancholia} & 0.063 \\
\concept{identity\_fragmentation} & 0.062 \\
\end{tabular}
&
\begin{tabular}[t]{@{}p{3.8cm} S[table-format=-1.3]@{}}
\concept{romantic\_obsession} & -0.126 \\
\concept{emotional\_dependency} & -0.112 \\
\concept{ambivalent\_attachment} & -0.006 \\
\end{tabular}
\\
Jahan &
\begin{tabular}[t]{@{}p{3.8cm} S[table-format=1.3]@{}}
\concept{emotional\_dependency} & 0.107 \\
\concept{romantic\_obsession} & 0.027 \\
\concept{melancholia} & 0.013 \\
\end{tabular}
&
\begin{tabular}[t]{@{}p{3.8cm} S[table-format=-1.3]@{}}
\concept{spiritual\_narcissism} & -0.063 \\
\concept{self\_destructive\_idealization} & -0.050 \\
\concept{identity\_fragmentation} & -0.029 \\
\end{tabular}
\\
\bottomrule
\end{tabular}
}
\end{table}

\subsection{Robustness under confidence thresholding}

Thresholding analyses show that individuality rankings are stable when low-confidence label instances are removed. Spearman correlation between the baseline divergence ranking and the $\tau=0.5$ ranking is $\rho=1.000$, and for $\tau=0.7$ it remains $\rho=0.964$. Table~\ref{tab:thresholds} reports divergence values for the most distinctive poets under each policy, showing that the strongest individuality signals persist under stricter evidence requirements.

\begin{table}[ht]
\centering
\caption{Jensen--Shannon divergence under confidence thresholding. Baseline uses all non-abstained labels; thresholded variants include only labels with $p_{v,c}\ge\tau$.}
\label{tab:thresholds}
\small
\begin{tabular}{lrrr}
\toprule
Poet & Baseline & $\tau=0.5$ & $\tau=0.7$ \\
\midrule
Khayyam & 0.0901 & 0.0900 & 0.1011 \\
Parvin & 0.0459 & 0.0460 & 0.0564 \\
Jahan & 0.0250 & 0.0251 & 0.0282 \\
Khaghani & 0.0179 & 0.0180 & 0.0227 \\
Saadi & 0.0106 & 0.0108 & 0.0160 \\
Vahshi & 0.0085 & 0.0085 & 0.0111 \\
\bottomrule
\end{tabular}
\end{table}

\subsection{Ablation: confidence weighting vs.\ uniform weighting}

Confidence scores are used both in aggregation (Eq.~\ref{eq:poet_concept_mass}) and in concept coupling (Eq.~\ref{eq:edge_weight}). To test whether confidence weighting materially changes conclusions, we recompute poet distributions, divergences, and the spectral basis under a uniform policy that assigns unit mass to every predicted label instance. The individuality ranking is unchanged: Spearman correlation between confidence-weighted and uniform $D_{JS}$ ranks is $\rho=1.000$, with small absolute shifts in divergence values (e.g., \textsc{Khayyam}: 0.0901 $\rightarrow$ 0.0871; \textsc{Parvin}: 0.0459 $\rightarrow$ 0.0431). The co-occurrence geometry is likewise stable: absolute correlations between corresponding Laplacian eigenvectors are 0.99999/0.99990/0.99963 for EM1--EM3, and correlations between poet coordinates exceed 0.9996 for each axis. These results indicate that our main comparisons do not hinge on a particular weighting choice.

\begin{figure}[ht]
\centering
\includegraphics[width=0.92\linewidth]{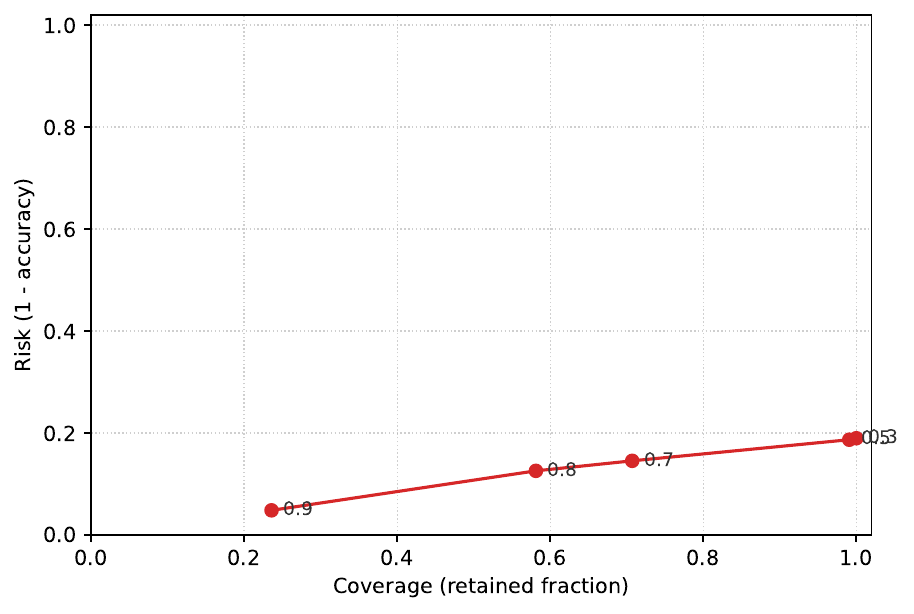}
\caption{Coverage--risk curve on the validation sample as the temperature-scaled confidence threshold $\tau$ increases.}
\label{fig:validation_coverage_risk}
\end{figure}

Confidence weighting is nevertheless beneficial for the intended epistemic role of confidence as graded evidence. Under confidence weighting, extreme verse retrieval along Eigenmood axes concentrates on higher-confidence instances: among the top 500 verses by $|s_v^{(k)}|$ in Eq.~\ref{eq:verse_score}, the mean confidence of contributing labels is higher than under uniform weighting (EM1: 0.793 vs 0.723; EM2: 0.759 vs 0.732; EM3: 0.776 vs 0.720). Consistent with this interpretation of confidence, the Two-Annotator Validation exhibits a monotonic coverage--risk curve: as the temperature-scaled threshold increases, risk decreases (Figure~\ref{fig:validation_coverage_risk}), supporting thresholded retrieval and uncertainty-aware aggregation.

\subsection{Bootstrap uncertainty of individuality and Eigenmood coordinates}

Point estimates of divergence and embedding coordinates can be sensitive to finite subcorpus size, especially for poets with few non-abstained verses. We estimate within-poet sampling variability by nonparametric bootstrap resampling of annotated verses with replacement (200 replicates per poet), recomputing $P_i$ and derived quantities each time while holding the global baseline $P_0$ and the Eigenmood basis fixed. Table~\ref{tab:bootstrap} reports mean and 95\% percentile intervals for $D_{JS}$ and EM2. As expected, \textsc{Khayyam} exhibits the widest interval because it has only 219 non-abstained verses, whereas large subcorpora yield tight uncertainty bands.

\begin{table}[ht]
\centering
\caption{Bootstrap uncertainty intervals (200 replicates) for individuality and EM2. Intervals are 2.5--97.5\% percentiles.}
\label{tab:bootstrap}
\small
{\renewcommand{\arraystretch}{1.15}%
\setlength{\tabcolsep}{4pt}%
\begin{tabular}{@{}l r p{4.1cm} p{3.8cm}@{}}
\toprule
Poet & \shortstack{Non-abstained\\verses} & \shortstack{$D_{JS}$ mean\\[2.5\%, 97.5\%]} & \shortstack{EM2 mean\\[2.5\%, 97.5\%]} \\
\midrule
Khayyam & 219 & 0.094 [0.072, 0.122] & -0.058 [-0.085, -0.028] \\
Parvin & 3432 & 0.046 [0.041, 0.051] & -0.027 [-0.034, -0.019] \\
Jahan & 11502 & 0.025 [0.024, 0.027] & 0.007 [0.005, 0.009] \\
Khaghani & 11111 & 0.018 [0.016, 0.020] & -0.011 [-0.014, -0.008] \\
Saadi & 5927 & 0.011 [0.010, 0.012] & -0.003 [-0.006, 0.000] \\
Vahshi & 4073 & 0.009 [0.007, 0.011] & 0.007 [0.002, 0.012] \\
\bottomrule
\end{tabular}
}
\end{table}

\subsection{Co-occurrence structure and Eigenmood embedding}

\begin{figure}[ht]
\centering
\includegraphics[width=0.92\linewidth]{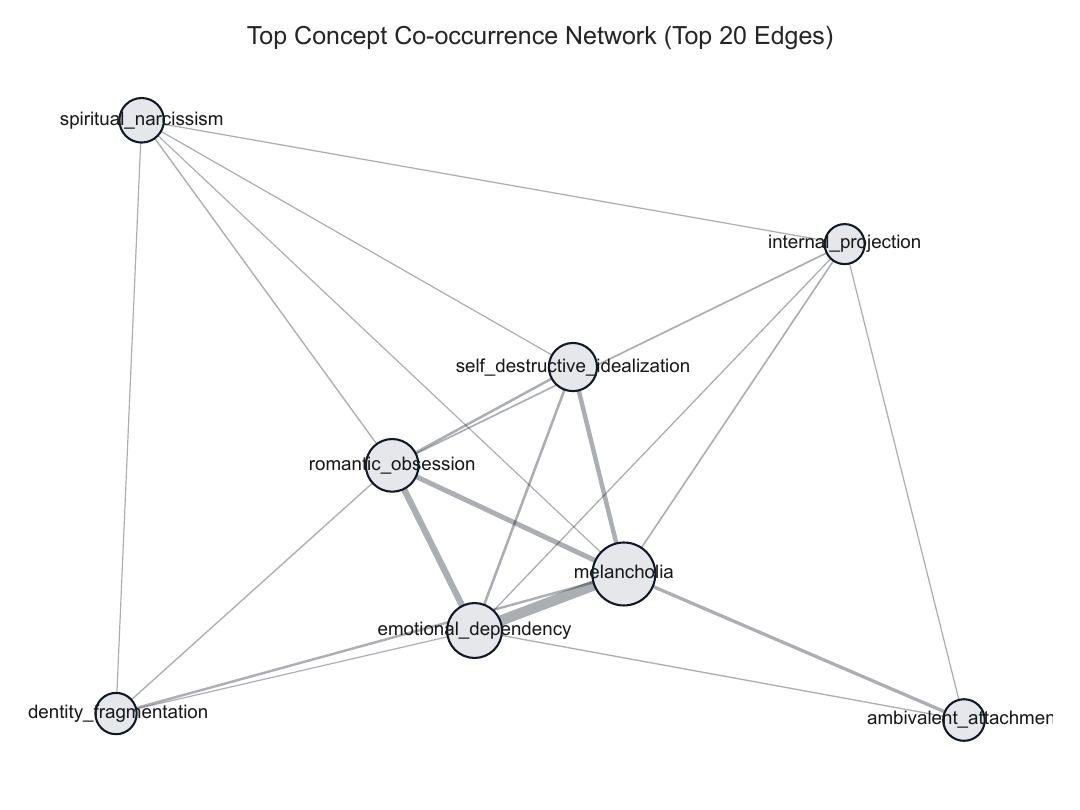}
\caption{Top concept co-occurrence network (top 20 edges by confidence-weighted co-activation).}
\label{fig:network}
\end{figure}

\begin{figure}[ht]
\centering
\includegraphics[width=0.82\linewidth]{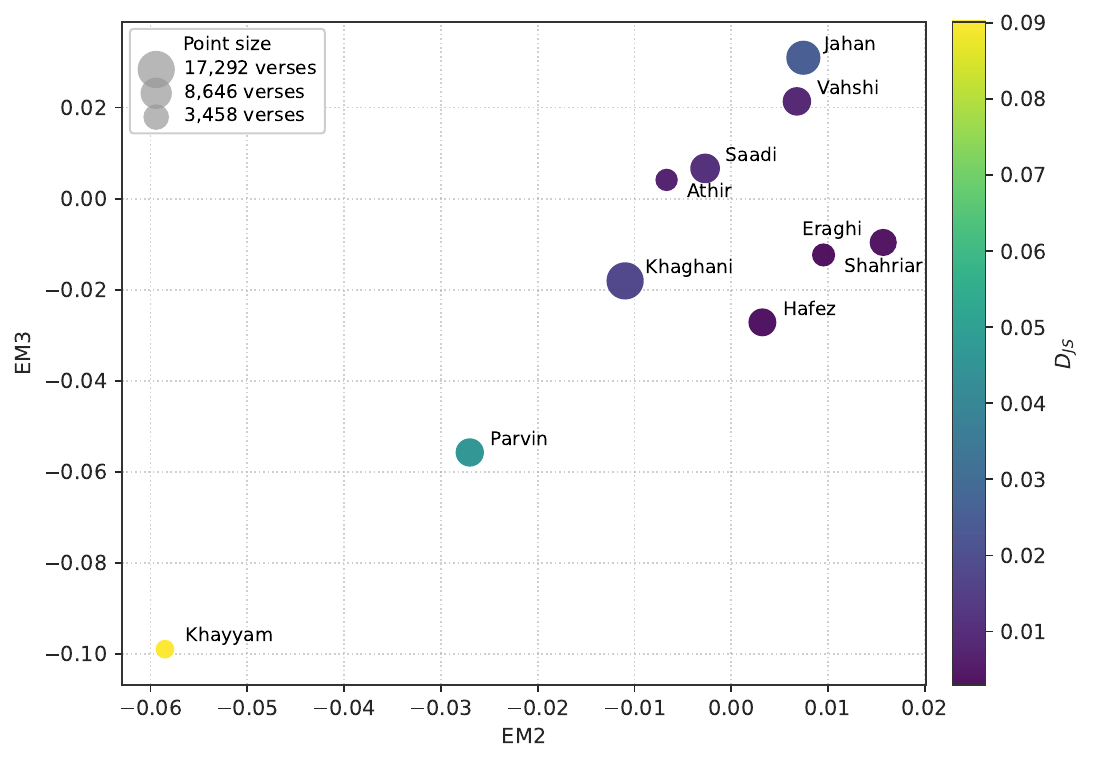}
\caption{Eigenmood embedding in two dimensions (EM2 vs EM3). Point size scales with verse count and color encodes $D_{JS}$.}
\label{fig:scatter}
\end{figure}

The strongest co-occurrence edges form a dense core anchored by \concept{melancholia}, \concept{emotional\_dependency}, and \concept{romantic\_obsession}. Figure~\ref{fig:network} visualizes the top 20 edges; Table~\ref{tab:edges} lists the top 12. Spectral decomposition of the filtered 8-concept graph yields non-trivial eigenvalues $\lambda_1=584.321$, $\lambda_2=1039.565$, and $\lambda_3=1290.029$. Table~\ref{tab:eigen_loadings} reports interpretable loadings of Eigenmood axes: EM2 contrasts \concept{internal\_projection} against \concept{identity\_fragmentation}, while EM1 highlights \concept{spiritual\_narcissism} against \concept{ambivalent\_attachment} and \concept{melancholia}. Poet embeddings in EM2--EM3 space are shown in Figure~\ref{fig:scatter}.

\begin{table}[ht]
\centering
\caption{Top confidence-weighted concept co-occurrence edges.}
\label{tab:edges}
\small
\begin{tabular}{lr}
\toprule
Concept pair & Edge weight \\
\midrule
\concept{emotional\_dependency}--\concept{melancholia} & 4739.1 \\
\concept{emotional\_dependency}--\concept{romantic\_obsession} & 2650.9 \\
\concept{melancholia}--\concept{romantic\_obsession} & 1981.4 \\
\concept{melancholia}--\concept{self\_destructive\_idealization} & 1726.7 \\
\concept{ambivalent\_attachment}--\concept{melancholia} & 1124.0 \\
\concept{emotional\_dependency}--\concept{self\_destructive\_idealization} & 770.6 \\
\concept{romantic\_obsession}--\concept{self\_destructive\_idealization} & 737.6 \\
\concept{identity\_fragmentation}--\concept{melancholia} & 699.0 \\
\concept{internal\_projection}--\concept{romantic\_obsession} & 324.4 \\
\concept{internal\_projection}--\concept{melancholia} & 286.3 \\
\concept{romantic\_obsession}--\concept{spiritual\_narcissism} & 166.4 \\
\concept{identity\_fragmentation}--\concept{romantic\_obsession} & 160.7 \\
\bottomrule
\end{tabular}
\end{table}

\begin{table}[ht]
\centering
\caption{Largest-magnitude concept loadings for the first three Eigenmood axes (filtered 8-concept graph).}
\label{tab:eigen_loadings}
\small
{\renewcommand{\arraystretch}{1.15}%
\begin{tabular}{l p{10.6cm}}
\toprule
Axis & Largest-magnitude loadings (concept, value) \\
\midrule
EM1 &
\begin{tabular}[t]{@{}p{7.8cm} S[table-format=-1.3]@{}}
\concept{spiritual\_narcissism} & 0.931 \\
\concept{ambivalent\_attachment} & -0.217 \\
\concept{melancholia} & -0.135 \\
\end{tabular}
\\
EM2 &
\begin{tabular}[t]{@{}p{7.8cm} S[table-format=-1.3]@{}}
\concept{internal\_projection} & 0.837 \\
\concept{identity\_fragmentation} & -0.531 \\
\concept{self\_destructive\_idealization} & -0.079 \\
\end{tabular}
\\
EM3 &
\begin{tabular}[t]{@{}p{7.8cm} S[table-format=-1.3]@{}}
\concept{identity\_fragmentation} & -0.707 \\
\concept{ambivalent\_attachment} & 0.537 \\
\concept{internal\_projection} & -0.381 \\
\end{tabular}
\\
\bottomrule
\end{tabular}
}
\end{table}

Eigenmood sensitivity depends on how the concept graph is normalized. Recomputing the spectral basis after replacing confidence-weighted edges with simple co-occurrence counts yields near-identical modes (absolute correlations $>0.999$ across the first three non-trivial eigenvectors, and $>0.9999$ correlation of poet coordinates), indicating that the embedding is not driven by confidence scaling alone. In contrast, switching from the unnormalized Laplacian to the symmetric normalized Laplacian changes the geometry more substantially: after best matching of modes, eigenvector correlations are 0.845/0.497/0.335 and poet coordinate correlations are 0.90 for EM2 and 0.70 for EM3. We therefore treat Laplacian normalization as a modeling choice and report unnormalized results as the main analysis, while recommending normalized variants as a sensitivity check.

\section{Poet-Level Psychological Profiles and Historical Interpretation}

\subsection{Poet: Khaghani}

Khaghani contributes 17,292 verses, with an abstention rate of 0.357 and a baseline divergence of $D_{JS}=0.0179$. This profile is moderately distinctive in magnitude but directionally specific in composition. The confidence-weighted distribution is centered on Melancholia, Spiritual Narcissism, and Self-Destructive Idealization, while the signed lift profile shows positive displacement in Spiritual Narcissism ($\Delta=+0.086$) and Self-Destructive Idealization ($\Delta=+0.034$), and negative displacement in Emotional Dependency ($\Delta=-0.089$) and Romantic Obsession ($\Delta=-0.044$). The embedding coordinates (EM1 $=0.092$, EM2 $=-0.011$, EM3 $=-0.018$) indicate a profile that is less defined by attachment volatility and more by rhetorical self-positioning and high-register stance. As shown in Appendix B (Figs.~\ref{fig:khaghani_p1}, \ref{fig:khaghani_p2}, \ref{fig:khaghani_p3}; Tables~\ref{tab:poet_profile_khaghani_metrics} and \ref{tab:khaghani_label_exemplars}), the distribution and retrieval evidence align: global-scale mass and verse-level exemplars point to the same dominant motifs.

This configuration is consistent with scholarship on Khaghani as a rhetorically dense qasida poet whose diction often stages authority, intellectual rank, and moral pressure under formal constraint \cite{iranica_khaghani}. The present profile does not claim a psychological diagnosis of the historical author. It identifies an operationally stable textual signature: a recurrent coupling of lament with elevated self-referential posture and sacrificial imagery. A plausible contextual factor is genre ecology: in panegyric and high-court registers, performative distinction and adversity can co-appear as mutually reinforcing rhetorical resources. The appendix evidence supports an interpretive alignment rather than causal attribution. In this sense, the profile contributes to DH comparison by quantifying where Khaghani departs from corpus baseline while preserving uncertainty and abstention as first-class limits on interpretation.

\subsection{Poet: Jahan}

Jahan's subcorpus is large (12,299 verses) with low abstention (0.065) and substantial divergence ($D_{JS}=0.0250$), indicating a strongly recoverable and comparatively distinctive signal under the ontology. The confidence-weighted mass concentrates in Melancholia, Emotional Dependency, and Romantic Obsession. Relative to baseline, the largest positive shifts are Emotional Dependency ($\Delta=+0.107$) and Romantic Obsession ($\Delta=+0.027$), while Spiritual Narcissism ($\Delta=-0.063$) and Self-Destructive Idealization ($\Delta=-0.050$) are comparatively suppressed. Eigenmood coordinates (EM1 $=-0.068$, EM2 $=0.007$, EM3 $=0.031$) place Jahan away from the high-distinction pole and closer to attachment-centered discourse. As shown in Appendix B (Figs.~\ref{fig:jahan_p1}, \ref{fig:jahan_p2}, \ref{fig:jahan_p3}; Tables~\ref{tab:poet_profile_jahan_metrics} and \ref{tab:jahan_axis_exemplars}), both aggregate and retrieval levels are coherent: the most elevated verses on Jahan's dominant axes repeatedly foreground dependence, longing, and oscillatory relational stance.

This profile is consistent with scholarship on Jahan Malek Khatun's lyric production within courtly and gendered literary conventions \cite{iranica_jahan_malek}. The relevant interpretive point is not to reduce the corpus to personal affect, but to identify which rhetorical-affective configurations are disproportionately activated across many verses. A plausible contextual factor is the formal economy of ghazal-like address, where devotion, complaint, and unresolved proximity are codified as productive tensions rather than contradictions. The uncertainty-aware aggregation matters directly here: the low abstention rate indicates that the ontology maps relatively often in this subcorpus, yet interpretive caution remains necessary because metaphorical density can still blur boundaries between dependence and obsession. The appendix figures and tables therefore function as auditable evidence for an alignment between quantitative profile and established literary framing.

\subsection{Poet: Saadi}

Saadi contributes 6,864 verses with abstention 0.137 and modest divergence from baseline ($D_{JS}=0.0106$). The profile is centered on Romantic Obsession, Melancholia, and Emotional Dependency, but the lift structure is asymmetrical: Romantic Obsession ($\Delta=+0.069$) and Emotional Dependency ($\Delta=+0.026$) are elevated, while Melancholia ($\Delta=-0.069$) and Spiritual Narcissism ($\Delta=-0.045$) are reduced. This suggests a distribution with meaningful positive emphasis in relational focus but without strong displacement toward despair-heavy or self-exalting registers. Eigenmood coordinates (EM1 $=-0.047$, EM2 $=-0.003$, EM3 $=0.007$) indicate baseline-adjacent positioning with mild directional pull. As shown in Appendix B (Figs.~\ref{fig:saadi_p1}, \ref{fig:saadi_p2}, \ref{fig:saadi_p3}; Tables~\ref{tab:poet_profile_saadi_metrics} and \ref{tab:saadi_label_exemplars}), this is a stable profile in which the top-concept mass is identifiable but not extreme, and verse retrieval reproduces the same moderate structure.

This pattern is consistent with scholarship describing Saadi's broad ethical and pedagogical range across lyric and didactic forms \cite{iranica_saadi,britannica_saadi}. A plausible contextual factor is Saadi's stylistic breadth: his corpus sustains intimate relational vocabulary while frequently re-routing affect into moral reflection and social instruction. That mixture can raise obsession and dependency markers without producing high divergence overall, because balancing registers remain close to the corpus baseline. The profile should therefore be interpreted as comparative weighting, not as categorical identity. In DH terms, Saadi appears as a stabilizing point in the inter-poet geometry: sufficiently distinctive to be measurable, yet sufficiently central to illuminate how stronger outliers depart from a shared classical distribution.

\subsection{Poet: Vahshi}

Vahshi's corpus contains 5,727 verses with abstention 0.289 and relatively low divergence ($D_{JS}=0.0085$). The top mass falls on Melancholia, Self-Destructive Idealization, and Emotional Dependency, while signed differences show positive lift in Self-Destructive Idealization ($\Delta=+0.035$) and Ambivalent Attachment ($\Delta=+0.031$), with negative lift in Emotional Dependency ($\Delta=-0.060$) and Romantic Obsession ($\Delta=-0.042$). This pattern indicates that Vahshi's distinctiveness is less about absolute dominance and more about a shift in how suffering is configured: less fixated pursuit, more volatile attachment and sacrificial framing. The embedding (EM1 $=0.020$, EM2 $=0.007$, EM3 $=0.021$) is near the central cloud but directionally tilted toward these contrasts. As shown in Appendix B (Figs.~\ref{fig:vahshi_p1}, \ref{fig:vahshi_p2}, \ref{fig:vahshi_p3}; Tables~\ref{tab:poet_profile_vahshi_metrics} and \ref{tab:vahshi_axis_exemplars}), both uncertainty bands and exemplar retrieval support this nuanced structure rather than a single-label reading.

This is consistent with scholarship on Vahshi Bafqi and intensified love-lament within Safavid lyric repertoires \cite{iranica_vahshi}. A plausible contextual factor is the prominence of poetic conventions that valorize suffering and unstable reciprocity, producing high local intensity even when global divergence remains moderate. The profile should not be read as evidence of authorial pathology. It is a distributional description of textual tendencies under an explicit ontology and abstention regime. The appendix evidence is especially useful here because it shows how modest $D_{JS}$ can still correspond to interpretable directional lift in selected constructs; in other words, low overall distance does not imply interpretive flatness.

\subsection{Poet: Parvin}

Parvin contributes 5,573 verses with a high abstention rate (0.384) and the second-largest divergence in the corpus ($D_{JS}=0.0459$). The profile concentrates on Melancholia, Self-Destructive Idealization, and Spiritual Narcissism, with strong positive lift for Self-Destructive Idealization ($\Delta=+0.073$) and Melancholia ($\Delta=+0.063$), and strong negative lift for Romantic Obsession ($\Delta=-0.126$) and Emotional Dependency ($\Delta=-0.112$). The coordinates (EM1 $=0.039$, EM2 $=-0.027$, EM3 $=-0.056$) place Parvin away from attachment-heavy zones and toward a more didactic, contrastive affective regime. As shown in Appendix B (Figs.~\ref{fig:parvin_p1}, \ref{fig:parvin_p2}, \ref{fig:parvin_p3}; Tables~\ref{tab:poet_profile_parvin_metrics} and \ref{tab:parvin_label_exemplars}), the profile is supported by coherent verse-level evidence despite high abstention, which signals that many lines remain ontologically underdetermined.

This configuration is consistent with scholarship on Parvin Etesami emphasizing dialogic didacticism and moral argumentation \cite{iranica_parvin}. A plausible contextual factor is that Parvin's rhetorical architecture frequently prioritizes ethical confrontation, allegorical reversal, and admonitory contrast, which can elevate constructs linked to suffering and sacrificial framing while suppressing intimate romantic fixation. The high abstention rate reinforces interpretive caution: a sizable fraction of verses does not map cleanly to the current ontology. Nevertheless, the non-abstained mass exhibits a stable and distinctive orientation. In DH terms, Parvin provides a clear case where divergence and abstention must be read jointly, not competitively.

\subsection{Poet: Hafez}

Hafez contributes 5,221 verses with abstention 0.182 and near-baseline divergence ($D_{JS}=0.0035$). The top concepts are Melancholia, Romantic Obsession, and Emotional Dependency, but the lift magnitudes are small: Romantic Obsession ($\Delta=+0.018$) and Identity Fragmentation ($\Delta=+0.016$) are mildly elevated, while Emotional Dependency ($\Delta=-0.038$) and Ambivalent Attachment ($\Delta=-0.016$) are mildly reduced. Coordinates (EM1 $=0.019$, EM2 $=0.003$, EM3 $=-0.027$) likewise indicate central placement. As shown in Appendix B (Figs.~\ref{fig:hafez_p1}, \ref{fig:hafez_p2}, \ref{fig:hafez_p3}; Tables~\ref{tab:poet_profile_hafez_metrics} and \ref{tab:hafez_axis_exemplars}), Hafez's profile is not flat but balanced: multiple constructs are active, yet none dominates enough to generate large baseline distance.

This pattern is consistent with scholarship on Hafez that emphasizes polyvalent wine and tavern imagery, semantic indirection, and sustained critique of hypocrisy across lyric registers \cite{iranica_hafez,britannica_hafez}. A plausible contextual factor is deliberate multivalence: ghazal rhetoric in Hafez repeatedly supports simultaneous devotional, erotic, and critical readings, which can distribute mass across neighboring constructs and damp aggregate divergence. The resulting quantitative centrality should not be mistaken for interpretive simplicity. Instead, it identifies Hafez as a useful baseline-adjacent anchor for comparative work: poets with higher $D_{JS}$ depart from a distribution that Hafez approximates, while close reading still reveals rich local tension not reducible to any single ontology label.

\subsection{Poet: Eraghi}

Eraghi contributes 4,668 verses with low abstention (0.081) and low divergence ($D_{JS}=0.0040$). The highest mass remains in Melancholia, Emotional Dependency, and Romantic Obsession, but the signed profile differs from neighboring poets: Emotional Dependency ($\Delta=+0.043$) and Internal Projection ($\Delta=+0.013$) are elevated, while Melancholia ($\Delta=-0.030$) and Self-Destructive Idealization ($\Delta=-0.028$) are reduced. Eigenmood coordinates (EM1 $=0.014$, EM2 $=0.016$, EM3 $=-0.010$) place Eraghi near center with a slight tilt toward inwardized affect. As shown in Appendix B (Figs.~\ref{fig:eraghi_p1}, \ref{fig:eraghi_p2}, \ref{fig:eraghi_p3}; Tables~\ref{tab:poet_profile_eraghi_metrics} and \ref{tab:eraghi_label_exemplars}), distributional and retrieval evidence converge on this orientation: internalization motifs appear consistently without requiring large global deviation.

This is consistent with scholarship on Fakhr-al-Din Eraghi and Sufi lyric-mystical poetics, where interiorization of union and affective inwardness are canonical strategies \cite{iranica_eraqi}. A plausible contextual factor is the metaphysical localization of meaning: instead of emphasizing external pursuit or complaint, many lines stage transformation within the self. Under the present ontology, that tendency is captured by Internal Projection and related constructs. Interpretively, the key point is proportionality: Eraghi is not a high-divergence outlier, but a directionally coherent baseline-adjacent poet. This distinction matters for DH inference because centrality and coherence can coexist.

\subsection{Poet: Shahriar}

Shahriar's corpus includes 1,970 verses, with abstention 0.141 and very low divergence ($D_{JS}=0.0030$). The profile is concentrated in Melancholia, Romantic Obsession, and Emotional Dependency, but lift is modest: Melancholia ($\Delta=+0.041$) and Internal Projection ($\Delta=+0.013$) are elevated, while Emotional Dependency ($\Delta=-0.046$) and Spiritual Narcissism ($\Delta=-0.007$) are reduced. Coordinates (EM1 $=-0.007$, EM2 $=0.010$, EM3 $=-0.012$) again place Shahriar near baseline geometry. As shown in Appendix B (Figs.~\ref{fig:shahriar_p1}, \ref{fig:shahriar_p2}, \ref{fig:shahriar_p3}; Tables~\ref{tab:poet_profile_shahriar_metrics} and \ref{tab:shahriar_axis_exemplars}), this is a coherent but low-distance profile, with retrieval evidence emphasizing lament-oriented expressions rather than extreme spectral displacement.

The profile is consistent with scholarship on Shahriar's modern reworking of classical lyric resources \cite{iranica_shahriar}. A plausible contextual factor is historical layering: modern diction and sensibility may intensify direct lament while still preserving many inherited affective conventions, yielding limited divergence from the corpus baseline. The abstention level is moderate, indicating usable coverage under the ontology but nontrivial limits for ambiguous lines. Accordingly, the interpretation remains comparative and probabilistic: Shahriar is best read as a baseline-near poet with selective emphasis, not as a categorical endpoint in the emotional space.

\subsection{Poet: Athir}

Athir contributes 1,603 verses with abstention 0.138 and low-to-moderate divergence ($D_{JS}=0.0073$). The top concept mass lies in Melancholia, Romantic Obsession, and Emotional Dependency, but the discriminating signal appears in lift: Romantic Obsession ($\Delta=+0.052$) and Ambivalent Attachment ($\Delta=+0.019$) are elevated, whereas Spiritual Narcissism ($\Delta=-0.041$) and Emotional Dependency ($\Delta=-0.031$) are reduced. Coordinates (EM1 $=-0.044$, EM2 $=-0.007$, EM3 $=0.004$) suggest a profile with mild orientation away from distinction-heavy poles. As shown in Appendix B (Figs.~\ref{fig:athir_p1}, \ref{fig:athir_p2}, \ref{fig:athir_p3}; Tables~\ref{tab:poet_profile_athir_metrics} and \ref{tab:athir_label_exemplars}), the uncertainty-aware pattern is stable: moderate divergence, but interpretable directional deviations in attachment-adjacent constructs.

This is consistent with scholarship on Athir Akhsikati as a court poet in a rhetorically compact qasida environment \cite{iranica_athir_akhsikati}. A plausible contextual factor is formal compression: rhetorical density can concentrate obsessional and ambivalent motifs while reducing explicit self-exalting posture under this ontology. Because the corpus is smaller than major poets, the profile should be interpreted with attention to uncertainty bands and abstention behavior. Even so, the alignment between aggregate distributions and retrieved exemplars supports a cautious claim of comparative individuality.

\subsection{Poet: Khayyam}

Khayyam is the smallest subcorpus (356 verses) and the strongest outlier by divergence ($D_{JS}=0.0901$), with high abstention (0.385). The confidence-weighted profile is dominated by Melancholia, Identity Fragmentation, and Spiritual Narcissism. Relative to baseline, Melancholia ($\Delta=+0.232$) and Identity Fragmentation ($\Delta=+0.101$) are strongly elevated, while Emotional Dependency ($\Delta=-0.169$) and Romantic Obsession ($\Delta=-0.116$) are strongly reduced. Embedding coordinates (EM1 $=0.024$, EM2 $=-0.059$, EM3 $=-0.099$) place Khayyam at a clear geometric extreme. As shown in Appendix B (Figs.~\ref{fig:khayyam_p1}, \ref{fig:khayyam_p2}, \ref{fig:khayyam_p3}; Tables~\ref{tab:poet_profile_khayyam_metrics} and \ref{tab:khayyam_axis_exemplars}), both lift and retrieval evidence indicate a sustained orientation toward existential and paradox-framed discourse rather than relational dependency.

This profile is consistent with scholarship on Omar Khayyam that foregrounds impermanence, skepticism, and existential reflection in the rubai tradition \cite{iranica_khayyam,britannica_khayyam}. The small sample and high abstention require caution: outlier status in $D_{JS}$ can be amplified when coverage is limited. Nevertheless, the directionality is robust across multiple indicators and is not reducible to a single concept spike. The most defensible reading is that Khayyam's corpus activates a different mixture of affective constructs than the corpus-wide norm, and that this difference is textually inspectable in the appendix evidence.

\subsection{Cross-Poet Comparative Patterns}

\begin{figure}[t]
\centering
\includegraphics[width=0.82\linewidth]{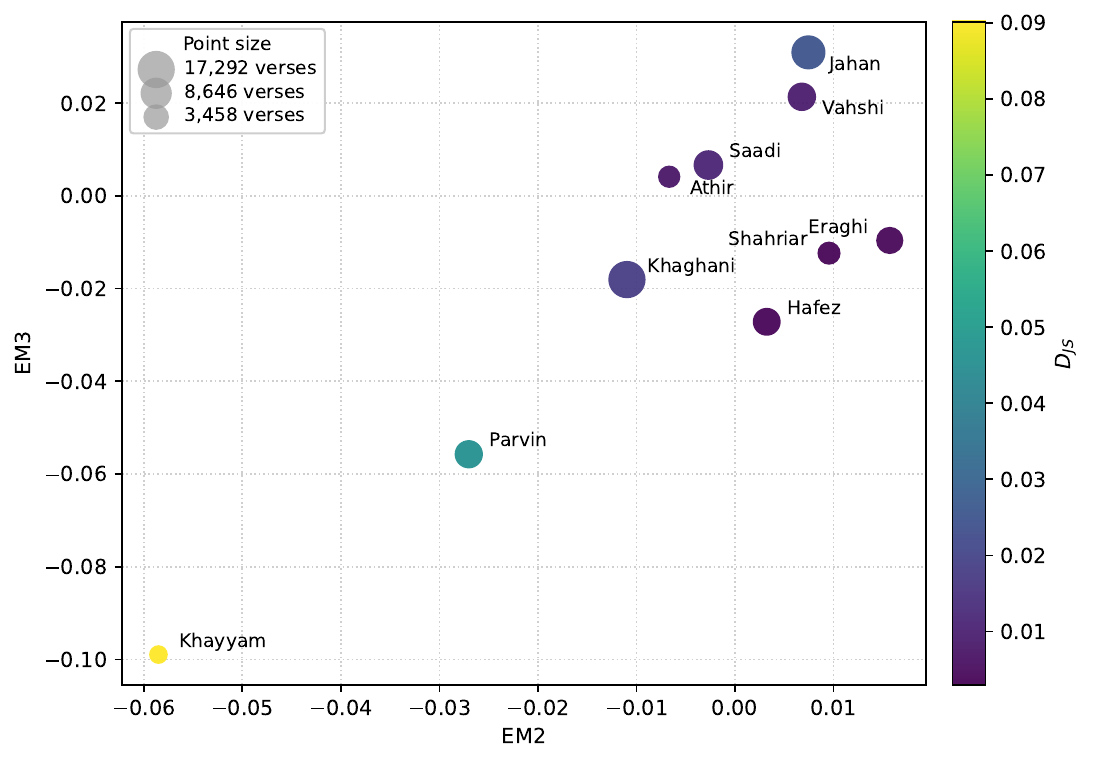}
\caption{Cross-poet Eigenmood embedding (EM2 vs EM3). Labels are offset outside markers; point size scales with verse count; color encodes $D_{JS}$.}
\label{fig:cross_poet_em2_em3}
\end{figure}

At the corpus level, the divergence ranking remains concentrated: Khayyam and Parvin occupy the high-distance end, while Hafez and Shahriar remain closest to baseline. Jahan and Khaghani occupy intermediate positions but for different reasons: Jahan with broad coverage and low abstention, Khaghani with high volume and high abstention but directional emphasis in distinction-associated constructs. As shown in Table~\ref{tab:cross_poet_js_rank} and Figure~\ref{fig:cross_poet_em2_em3}, magnitude and direction must be read together. $D_{JS}$ captures how far a poet moves from baseline, while EM2--EM3 geometry captures where the movement points in concept-coupled space.

\begin{table}[t]
\centering
\caption{Ranking of poets by divergence from the global baseline ($D_{JS}$).}
\label{tab:cross_poet_js_rank}
\footnotesize
\renewcommand{\arraystretch}{1.1}
\begin{tabular}{l S[table-format=1.4] S[table-format=1.3] S[table-format=5.0]}
\toprule
Poet & {$D_{JS}$} & {Abstain rate} & {Verses} \\
\midrule
Khayyam & 0.0901 & 0.385 & 356 \\
Parvin & 0.0459 & 0.384 & 5573 \\
Jahan & 0.0250 & 0.065 & 12299 \\
Khaghani & 0.0179 & 0.357 & 17292 \\
Saadi & 0.0106 & 0.137 & 6864 \\
Vahshi & 0.0085 & 0.289 & 5727 \\
Athir & 0.0073 & 0.138 & 1603 \\
Eraghi & 0.0040 & 0.081 & 4668 \\
Hafez & 0.0035 & 0.182 & 5221 \\
Shahriar & 0.0030 & 0.141 & 1970 \\
\bottomrule
\end{tabular}
\end{table}

A cautious historical interpretation follows from this joint reading. One cluster remains near the baseline cloud with low divergence and balanced concept mass (for example Hafez, Shahriar, Eraghi), while a second cluster includes poets with stronger directional displacement and higher abstention-sensitive individuality (especially Khayyam and Parvin). The abstention-divergence association remains moderate and informative rather than determinative. In DH terms, these comparative profiles support probabilistic cultural-psychological modeling: they identify differential rhetorical-affective tendencies across poets without collapsing textual tradition into clinical categories or monocausal historical claims.

\FloatBarrier

\section{Interpretive Case Studies}

Eigenmood geometry is designed to support a distant-to-close workflow: the same spectral basis used for poet embeddings can score individual verses. Given eigenvector $u_k$, we define a verse score
\begin{equation}
s_v^{(k)}=\sum_{c\in\mathcal{C}} (1-a_v)\,p_{v,c}\,\mathbf{1}[c\in L_v]\,u_k(c),
\label{eq:verse_score}
\end{equation}
which provides an index for retrieving extreme-scoring verses that exemplify an axis in textual practice. Importantly, abstained verses do not enter the retrieval pool, and low-confidence labels contribute weakly, so interpretive exemplars are selected from comparatively well-supported evidence rather than from forced assignments.

\subsection{Eigenmood axis case studies}

\subsubsection{EM1: spiritual distinction vs.\ melancholic dependence}

\begin{figure}[!htbp]
\centering
\includegraphics[width=0.92\linewidth]{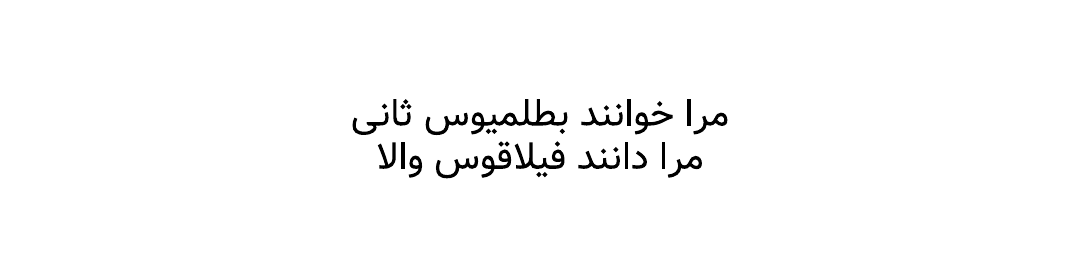}
\caption{EM1 high-side retrieval exemplar for Khaghani.}
\label{fig:app_em1_khaghani}
\end{figure}

EM1 loads positively on \concept{spiritual\_narcissism} and negatively on \concept{melancholia} and \concept{ambivalent\_attachment}, contrasting rhetorical self-distinction with registers of sorrowful dependence. A high-EM1 exemplar from \textsc{Khaghani} is retrieved with model label \concept{spiritual\_narcissism} at confidence 0.90 (Fig.~\ref{fig:app_em1_khaghani}):
\noindent\textit{Translation (gloss).} ``They call me a second Ptolemy; they know me as a sage of the age.''

\begin{figure}[!htbp]
\centering
\includegraphics[width=0.92\linewidth]{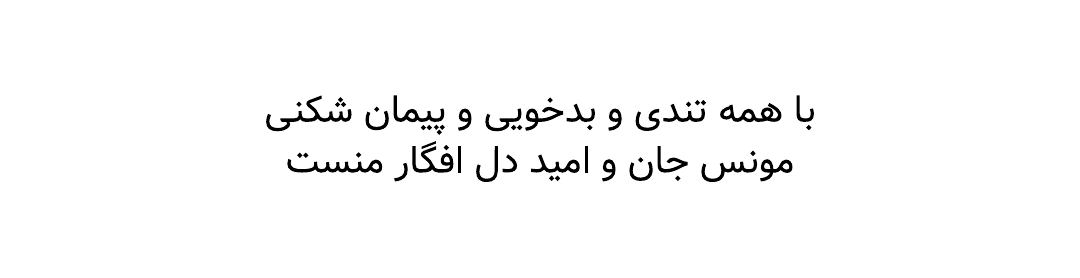}
\caption{EM1 low-side retrieval exemplar for Jahan.}
\label{fig:app_em1_jahan}
\end{figure}

By contrast, a low-EM1 exemplar from \textsc{Jahan} combines \concept{melancholia}, \concept{emotional\_dependency}, and \concept{ambivalent\_attachment} with maximum confidence 0.85 (Fig.~\ref{fig:app_em1_jahan}):
\noindent\textit{Translation (gloss).} ``With all your harshness and ill-temper and vow-breaking, you are still the companion of my soul and the hope of my afflicted heart.''

Close reading clarifies why these exemplars are interpretable in a DH sense. Khaghani's couplet belongs to a courtly rhetoric in which learned authority and cosmological knowledge can be staged as poetic capital, including self-ascription of exceptional insight \cite{meisami1987medieval}. Our label \concept{spiritual\_narcissism} does not claim clinical grandiosity; it operationalizes a rhetorical posture of spiritual or intellectual distinction. Jahan's couplet, by contrast, performs attachment under injury: the beloved is simultaneously blamed (harshness, broken promises) and affirmed as the speaker's sole support. This oscillatory dependence is a recognizable lyric configuration in Persian love discourse, where suffering and devotion are often co-constitutive \cite{schimmel1992brocade}. Uncertainty matters here because ambiguous panegyric or complaint is common; the workflow therefore anchors interpretation in verses with strong confidence and excludes abstained lines that the model could not reliably map to the ontology.

\subsubsection{EM2: internal projection vs.\ fragmented self-positioning}

\begin{figure}[!htbp]
\centering
\includegraphics[width=0.92\linewidth]{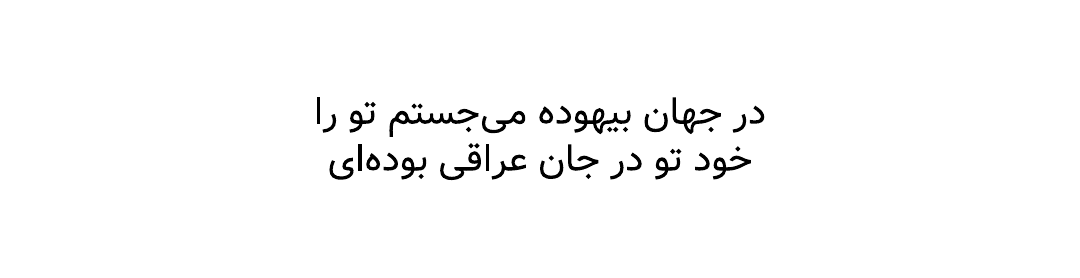}
\caption{EM2 high-side retrieval exemplar for Eraghi.}
\label{fig:app_em2_eraghi}
\end{figure}

EM2 contrasts \concept{internal\_projection} against \concept{identity\_fragmentation}. A high-EM2 exemplar from \textsc{Eraghi} is retrieved with \concept{internal\_projection} at confidence 0.95 (Fig.~\ref{fig:app_em2_eraghi}):
\noindent\textit{Translation (gloss).} ``In the world I searched for you in vain; you were yourself within the soul of Eraghi.''

\begin{figure}[!htbp]
\centering
\includegraphics[width=0.92\linewidth]{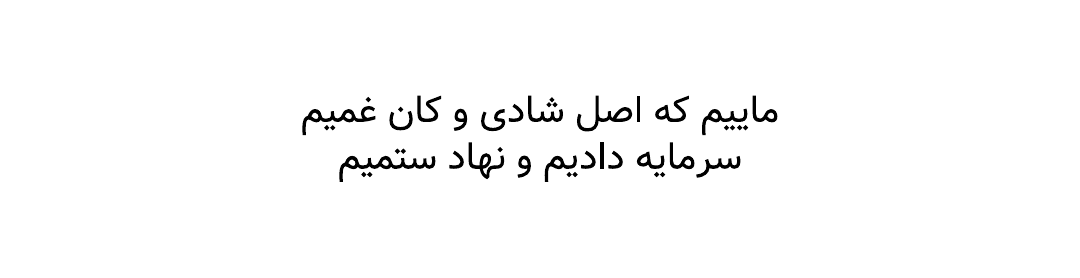}
\caption{EM2 low-side retrieval exemplar for Khayyam.}
\label{fig:app_em2_khayyam}
\end{figure}

A low-EM2 exemplar from \textsc{Khayyam} is retrieved with \concept{identity\_fragmentation} at confidence 0.90 (Fig.~\ref{fig:app_em2_khayyam}):
\noindent\textit{Translation (gloss).} ``We are at once the harvest of joy and the blood of sorrow; we paid the capital and reaped only injustice.''

Eraghi's inward localization of the beloved or the divine is a conventional Sufi move: union is expressed not as external possession but as interior presence \cite{debruijn1997sufi,schimmel1992brocade}. In this rhetorical economy, the ``I'' becomes a site of transformation, and searching the world is figured as ignorance of what is already within. Khayyam's quatrain, by contrast, stages a divided speaker position in which joy and grief, profit and injustice, are held in paradoxical coexistence, aligning with scholarly readings of Khayyam's skeptical and existential register \cite{aminrazavi2005wine}. The uncertainty-aware retrieval procedure is critical because such paradox is ubiquitous in Persian verse; forcing a single affect label would flatten the texture of the quatrain. By selecting high-confidence instances and reporting confidence explicitly, the workflow supports cautious interpretive anchoring while keeping the possibility of alternative readings open.

\subsubsection{EM3: ambivalent attachment vs.\ ontological division}

\begin{figure}[!htbp]
\centering
\includegraphics[width=0.92\linewidth]{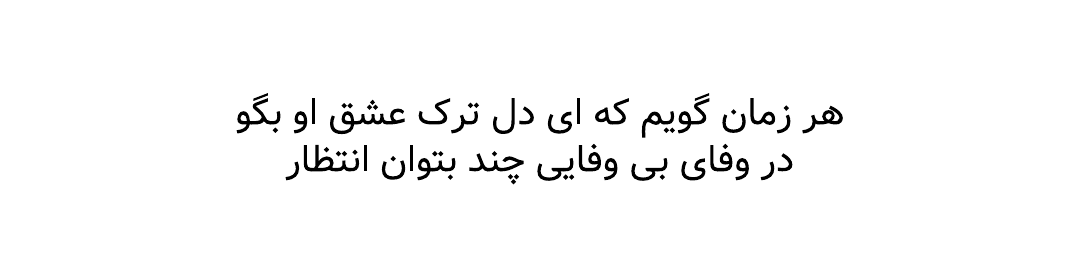}
\caption{EM3 high-side retrieval exemplar for Jahan.}
\label{fig:app_em3_jahan}
\end{figure}

EM3 loads positively on \concept{ambivalent\_attachment} and negatively on \concept{identity\_fragmentation} and \concept{internal\_projection}, yielding an axis that contrasts oscillatory attachment discourse with forms of self or world division. A positive-EM3 exemplar from \textsc{Jahan} is retrieved with \concept{ambivalent\_attachment} and \concept{melancholia} at maximum confidence 0.85 (Fig.~\ref{fig:app_em3_jahan}):
\noindent\textit{Translation (gloss).} ``Each moment I tell my heart: leave love and speak of something else; yet in fidelity, how long can one wait for the faithless?''

\begin{figure}[!htbp]
\centering
\includegraphics[width=0.92\linewidth]{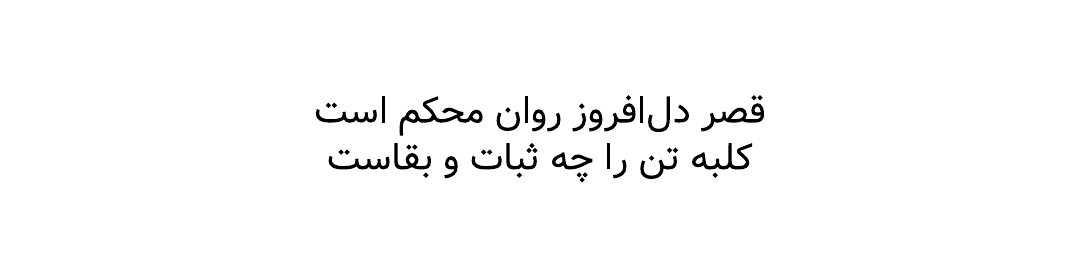}
\caption{EM3 low-side retrieval exemplar for Parvin.}
\label{fig:app_em3_parvin}
\end{figure}

A negative-EM3 exemplar from \textsc{Parvin} is retrieved with \concept{identity\_fragmentation} at confidence 0.85 (Fig.~\ref{fig:app_em3_parvin}):
\noindent\textit{Translation (gloss).} ``The soul's radiant palace is firm; what stability and permanence has the body's hut?''

The Jahan couplet performs a characteristic attachment oscillation: it commands detachment while simultaneously returning to the beloved as the horizon of expectation. This tension between renunciation and return is central to Persian lyric and is often thematized through the formal opposition of separation and union \cite{schimmel1992brocade,debruijn1997sufi}. Parvin's didactic contrast between soul and body stages a different kind of division: the self is split across metaphysical registers, and the body is rhetorically demoted to an unstable dwelling. Our label \concept{identity\_fragmentation} operationalizes such divided self-positioning without implying psychopathology. Uncertainty is crucial for this axis because moral and metaphysical couplets can be semantically dense; the abstention mechanism prevents over-interpretation when the model cannot justify mapping to the ontology.

\subsection{Divergence-driven poet contrasts}

\subsubsection{Khayyam (high divergence) vs.\ Hafez (baseline-adjacent)}

\begin{figure}[!htbp]
\centering
\includegraphics[width=0.92\linewidth]{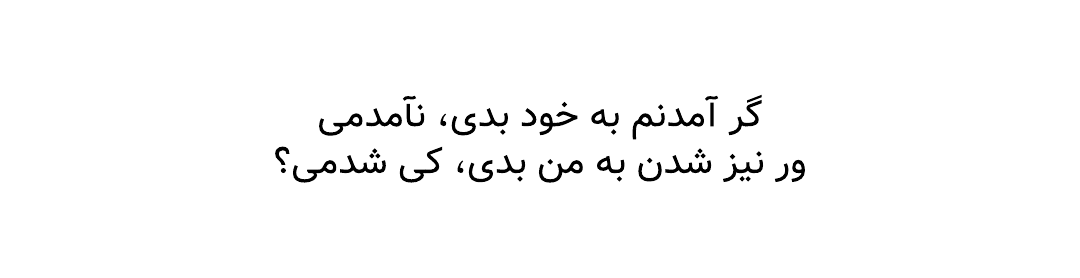}
\caption{Divergence-linked retrieval exemplar for Khayyam.}
\label{fig:app_jsd_khayyam}
\end{figure}

Distributional individuality identifies \textsc{Khayyam} as the most divergent poet from the global baseline ($D_{JS}=0.0901$), whereas \textsc{Hafez} is close to baseline ($0.0035$). A Khayyam exemplar aligned with this divergence is a high-confidence \concept{identity\_fragmentation} quatrain (confidence 0.85; Fig.~\ref{fig:app_jsd_khayyam}):
\noindent\textit{Translation (gloss).} ``If my coming were by my own choosing, I would not have come; and if becoming me were possible, why did I become?''

\begin{figure}[!htbp]
\centering
\includegraphics[width=0.92\linewidth]{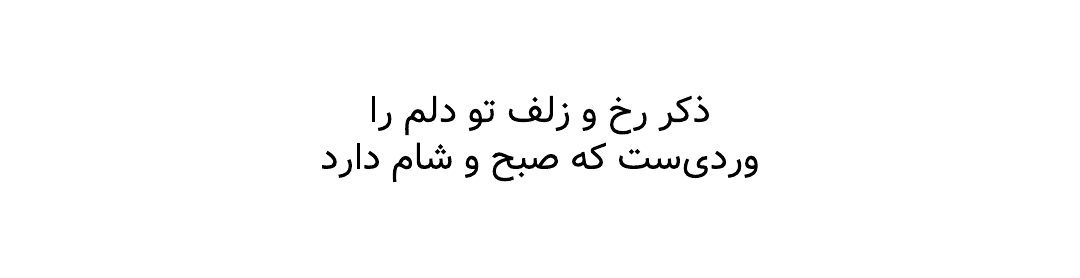}
\caption{Divergence-linked retrieval exemplar for Hafez.}
\label{fig:app_jsd_hafez}
\end{figure}

By contrast, a Hafez exemplar emphasizes fixation on the beloved's image and returns repeatedly to sensory memory (confidence 0.90 for \concept{romantic\_obsession}; Fig.~\ref{fig:app_jsd_hafez}):
\noindent\textit{Translation (gloss).} ``The memory of your face and your tress is in my heart; it is the remedy for my pain, morning and night.''

Close reading illustrates how divergence connects to interpretive themes without collapsing into diagnosis. Khayyam's quatrain interrogates agency and identity through conditional paradox, a move frequently noted in philosophical readings of the \textit{rub\={a}\={\i}} tradition \cite{aminrazavi2005wine}. Hafez's couplet, by contrast, exemplifies the ghazal's ability to stage obsession as devotion and as aestheticized suffering, where the beloved's image becomes an interiorized medicine \cite{schimmel1992brocade,debruijn1997sufi}. Uncertainty-aware profiling matters for this contrast because Khayyam's subcorpus is small and has high abstention; the divergence signal is therefore interpreted with wider uncertainty bands and exemplified only through high-confidence verses rather than through marginal cases.

\subsubsection{Parvin (high divergence) vs.\ Shahriar (baseline-adjacent)}

\begin{figure}[!htbp]
\centering
\includegraphics[width=0.92\linewidth]{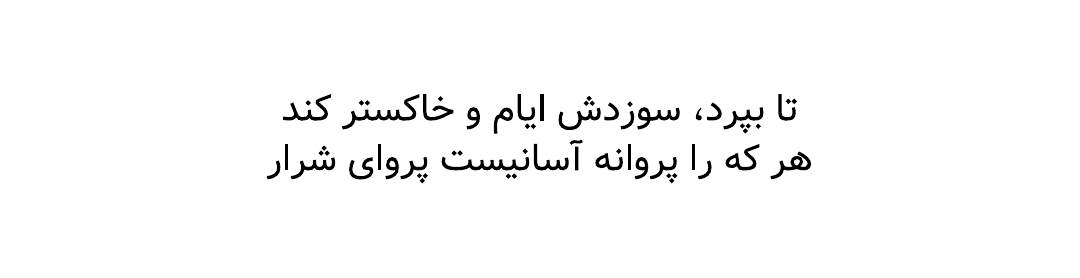}
\caption{Divergence-linked retrieval exemplar for Parvin.}
\label{fig:app_jsd_parvin}
\end{figure}

\textsc{Parvin} is the second most divergent poet ($D_{JS}=0.0459$), while \textsc{Shahriar} is close to baseline ($0.0030$). A Parvin exemplar associated with this divergence uses the moth-and-flame motif to frame self-endangering devotion, labeled \concept{self\_destructive\_idealization} at confidence 0.85 (Fig.~\ref{fig:app_jsd_parvin}):
\noindent\textit{Translation (gloss).} ``Let it fly; it burns, and the ashes smile. Whoever is a moth has no fear of the spark.''

\begin{figure}[!htbp]
\centering
\includegraphics[width=0.92\linewidth]{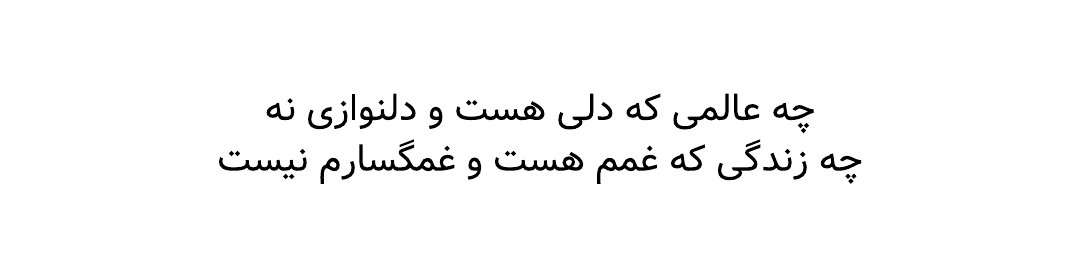}
\caption{Divergence-linked retrieval exemplar for Shahriar.}
\label{fig:app_jsd_shahriar}
\end{figure}

A Shahriar exemplar expresses sorrow and lack of consolation, labeled \concept{melancholia} at confidence 0.90 (Fig.~\ref{fig:app_jsd_shahriar}):
\noindent\textit{Translation (gloss).} ``What kind of world has a heart but no tenderness? What kind of life is sorrowful, yet has no companion in sorrow?''

Both couplets draw on long-standing Persian image repertoires, but they activate different rhetorical-affective configurations. Parvin's line deploys a canonical image of sacrificial attraction to the flame to moralize the cost of devotion, a pattern of idealization intertwined with self-harm that is widely documented in Persian poetic imagery \cite{schimmel1992brocade}. Shahriar's line, by contrast, uses direct lament to foreground the absence of care and shared grief, aligning with melancholic complaint. The uncertainty-aware framework shapes interpretation by specifying evidential boundaries: where the model abstains, we do not impute constructs; where confidence is high, we treat the verse as an anchor for cautious close reading rather than as proof of a psychological state.

\section{Validation and Human Evaluation}

Automatic annotation is only defensible for psychological reading if uncertainty claims correspond to annotator judgments. We therefore implement a controlled two-annotator evaluation on a stratified 500-verse sample drawn from the corpus. The evaluation targets three questions: (i) whether the ontology labels align with annotator judgments at the verse level, (ii) whether the abstention flag functions as a meaningful selective-prediction signal, and (iii) whether confidence scores support thresholded analysis and uncertainty-aware aggregation. The two-annotator validation was conducted by domain experts in psychology. All reported agreement, precision, abstention, and calibration statistics are computed directly from the completed validation sheet.

\subsection{Sampling and stratification}

We draw a stratified sample of 500 verses from the 61{,}573-verse corpus. Stratification is defined over (i) abstention status, (ii) confidence regime for non-abstained verses (low/medium/high bins based on the maximum predicted confidence in a verse), and (iii) label presence to ensure coverage of all concepts. Concretely, non-abstained verses are partitioned into low ($\max_c p_{v,c}<0.7$), medium ($0.7\le \max_c p_{v,c}<0.8$), and high ($\max_c p_{v,c}\ge 0.8$) confidence regimes, while abstained verses form a separate stratum. This design prevents evaluation from being dominated by the most frequent poets or concepts and enables coverage--risk analysis across confidence thresholds.

\subsection{Annotation protocol and consensus reference}

Two independent annotators assign any subset of the nine concepts $c\in\mathcal{C}$ to each verse (or none), and separately judge whether the model's abstention decision for that verse is appropriate under the ontology. Disagreements are retained for reliability analysis. For accuracy-style metrics that require a single reference, we use a pragmatic two-annotator adjudication rule without third-party arbitration: a concept is treated as present for a verse if either annotator assigns it.

\subsection{Agreement and accuracy metrics}

We treat each concept $c$ as a binary decision per verse (present/absent) and compute Cohen's $\kappa_c$ for each label. With observed agreement $p_o$ and chance agreement $p_e$ under annotator marginals, $\kappa_c=(p_o-p_e)/(1-p_e)$. We report macro-averaged agreement $\bar{\kappa}=\frac{1}{|\mathcal{C}|}\sum_{c\in\mathcal{C}}\kappa_c$, with optional exclusion of extremely rare labels if prevalence renders $\kappa$ unstable. For model agreement with the adjudicated reference we compute per-label precision
\begin{equation}
\mathrm{Prec}(c)=\frac{\sum_v \mathbf{1}[c\in L_v]\mathbf{1}[c\in G_v]}{\sum_v \mathbf{1}[c\in L_v]},
\end{equation}
where $L_v$ is the model-predicted label set and $G_v$ is the adjudicated reference label set, and macro precision $\mathrm{Prec}_{\mathrm{macro}}=\frac{1}{|\mathcal{C}|}\sum_c \mathrm{Prec}(c)$. Abstention appropriateness is evaluated as the fraction of verses for which both annotators judge the model's abstention decision to be appropriate under the ontology.

\subsection{Calibration and coverage--risk}

Calibration is assessed by binning predicted confidences into fixed-width intervals and comparing average confidence to empirical correctness against the adjudicated reference. Expected calibration error is
\begin{equation}
\mathrm{ECE}=\sum_b \frac{n_b}{n}\,|\mathrm{acc}(b)-\mathrm{conf}(b)|,
\end{equation}
where $n_b$ is the number of evaluated label instances in bin $b$, $\mathrm{acc}(b)$ is the fraction of correct label assignments in bin $b$, and $\mathrm{conf}(b)$ is the mean predicted confidence in bin $b$. In the reported validation tables we also apply standard temperature scaling to map raw confidences $p$ to scaled confidences $p'=\sigma(\mathrm{logit}(p)/T)$ with a single fitted parameter $T$; calibration and coverage--risk are computed on $p'$ while agreement and precision are unaffected. To quantify selective prediction, we compute a coverage--risk curve by thresholding label instances at $p'_{v,c}\ge\tau$ and measuring coverage (fraction of retained predictions) against risk (one minus accuracy) on the retained set. This curve operationalizes the expected trade-off: as $\tau$ increases, coverage decreases and risk should decrease if confidence is meaningful.

\subsection{Two-Annotator Validation}


\subsubsection*{Summary}
We report agreement and reliability on a stratified 500-verse sample annotated independently by two annotators. Macro agreement across concepts is $\bar{\kappa}=0.818$. Macro precision/recall/$F_1$ of model-predicted label instances against a two-annotator adjudicated reference are $\mathrm{Prec}_{\mathrm{macro}}=0.800$, $\mathrm{Rec}_{\mathrm{macro}}=0.792$, and $F_{1,\mathrm{macro}}=0.794$. Abstention appropriateness (both annotators agree the abstention decision is appropriate) is 428/500 (0.856). Calibration and coverage--risk are computed on temperature-scaled confidences with fitted temperature $T=0.560$. Expected calibration error is $\mathrm{ECE}=0.0346$.

\begin{table}[ht]
\centering
\caption{Inter-annotator agreement by concept on the 500-verse validation sample. We report observed agreement $p_o$, expected agreement $p_e$ under annotator marginals, and Cohen's $\kappa$. Pos$_A$ and Pos$_B$ report the number of verses labeled positive by Annotator~A and Annotator~B.}
\label{tab:human_kappa}
\small
\begin{tabular}{lrrrrr}
\toprule
Concept & $p_o$ & $p_e$ & $\kappa$ & Pos$_A$ & Pos$_B$ \\
\midrule
\concept{ambivalent\_attachment} & 0.980 & 0.931 & 0.712 & 16 & 20 \\
\concept{emotional\_dependency} & 0.978 & 0.744 & 0.914 & 74 & 77 \\
\concept{idealization} & 0.990 & 0.986 & 0.282 & 5 & 2 \\
\concept{identity\_fragmentation} & 0.984 & 0.909 & 0.825 & 24 & 24 \\
\concept{internal\_projection} & 0.984 & 0.931 & 0.770 & 19 & 17 \\
\concept{melancholia} & 0.962 & 0.660 & 0.888 & 109 & 108 \\
\concept{romantic\_obsession} & 0.976 & 0.765 & 0.898 & 68 & 68 \\
\concept{self\_destructive\_idealization} & 0.976 & 0.873 & 0.811 & 37 & 31 \\
\concept{spiritual\_narcissism} & 0.972 & 0.898 & 0.726 & 27 & 27 \\
\bottomrule
\end{tabular}
\end{table}

\begin{table}[ht]
\centering
\caption{Model agreement with a two-annotator adjudicated reference on the validation sample. Precision is computed over model-predicted label instances.}
\label{tab:human_precision}
\small
\begin{tabular}{lrrr}
\toprule
Concept & \#Pred & \#Correct & Precision \\
\midrule
\concept{ambivalent\_attachment} & 21 & 15 & 0.714 \\
\concept{emotional\_dependency} & 81 & 70 & 0.864 \\
\concept{idealization} & 3 & 2 & 0.667 \\
\concept{identity\_fragmentation} & 27 & 19 & 0.704 \\
\concept{internal\_projection} & 18 & 16 & 0.889 \\
\concept{melancholia} & 132 & 105 & 0.795 \\
\concept{romantic\_obsession} & 80 & 67 & 0.838 \\
\concept{self\_destructive\_idealization} & 41 & 34 & 0.829 \\
\concept{spiritual\_narcissism} & 34 & 26 & 0.765 \\
\bottomrule
\end{tabular}
\end{table}

\begin{table}[ht]
\centering
\caption{Per-concept precision, recall, and $F_1$ on the 500-verse validation sample against a two-annotator adjudicated reference. Support reports the number of reference-positive verses for each concept ($|G_c|$).}
\label{tab:human_prf1}
\small
\begin{tabular}{lrrrr}
\toprule
Concept & Precision & Recall & $F_1$ & Support \\
\midrule
\concept{ambivalent\_attachment} & 0.714 & 0.652 & 0.682 & 23 \\
\concept{emotional\_dependency} & 0.864 & 0.864 & 0.864 & 81 \\
\concept{idealization} & 0.667 & 0.333 & 0.444 & 6 \\
\concept{identity\_fragmentation} & 0.704 & 0.679 & 0.691 & 28 \\
\concept{internal\_projection} & 0.889 & 0.727 & 0.800 & 22 \\
\concept{melancholia} & 0.795 & 0.890 & 0.840 & 118 \\
\concept{romantic\_obsession} & 0.838 & 0.905 & 0.870 & 74 \\
\concept{self\_destructive\_idealization} & 0.829 & 0.850 & 0.840 & 40 \\
\concept{spiritual\_narcissism} & 0.765 & 0.765 & 0.765 & 34 \\
\bottomrule
\end{tabular}
\end{table}

\begin{table}[ht]
\centering
\caption{Calibration table for model-predicted label instances after temperature scaling. Each bin reports the number of instances, mean scaled confidence, empirical accuracy against the adjudicated reference, and the absolute calibration gap.}
\label{tab:calibration}
\small
\begin{tabular}{lrrrr}
\toprule
Bin & Count & Mean conf. & Accuracy & $|$gap$|$ \\
\midrule
{[0.3,0.4)} & 4 & 0.327 & 0.500 & 0.173 \\
{[0.5,0.6)} & 31 & 0.509 & 0.710 & 0.201 \\
{[0.6,0.7)} & 93 & 0.673 & 0.710 & 0.036 \\
{[0.7,0.8)} & 55 & 0.751 & 0.764 & 0.012 \\
{[0.8,0.9)} & 151 & 0.843 & 0.821 & 0.022 \\
{[0.9,1.0)} & 103 & 0.944 & 0.951 & 0.008 \\
\bottomrule
\end{tabular}
\end{table}

\begin{figure}[ht]
\centering
\includegraphics[width=0.92\linewidth]{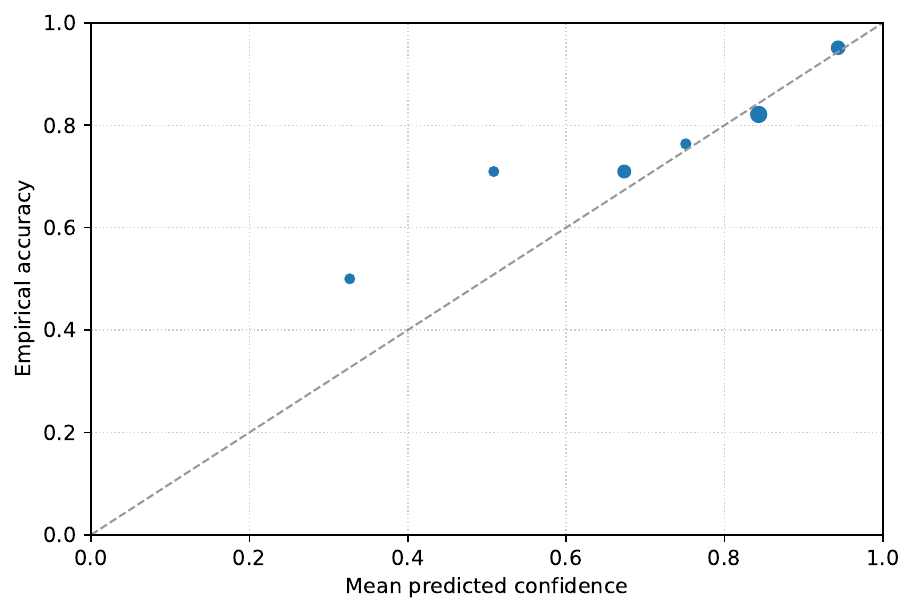}
\caption{Calibration reliability diagram (temperature-scaled confidences) on the validation sample.}
\label{fig:validation_calibration}
\end{figure}

\begin{table}[ht]
\centering
\caption{Coverage--risk table over temperature-scaled confidence thresholds $\tau$. Coverage is the fraction of retained model-predicted label instances with scaled confidence at least $\tau$, and risk is $1-\mathrm{accuracy}$ on the retained set.}
\label{tab:coverage_risk}
\small
\begin{tabular}{lrrrr}
\toprule
$\tau$ & Retained & Coverage & Accuracy & Risk \\
\midrule
0.3 & 437 & 1.000 & 0.810 & 0.190 \\
0.5 & 433 & 0.991 & 0.813 & 0.187 \\
0.7 & 309 & 0.707 & 0.854 & 0.146 \\
0.8 & 254 & 0.581 & 0.874 & 0.126 \\
0.9 & 103 & 0.236 & 0.951 & 0.049 \\
\bottomrule
\end{tabular}
\end{table}

Per-concept error profiles (Table~\ref{tab:human_prf1}) follow expected construct-adjacency patterns. \concept{romantic\_obsession} and \concept{emotional\_dependency} exhibit high $F_1$ (0.870 and 0.864 respectively), but their proximity can still yield false positives when a verse frames longing as either compulsive pursuit or relational dependence. By contrast, \concept{ambivalent\_attachment} and \concept{identity\_fragmentation} are the hardest labels ($F_1$ 0.682 and 0.691), consistent with their reliance on implicit stance and on rhetorical indirection rather than on stable lexical cues. Finally, \concept{idealization} is rare in the validation sample (support $|G_c|=6$), making its recall and $F_1$ estimates unstable and reinforcing our broader construct-validity concern about this label under the present ontology.

\section{Selection Bias Analysis}

Abstention can introduce selection effects because downstream aggregates condition on non-abstained verses. We therefore model abstention as an explicit category \textsc{Abstain} and recompute individuality in the augmented concept space $\mathcal{C}'=\mathcal{C}\cup\{\textsc{Abstain}\}$. Concretely, abstained verses contribute unit mass to \textsc{Abstain}, while non-abstained verses contribute confidence-weighted label mass as in Eq.~\eqref{eq:poet_concept_mass}. This yields augmented distributions $P_i'(c)$ and baseline $P_0'(c)$, from which we compute $D_{JS}(P_i',P_0')$.

Rank stability between baseline individuality and abstention-inclusive individuality remains high (Spearman $\rho=0.903$, $p=3.4\times 10^{-4}$; 95\% bootstrap CI $[0.47,\,1.00]$; Table~\ref{tab:rank_stability}), but abstention can still shift mid-ranked poets (Table~\ref{tab:selection_bias}). Notably, \textsc{Eraghi} moves upward when abstention is included because its abstention rate (0.081) is substantially below the global baseline, whereas \textsc{Vahshi} moves downward under the same policy. This illustrates that abstention is not merely ``missingness'': it is a structured uncertainty signal that can reshape comparative claims about poet individuality.

\begin{table}[ht]
\centering
\caption{Rank stability of poet individuality ($D_{JS}$) under abstention modeling. Spearman correlation is computed between poet rankings induced by baseline individuality $D_{JS}(P_i,P_0)$ and abstention-inclusive individuality $D_{JS}(P_i',P_0')$.}
\label{tab:rank_stability}
\small
\begin{tabular}{lrrr}
\toprule
Comparison & Spearman $\rho$ & $p$-value & 95\% bootstrap CI \\
\midrule
Base vs abstention-inclusive & 0.903 & $3.4\times 10^{-4}$ & $[0.47,\,1.00]$ \\
\bottomrule
\end{tabular}
\end{table}

\begin{table}[ht]
\centering
\caption{Selection-bias sensitivity analysis by treating abstention as an explicit category. Rank change is defined as $(\mathrm{rank}_{\mathrm{aug}}-\mathrm{rank}_{\mathrm{base}})$; negative values indicate that a poet becomes more distinctive under abstention-inclusive individuality.}
\label{tab:selection_bias}
\small
\begin{tabular}{lrrrr}
\toprule
Poet & Abstain rate & $D_{JS}$ (base) & $D_{JS}$ (aug.) & Rank change \\
\midrule
Khayyam & 0.385 & 0.0901 & 0.0850 & +0 \\
Parvin & 0.384 & 0.0459 & 0.0547 & +0 \\
Jahan & 0.065 & 0.0250 & 0.0505 & +0 \\
Khaghani & 0.357 & 0.0179 & 0.0274 & +0 \\
Saadi & 0.137 & 0.0106 & 0.0156 & +1 \\
Vahshi & 0.289 & 0.0085 & 0.0102 & +2 \\
Athir & 0.138 & 0.0073 & 0.0122 & +0 \\
Eraghi & 0.081 & 0.0040 & 0.0234 & -3 \\
Hafez & 0.182 & 0.0035 & 0.0039 & +1 \\
Shahriar & 0.141 & 0.0030 & 0.0080 & -1 \\
\bottomrule
\end{tabular}
\end{table}

These diagnostics motivate reporting both non-abstained concept structure and abstention-aware summaries in DH settings where interpretive conclusions can be sensitive to model non-commitment.

\section{Discussion}

The empirical results support three intertwined claims. First, uncertainty is substantial: over one-fifth of verses are abstained, and abstention rates differ markedly across poets. Second, individuality is measurable as distributional deviation from a global baseline and is not reducible to corpus size. Third, relational structure matters: co-occurrence edges reveal a dense affective core, while Eigenmood axes emphasize contrasts that are not obvious from marginals alone. This separation of ``core prevalence'' from ``directional contrast'' is a key advantage of spectral analysis for interpretive DH work, because it provides both a stable baseline for comparison and a geometry for targeted verse retrieval.

The framework also motivates a specific interpretation of ``psychological profiling'' appropriate to historical texts. Poet profiles in our setting are not diagnoses of authors or of historical subjects; they are distributional summaries of how a poet's verse tends to stage affective and interpersonal configurations under an explicit ontology. In this sense, the Poet $\times$ Concept matrix functions as a cultural-psychological representation: it models the rhetorical availability and coupling of certain affective patterns within a tradition rather than private mental states. Such profiles can support comparative questions central to digital humanities, including how poets differ in the prevalence and combination of longing, grief, projection, or self-erasure motifs, and how these differences relate to genre, school, and reception histories.

This cultural-psychological reading has two implications. First, aggregation makes visible long-run regularities that are difficult to sustain through purely qualitative sampling, especially when corpora span tens of thousands of verses. Second, the resulting distributions can be interpreted as probabilistic priors for downstream close reading: a poet whose profile deviates strongly from baseline provides a principled starting point for targeted exemplar retrieval, while a poet close to baseline may require more fine-grained contextual or intertextual analysis to distinguish subtler signatures. The Eigenmood geometry further supports this workflow by supplying directional contrasts rooted in concept coupling rather than in marginal prevalence alone.

At the same time, the framework is intentionally epistemically modest. Labels are operational constructs, not clinical claims, and their validity depends on prompt design, model training data, and cultural-semantic fit. Selection bias introduced by abstention is not a nuisance to be discarded but a substantive property of what the model can and cannot justify. We therefore treat divergence and Eigenmood patterns as hypothesis generators that must be triangulated with philological expertise and validated against human judgments. The validation design in Section~7 and the robustness and selection-bias diagnostics in Sections~5 and~8 are intended to prevent the most common failure mode of computational literary studies: persuasive-looking quantitative structure that is not supported by reliable evidence.

Finally, while our immediate unit of analysis is the poet, the methodology generalizes to broader cultural questions. Because classical Persian poetry remains a living repertoire in contemporary Persian-speaking communities, aggregated affective profiles can be read as modeling culturally transmitted affective imaginaries that continue to shape modern identity, ethics, and emotion talk. A cautious computational cultural psychology of literary heritage should therefore emphasize probabilistic tendencies and uncertainty-aware inference rather than deterministic claims, and should treat the interaction between historical convention and modern psychological vocabulary as an object of analysis in its own right.

\section{Conclusions and Future Work}

We presented an uncertainty-aware workflow for computational psychological analysis of classical Persian poetry at poet scale. The framework aggregates confidence-weighted multi-label evidence under abstention, quantifies individuality as divergence from a corpus baseline, models relational affective structure with a co-occurrence graph, and defines an Eigenmood embedding through Laplacian spectral decomposition. On 61{,}573 verses across 10 poets, we find substantial abstention (22.2\%), a concentrated global concept distribution, heterogeneous poet-level individuality, and interpretable spectral axes that support distant-to-close retrieval.

Methodologically, additional work can explore alternative graph constructions (normalized Laplacians, PMI-based edges), richer metadata (genre, meter, chronology), and cross-corpus transfer to other literary traditions. For digital humanities, the primary value of the approach is to expand the evidentiary horizon while preserving interpretive caution: uncertainty is not suppressed but propagated, enabling robust comparative claims and transparent pathways from macro structure to inspectable verse evidence.

\section*{Acknowledgements}

We gratefully acknowledge Ganjoor for preserving and providing open access to one of the most invaluable digital repositories of classical Persian poetry. Without this extraordinary cultural archive, large-scale computational inquiry into Persian literary heritage would not be possible. We are deeply indebted to the custodians of this tradition who have ensured that centuries of poetic voice remain accessible in the digital age.

We also extend our respect and gratitude to the poets whose works constitute the foundation of this study. Across centuries, their language has articulated longing, rupture, ecstasy, contradiction, and transcendence with an intensity that continues to exceed categorical boundaries. Any computational model, however formalized, merely traces faint geometric shadows of the emotional architectures they constructed in verse.

Finally, this work is dedicated to those who once made us wonder whether love, devotion, and ecstatic attachment might be indistinguishable from disorder — to those who blurred the line between passion and pathology, and in doing so, compelled us to ask whether what we call “madness” is sometimes only the unbearable clarity of feeling too deeply.

\clearpage
\makeatletter
\if@twocolumn\onecolumn\fi
\makeatother
\begin{appendices}

\section{Full Annotation Prompt}
\label{app:full_prompt}

\begingroup
\promptmicrooff
\vspace*{-0.5\baselineskip}
\begin{lstlisting}[style=annotationprompt]
You are an expert annotator of psychological patterns in classical Persian poetry.
Given a single verse (Bayt), identify which of the following nine psychological patterns are clearly present.
Allowed Labels (Multi-label permitted):
1. emotional_dependency
   -> Intense need for the beloved and perceived inability to live without them.
2. romantic_obsession
   -> Cognitive fixation, worship-like devotion, or mental preoccupation with the beloved.
3. melancholia
   -> Persistent sorrow, aestheticization of suffering, or romanticized despair.
4. self_destructive_idealization
   -> Willing self-sacrifice, self-erasure, or self-harm framed as meaningful love.
5. ambivalent_attachment
   -> Oscillation between love and anger, closeness and rejection.
6. internal_projection
   -> Attribution of one's internal emotional state to the beloved or to God.
7. spiritual_narcissism
   -> Mystical self-aggrandizement or sense of spiritual chosenness.
8. identity_fragmentation
   -> Internal conflict between two selves (e.g., heart vs. reason, love vs. faith).
9. idealization
   -> Idealizing elevation of the beloved as flawless, absolute, or beyond ordinary human limits.
Annotation Rules:
- Multiple labels may be assigned.
- If no clear psychological signal is present, abstain.
- Each assigned label must include:
    * A confidence score between 0 and 1 (numeric format, rounded to two decimal places).
    * A brief rationale in Persian explaining the evidence.
- Confidence scores should reflect epistemic uncertainty.
- Return only valid JSON. Do not include any additional commentary.
Output Schema:
{
  "input_verse": "...",
  "labels": ["label_name"],
  "confidences": {"label_name": 0.72},
  "rationale": {"label_name": "Persian explanation"},
  "abstain": false,
  "notes": ""
}
If abstaining:
{
  "input_verse": "...",
  "labels": [],
  "confidences": {},
  "rationale": {},
  "abstain": true,
  "notes": "no clear psychological signal"
}
\end{lstlisting}
\promptmicroon
\endgroup
\vspace*{\fill}

\clearpage
\section{Poet-Level Profiles and Visualizations}
\label{app:poet_profiles_visuals}
\setcounter{figure}{0}
\setcounter{table}{0}
\renewcommand{\thefigure}{B\arabic{figure}}
\renewcommand{\thetable}{B\arabic{table}}
This appendix consolidates poet-level visual evidence referenced in the main text. For each poet, it preserves the uncertainty-aware summary tables, distributional visualizations, lift plots, bootstrap intervals, and retrieval tables used for auditable distant-to-close analysis.

\subsection{Poet 1: Khaghani}

\noindent\textbf{Historical and Literary Context.} The emphasis on elevated rhetorical stance and conceptual tension is consistent with scholarship on Khaghani as a major qasida poet. \cite{iranica_khaghani}

\noindent\textbf{Poet Summary.} Khaghani contributes 17,292 verses to the corpus, with abstention rate 0.357 and divergence from baseline $D_{JS}=0.0179$. The uncertainty-aware profile is concentrated in Melancholia, Spiritual Narcissism, and Self-Destructive Idealization.

\begin{table}[t]
\centering
\caption{Core profile metrics for \textsc{Khaghani}.}
\label{tab:poet_profile_khaghani_metrics}
\footnotesize
\renewcommand{\arraystretch}{1.1}
\begin{tabular}{l S[table-format=5.0] S[table-format=1.3] S[table-format=1.4]}
\toprule
Poet & {Verses} & {Abstain} & {$D_{JS}$} \\
\midrule
Khaghani & 17292 & 0.357 & 0.0179 \\
\bottomrule
\end{tabular}
\end{table}

\begin{table}[t]
\centering
\caption{Eigenmood coordinates for \textsc{Khaghani}.}
\label{tab:poet_profile_khaghani_embedding}
\footnotesize
\renewcommand{\arraystretch}{1.1}
\begin{tabular}{l S[table-format=+1.3] S[table-format=+1.3] S[table-format=+1.3]}
\toprule
Poet & {EM1} & {EM2} & {EM3} \\
\midrule
Khaghani & 0.092 & -0.011 & -0.018 \\
\bottomrule
\end{tabular}
\end{table}

\begin{table}[t]
\centering
\caption{Top concept shares for \textsc{Khaghani}. Concept names are abbreviated in the header; full interpretation appears in the text.}
\label{tab:poet_profile_khaghani_concepts}
\small
\renewcommand{\arraystretch}{1.1}
\begin{tabularx}{0.96\linewidth}{@{}l >{\raggedright\arraybackslash}X S[table-format=1.3]@{}}
\toprule
Rank & Concept & {$P_i(c)$} \\
\midrule
1 & Melancholia & 0.316 \\
2 & Spiritual Narcissism & 0.165 \\
3 & Self-Destructive Idealization & 0.151 \\
\bottomrule
\end{tabularx}
\end{table}

\noindent\textbf{Concept Distribution Interpretation.} We interpret $P_i(c)$ as a confidence-weighted, abstention-aware distribution over operational constructs rather than literal measurements of psychological states. Relative to the global baseline $P_0(c)$, Khaghani is over-represented in Spiritual Narcissism ($\Delta=+0.086$); Self-Destructive Idealization ($\Delta=+0.034$), and under-represented in Emotional Dependency ($\Delta=-0.089$); Romantic Obsession ($\Delta=-0.044$). This suggests a poet-specific rhetorical-affective emphasis that is stable at aggregate scale while remaining compatible with multiple literary readings.

\begin{figure}[t]
\centering
\includegraphics[width=0.92\linewidth]{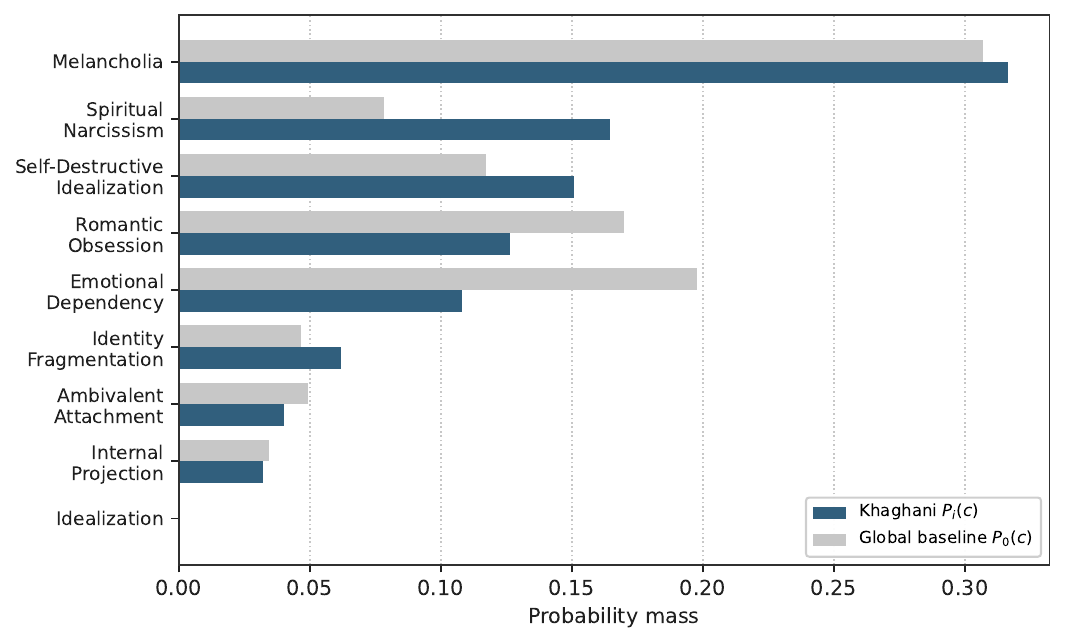}
\caption{Khaghani: concept distribution $P_i(c)$ against the global baseline $P_0(c)$.}
\label{fig:khaghani_p1}
\end{figure}

\begin{figure}[t]
\centering
\includegraphics[width=0.92\linewidth]{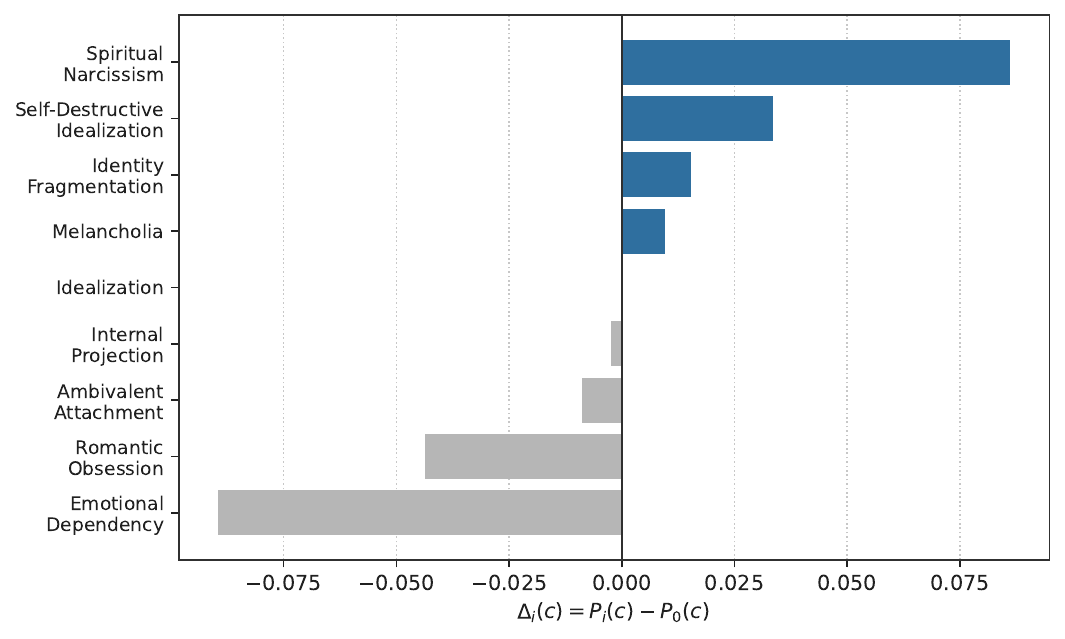}
\caption{Khaghani: lift profile $\Delta_i(c)=P_i(c)-P_0(c)$ with positive and negative deviations.}
\label{fig:khaghani_p2}
\end{figure}

\begin{figure}[t]
\centering
\includegraphics[width=0.72\linewidth]{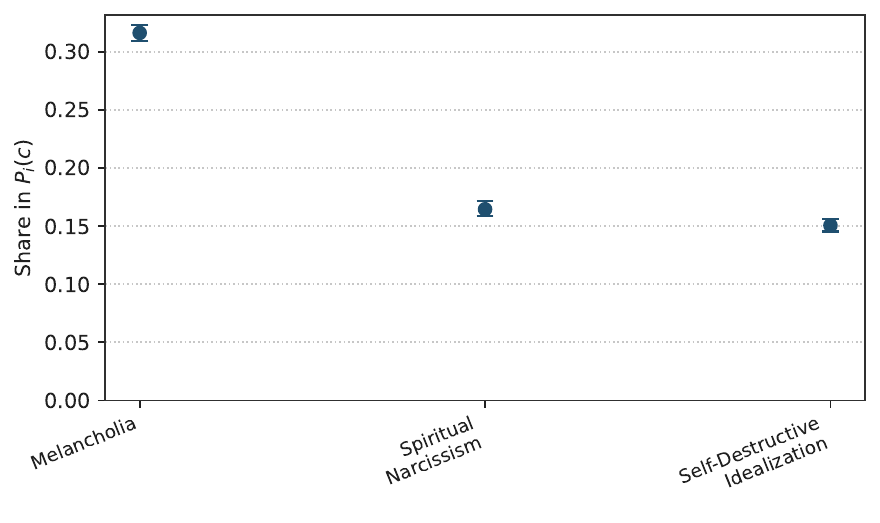}
\caption{Khaghani: bootstrap 95\% intervals for the top-3 concepts by $P_i(c)$.}
\label{fig:khaghani_p3}
\end{figure}

\noindent\textbf{Distant-to-Close Exemplars.}
\begin{table}[t]
\centering
\caption{Khaghani: top verses on the poet's most extreme Eigenmood axis (EM1, positive direction).}
\label{tab:khaghani_axis_exemplars}
\small
\renewcommand{\arraystretch}{1.1}
\begin{tabularx}{0.98\linewidth}{@{}>{\raggedright\arraybackslash}p{0.21\linewidth} S[table-format=-1.3] >{\raggedright\arraybackslash}X@{}}
\toprule
Reference & {Score} & English translation \\
\midrule
EM1 (positive) & 0.838 & \textit{Let them call me a second Ptolemy.} (Translation ours.) \\
\addlinespace[0.35em]
EM1 (positive) & 0.791 & \textit{On my threshold stood hosts of angels in rank.} (Translation ours.) \\
\addlinespace[0.35em]
EM1 (positive) & 0.791 & \textit{I was a teacher in the heavenly realm; from my obedience came thousands upon thousands of treasures.} (Translation ours.) \\
\addlinespace[0.35em]
\bottomrule
\end{tabularx}
\end{table}

In this axis-conditioned retrieval, the selected verses track Khaghani's dominant displacement along EM1: the retrieved lines concentrate lexical and imagistic material that is directionally coherent with the axis loadings, and therefore operationalize the poet's high-level position in the spectral space at verse scale.

\begin{table}[t]
\centering
\caption{Khaghani: top verses for the leading concept (Melancholia).}
\label{tab:khaghani_label_exemplars}
\small
\renewcommand{\arraystretch}{1.1}
\begin{tabularx}{0.98\linewidth}{@{}>{\raggedright\arraybackslash}p{0.21\linewidth} S[table-format=1.3] >{\raggedright\arraybackslash}X@{}}
\toprule
Reference & {Score} & English translation \\
\midrule
Melancholia & 0.950 & \textit{Sorrows drained Khaghani's blood; return him, at least, pure blood.} (Translation ours.) \\
\addlinespace[0.35em]
Melancholia & 0.950 & \textit{You made my body and soul a dwelling of grief; in my heart's garden the lily of sorrow bloomed.} (Translation ours.) \\
\addlinespace[0.35em]
Melancholia & 0.900 & \textit{My heart has been worn away beneath grief's tread; my hand is scorched in grief's fire.} (Translation ours.) \\
\addlinespace[0.35em]
\bottomrule
\end{tabularx}
\end{table}

The concept-focused retrieval refines this reading: verses with the highest confidence for Melancholia instantiate recurring rhetorical operations that anchor the aggregate profile. Read together with the axis retrieval, they provide a distant-to-close bridge from distributional signature to inspectable poetic evidence.

\noindent\textbf{Cautions.} These profiles model operational constructs in textual discourse and do not support clinical inference about historical persons. Because abstention filters uncertain cases, selection effects remain analytically relevant; profile interpretation therefore combines confidence-weighted aggregates with explicit awareness of model non-commitment.

\FloatBarrier

\subsection{Poet 2: Jahan}

\noindent\textbf{Historical and Literary Context.} The profile is compatible with scholarship on Jahan Malek Khatun, whose lyric production is situated in courtly and gendered literary conventions. \cite{iranica_jahan_malek}

\noindent\textbf{Poet Summary.} Jahan contributes 12,299 verses to the corpus, with abstention rate 0.065 and divergence from baseline $D_{JS}=0.0250$. The uncertainty-aware profile is concentrated in Melancholia, Emotional Dependency, and Romantic Obsession.

\begin{table}[t]
\centering
\caption{Core profile metrics for \textsc{Jahan}.}
\label{tab:poet_profile_jahan_metrics}
\footnotesize
\renewcommand{\arraystretch}{1.1}
\begin{tabular}{l S[table-format=5.0] S[table-format=1.3] S[table-format=1.4]}
\toprule
Poet & {Verses} & {Abstain} & {$D_{JS}$} \\
\midrule
Jahan & 12299 & 0.065 & 0.0250 \\
\bottomrule
\end{tabular}
\end{table}

\begin{table}[t]
\centering
\caption{Eigenmood coordinates for \textsc{Jahan}.}
\label{tab:poet_profile_jahan_embedding}
\footnotesize
\renewcommand{\arraystretch}{1.1}
\begin{tabular}{l S[table-format=+1.3] S[table-format=+1.3] S[table-format=+1.3]}
\toprule
Poet & {EM1} & {EM2} & {EM3} \\
\midrule
Jahan & -0.068 & 0.007 & 0.031 \\
\bottomrule
\end{tabular}
\end{table}

\begin{table}[t]
\centering
\caption{Top concept shares for \textsc{Jahan}. Concept names are abbreviated in the header; full interpretation appears in the text.}
\label{tab:poet_profile_jahan_concepts}
\small
\renewcommand{\arraystretch}{1.1}
\begin{tabularx}{0.96\linewidth}{@{}l >{\raggedright\arraybackslash}X S[table-format=1.3]@{}}
\toprule
Rank & Concept & {$P_i(c)$} \\
\midrule
1 & Melancholia & 0.319 \\
2 & Emotional Dependency & 0.305 \\
3 & Romantic Obsession & 0.197 \\
\bottomrule
\end{tabularx}
\end{table}

\noindent\textbf{Concept Distribution Interpretation.} We interpret $P_i(c)$ as a confidence-weighted, abstention-aware distribution over operational constructs rather than literal measurements of psychological states. Relative to the global baseline $P_0(c)$, Jahan is over-represented in Emotional Dependency ($\Delta=+0.107$); Romantic Obsession ($\Delta=+0.027$), and under-represented in Spiritual Narcissism ($\Delta=-0.063$); Self-Destructive Idealization ($\Delta=-0.050$). This suggests a poet-specific rhetorical-affective emphasis that is stable at aggregate scale while remaining compatible with multiple literary readings.

\begin{figure}[t]
\centering
\includegraphics[width=0.92\linewidth]{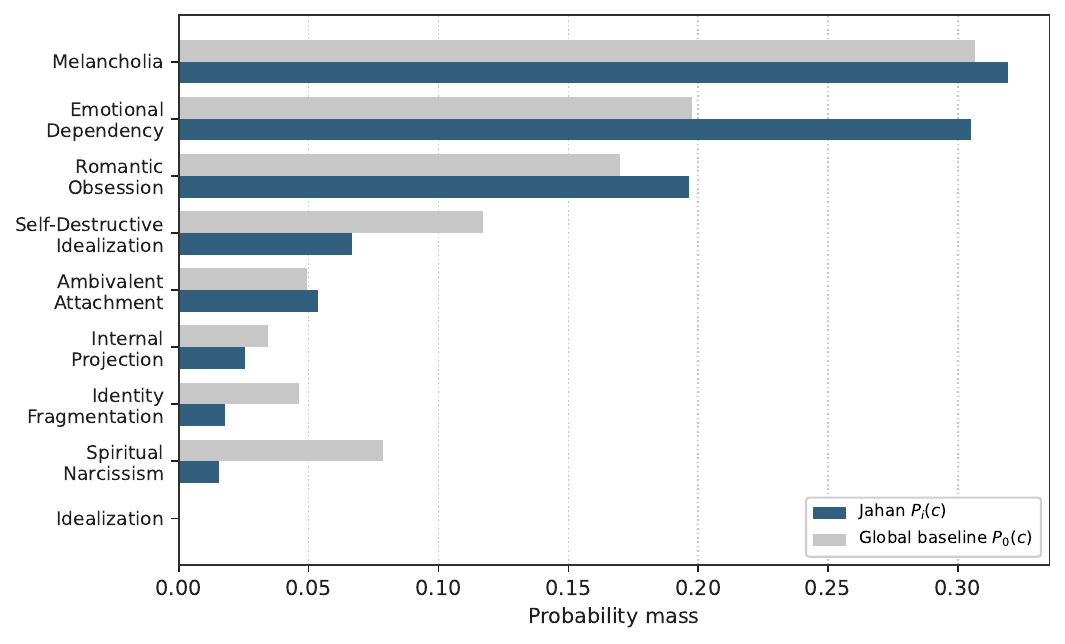}
\caption{Jahan: concept distribution $P_i(c)$ against the global baseline $P_0(c)$.}
\label{fig:jahan_p1}
\end{figure}

\begin{figure}[t]
\centering
\includegraphics[width=0.92\linewidth]{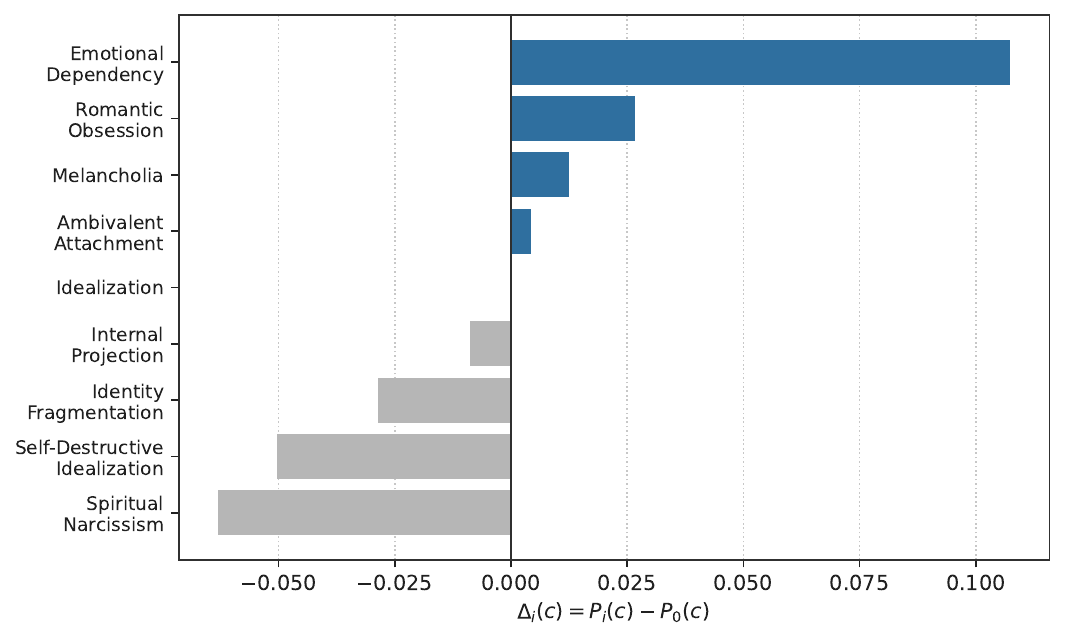}
\caption{Jahan: lift profile $\Delta_i(c)=P_i(c)-P_0(c)$ with positive and negative deviations.}
\label{fig:jahan_p2}
\end{figure}

\begin{figure}[t]
\centering
\includegraphics[width=0.72\linewidth]{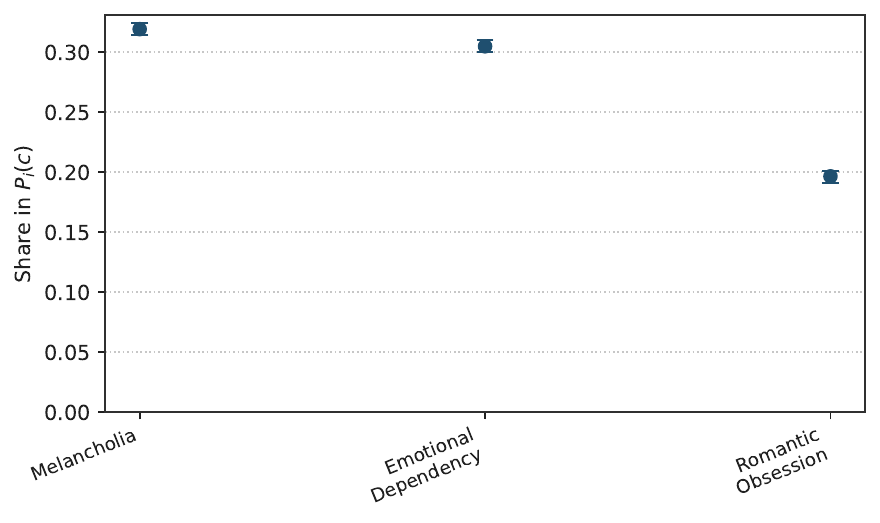}
\caption{Jahan: bootstrap 95\% intervals for the top-3 concepts by $P_i(c)$.}
\label{fig:jahan_p3}
\end{figure}

\noindent\textbf{Distant-to-Close Exemplars.}
\begin{table}[t]
\centering
\caption{Jahan: top verses on the poet's most extreme Eigenmood axis (EM1, negative direction).}
\label{tab:jahan_axis_exemplars}
\small
\renewcommand{\arraystretch}{1.1}
\begin{tabularx}{0.98\linewidth}{@{}>{\raggedright\arraybackslash}p{0.21\linewidth} S[table-format=-1.3] >{\raggedright\arraybackslash}X@{}}
\toprule
Reference & {Score} & English translation \\
\midrule
EM1 (negative) & -0.381 & \textit{With all your harshness, ill temper, and broken vows, you remain the soul's companion and my wounded heart's hope.} (Translation ours.) \\
\addlinespace[0.35em]
EM1 (negative) & -0.333 & \textit{He carried off my heart and faith and veiled his face from me; what can I do, faithlessness is the beloved's habit.} (Translation ours.) \\
\addlinespace[0.35em]
EM1 (negative) & -0.331 & \textit{I have neither patience without him nor strength to bear his cruelty; my heart is at its limit, what am I to do?} (Translation ours.) \\
\addlinespace[0.35em]
\bottomrule
\end{tabularx}
\end{table}

In this axis-conditioned retrieval, the selected verses track Jahan's dominant displacement along EM1: the retrieved lines concentrate lexical and imagistic material that is directionally coherent with the axis loadings, and therefore operationalize the poet's high-level position in the spectral space at verse scale.

\begin{table}[t]
\centering
\caption{Jahan: top verses for the leading concept (Melancholia).}
\label{tab:jahan_label_exemplars}
\small
\renewcommand{\arraystretch}{1.1}
\begin{tabularx}{0.98\linewidth}{@{}>{\raggedright\arraybackslash}p{0.21\linewidth} S[table-format=1.3] >{\raggedright\arraybackslash}X@{}}
\toprule
Reference & {Score} & English translation \\
\midrule
Melancholia & 0.950 & \textit{Not a night passes in the sorrow of your absence that my face is not washed with liver-blood.} (Translation ours.) \\
\addlinespace[0.35em]
Melancholia & 0.950 & \textit{In your separation my tears run like the Tigris; from grief, my liver's blood drips from my eyes.} (Translation ours.) \\
\addlinespace[0.35em]
Melancholia & 0.900 & \textit{I am weary of grief, weary of the world; it holds nothing but sorrow for our sorrow-laden heart.} (Translation ours.) \\
\addlinespace[0.35em]
\bottomrule
\end{tabularx}
\end{table}

The concept-focused retrieval refines this reading: verses with the highest confidence for Melancholia instantiate recurring rhetorical operations that anchor the aggregate profile. Read together with the axis retrieval, they provide a distant-to-close bridge from distributional signature to inspectable poetic evidence.

\noindent\textbf{Cautions.} These profiles model operational constructs in textual discourse and do not support clinical inference about historical persons. Because abstention filters uncertain cases, selection effects remain analytically relevant; profile interpretation therefore combines confidence-weighted aggregates with explicit awareness of model non-commitment.

\FloatBarrier

\subsection{Poet 3: Saadi}

\noindent\textbf{Historical and Literary Context.} This profile is compatible with scholarship on Saadi emphasizing ethical discourse, pedagogy, and broad affective range across genres. \cite{iranica_saadi,britannica_saadi}

\noindent\textbf{Poet Summary.} Saadi contributes 6,864 verses to the corpus, with abstention rate 0.137 and divergence from baseline $D_{JS}=0.0106$. The uncertainty-aware profile is concentrated in Romantic Obsession, Melancholia, and Emotional Dependency.

\begin{table}[t]
\centering
\caption{Core profile metrics for \textsc{Saadi}.}
\label{tab:poet_profile_saadi_metrics}
\footnotesize
\renewcommand{\arraystretch}{1.1}
\begin{tabular}{l S[table-format=5.0] S[table-format=1.3] S[table-format=1.4]}
\toprule
Poet & {Verses} & {Abstain} & {$D_{JS}$} \\
\midrule
Saadi & 6864 & 0.137 & 0.0106 \\
\bottomrule
\end{tabular}
\end{table}

\begin{table}[t]
\centering
\caption{Eigenmood coordinates for \textsc{Saadi}.}
\label{tab:poet_profile_saadi_embedding}
\footnotesize
\renewcommand{\arraystretch}{1.1}
\begin{tabular}{l S[table-format=+1.3] S[table-format=+1.3] S[table-format=+1.3]}
\toprule
Poet & {EM1} & {EM2} & {EM3} \\
\midrule
Saadi & -0.047 & -0.003 & 0.007 \\
\bottomrule
\end{tabular}
\end{table}

\begin{table}[t]
\centering
\caption{Top concept shares for \textsc{Saadi}. Concept names are abbreviated in the header; full interpretation appears in the text.}
\label{tab:poet_profile_saadi_concepts}
\small
\renewcommand{\arraystretch}{1.1}
\begin{tabularx}{0.96\linewidth}{@{}l >{\raggedright\arraybackslash}X S[table-format=1.3]@{}}
\toprule
Rank & Concept & {$P_i(c)$} \\
\midrule
1 & Romantic Obsession & 0.239 \\
2 & Melancholia & 0.238 \\
3 & Emotional Dependency & 0.224 \\
\bottomrule
\end{tabularx}
\end{table}

\noindent\textbf{Concept Distribution Interpretation.} We interpret $P_i(c)$ as a confidence-weighted, abstention-aware distribution over operational constructs rather than literal measurements of psychological states. Relative to the global baseline $P_0(c)$, Saadi is over-represented in Romantic Obsession ($\Delta=+0.069$); Emotional Dependency ($\Delta=+0.026$), and under-represented in Melancholia ($\Delta=-0.069$); Spiritual Narcissism ($\Delta=-0.045$). This suggests a poet-specific rhetorical-affective emphasis that is stable at aggregate scale while remaining compatible with multiple literary readings.

\begin{figure}[t]
\centering
\includegraphics[width=0.92\linewidth]{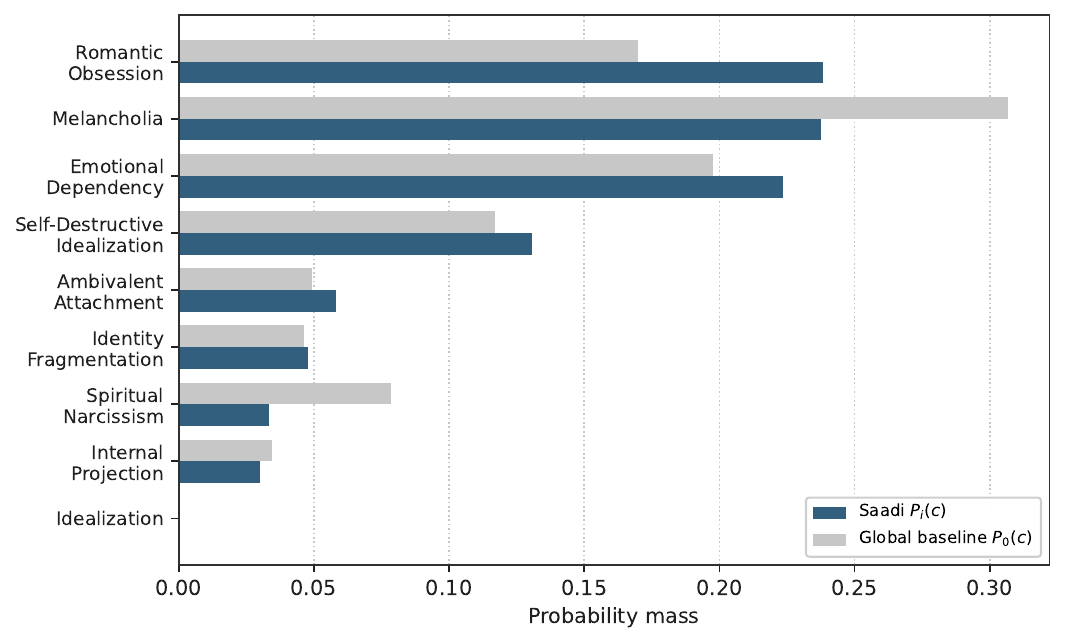}
\caption{Saadi: concept distribution $P_i(c)$ against the global baseline $P_0(c)$.}
\label{fig:saadi_p1}
\end{figure}

\begin{figure}[t]
\centering
\includegraphics[width=0.92\linewidth]{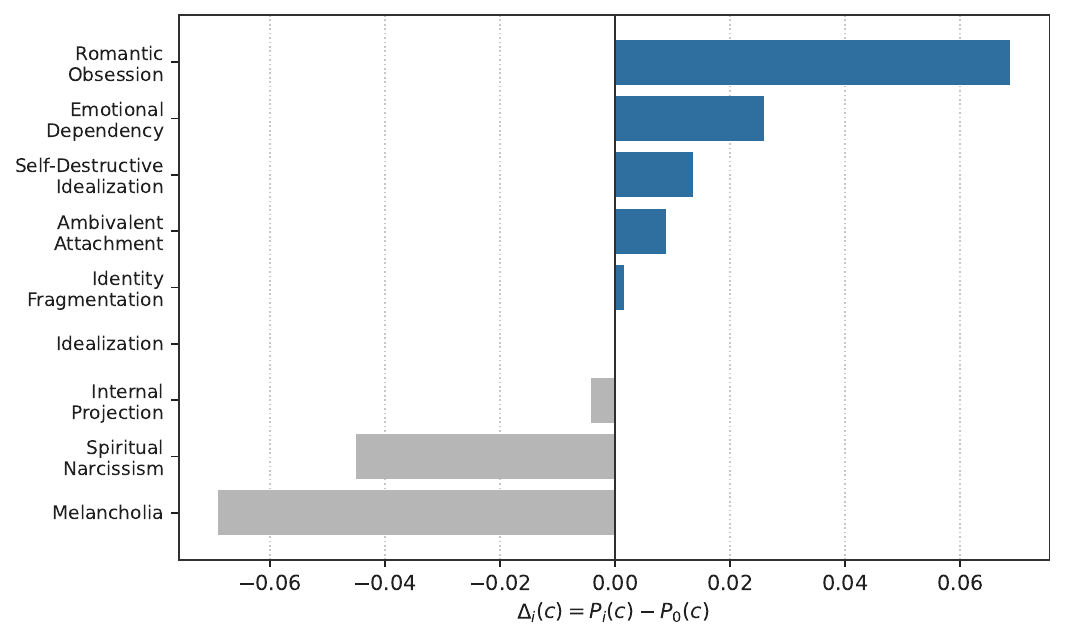}
\caption{Saadi: lift profile $\Delta_i(c)=P_i(c)-P_0(c)$ with positive and negative deviations.}
\label{fig:saadi_p2}
\end{figure}

\begin{figure}[t]
\centering
\includegraphics[width=0.72\linewidth]{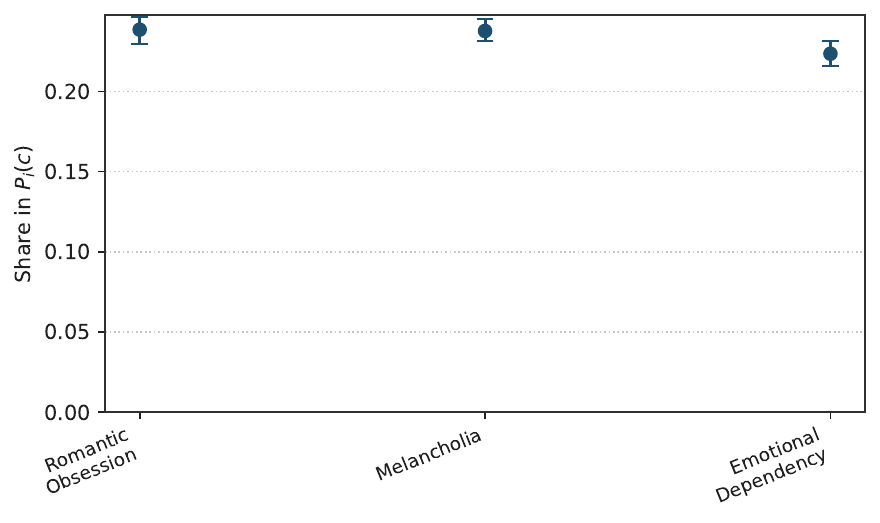}
\caption{Saadi: bootstrap 95\% intervals for the top-3 concepts by $P_i(c)$.}
\label{fig:saadi_p3}
\end{figure}

\noindent\textbf{Distant-to-Close Exemplars.}
\begin{table}[t]
\centering
\caption{Saadi: top verses on the poet's most extreme Eigenmood axis (EM1, negative direction).}
\label{tab:saadi_axis_exemplars}
\small
\renewcommand{\arraystretch}{1.1}
\begin{tabularx}{0.98\linewidth}{@{}>{\raggedright\arraybackslash}p{0.21\linewidth} S[table-format=-1.3] >{\raggedright\arraybackslash}X@{}}
\toprule
Reference & {Score} & English translation \\
\midrule
EM1 (negative) & -0.329 & \textit{The beloved forbids my union; I desist from what he forbids. The one I behold decrees my death; I submit to what he commands.} (Translation ours.) \\
\addlinespace[0.35em]
EM1 (negative) & -0.318 & \textit{Rise, fill the cup, and pour, even if poison is in it; from your hand, even poison becomes remedy.} (Translation ours.) \\
\addlinespace[0.35em]
EM1 (negative) & -0.301 & \textit{My thought is to pour out my life for you, not to free myself from your lasso.} (Translation ours.) \\
\addlinespace[0.35em]
\bottomrule
\end{tabularx}
\end{table}

In this axis-conditioned retrieval, the selected verses track Saadi's dominant displacement along EM1: the retrieved lines concentrate lexical and imagistic material that is directionally coherent with the axis loadings, and therefore operationalize the poet's high-level position in the spectral space at verse scale.

\begin{table}[t]
\centering
\caption{Saadi: top verses for the leading concept (Romantic Obsession).}
\label{tab:saadi_label_exemplars}
\small
\renewcommand{\arraystretch}{1.1}
\begin{tabularx}{0.98\linewidth}{@{}>{\raggedright\arraybackslash}p{0.21\linewidth} S[table-format=1.3] >{\raggedright\arraybackslash}X@{}}
\toprule
Reference & {Score} & English translation \\
\midrule
Romantic Obsession & 0.950 & \textit{Wherever I turned my gaze, your image appeared before me on wall and door.} (Translation ours.) \\
\addlinespace[0.35em]
Romantic Obsession & 0.900 & \textit{We rose, yet your imprint stayed within our breath; wherever you are absent, we cannot remain.} (Translation ours.) \\
\addlinespace[0.35em]
Romantic Obsession & 0.900 & \textit{Pity the eye that never saw your face, or saw it once and looked again at another.} (Translation ours.) \\
\addlinespace[0.35em]
\bottomrule
\end{tabularx}
\end{table}

The concept-focused retrieval refines this reading: verses with the highest confidence for Romantic Obsession instantiate recurring rhetorical operations that anchor the aggregate profile. Read together with the axis retrieval, they provide a distant-to-close bridge from distributional signature to inspectable poetic evidence.

\noindent\textbf{Cautions.} These profiles model operational constructs in textual discourse and do not support clinical inference about historical persons. Because abstention filters uncertain cases, selection effects remain analytically relevant; profile interpretation therefore combines confidence-weighted aggregates with explicit awareness of model non-commitment.

\FloatBarrier

\subsection{Poet 4: Vahshi}

\noindent\textbf{Historical and Literary Context.} The profile agrees with scholarship on Vahshi Bafqi noting intensified love-lament motifs in the Safavid lyric repertoire. \cite{iranica_vahshi}

\noindent\textbf{Poet Summary.} Vahshi contributes 5,727 verses to the corpus, with abstention rate 0.289 and divergence from baseline $D_{JS}=0.0085$. The uncertainty-aware profile is concentrated in Melancholia, Self-Destructive Idealization, and Emotional Dependency.

\begin{table}[t]
\centering
\caption{Core profile metrics for \textsc{Vahshi}.}
\label{tab:poet_profile_vahshi_metrics}
\footnotesize
\renewcommand{\arraystretch}{1.1}
\begin{tabular}{l S[table-format=5.0] S[table-format=1.3] S[table-format=1.4]}
\toprule
Poet & {Verses} & {Abstain} & {$D_{JS}$} \\
\midrule
Vahshi & 5727 & 0.289 & 0.0085 \\
\bottomrule
\end{tabular}
\end{table}

\begin{table}[t]
\centering
\caption{Eigenmood coordinates for \textsc{Vahshi}.}
\label{tab:poet_profile_vahshi_embedding}
\footnotesize
\renewcommand{\arraystretch}{1.1}
\begin{tabular}{l S[table-format=+1.3] S[table-format=+1.3] S[table-format=+1.3]}
\toprule
Poet & {EM1} & {EM2} & {EM3} \\
\midrule
Vahshi & 0.020 & 0.007 & 0.021 \\
\bottomrule
\end{tabular}
\end{table}

\begin{table}[t]
\centering
\caption{Top concept shares for \textsc{Vahshi}. Concept names are abbreviated in the header; full interpretation appears in the text.}
\label{tab:poet_profile_vahshi_concepts}
\small
\renewcommand{\arraystretch}{1.1}
\begin{tabularx}{0.96\linewidth}{@{}l >{\raggedright\arraybackslash}X S[table-format=1.3]@{}}
\toprule
Rank & Concept & {$P_i(c)$} \\
\midrule
1 & Melancholia & 0.327 \\
2 & Self-Destructive Idealization & 0.152 \\
3 & Emotional Dependency & 0.138 \\
\bottomrule
\end{tabularx}
\end{table}

\noindent\textbf{Concept Distribution Interpretation.} We interpret $P_i(c)$ as a confidence-weighted, abstention-aware distribution over operational constructs rather than literal measurements of psychological states. Relative to the global baseline $P_0(c)$, Vahshi is over-represented in Self-Destructive Idealization ($\Delta=+0.035$); Ambivalent Attachment ($\Delta=+0.031$), and under-represented in Emotional Dependency ($\Delta=-0.060$); Romantic Obsession ($\Delta=-0.042$). This suggests a poet-specific rhetorical-affective emphasis that is stable at aggregate scale while remaining compatible with multiple literary readings.

\begin{figure}[t]
\centering
\includegraphics[width=0.92\linewidth]{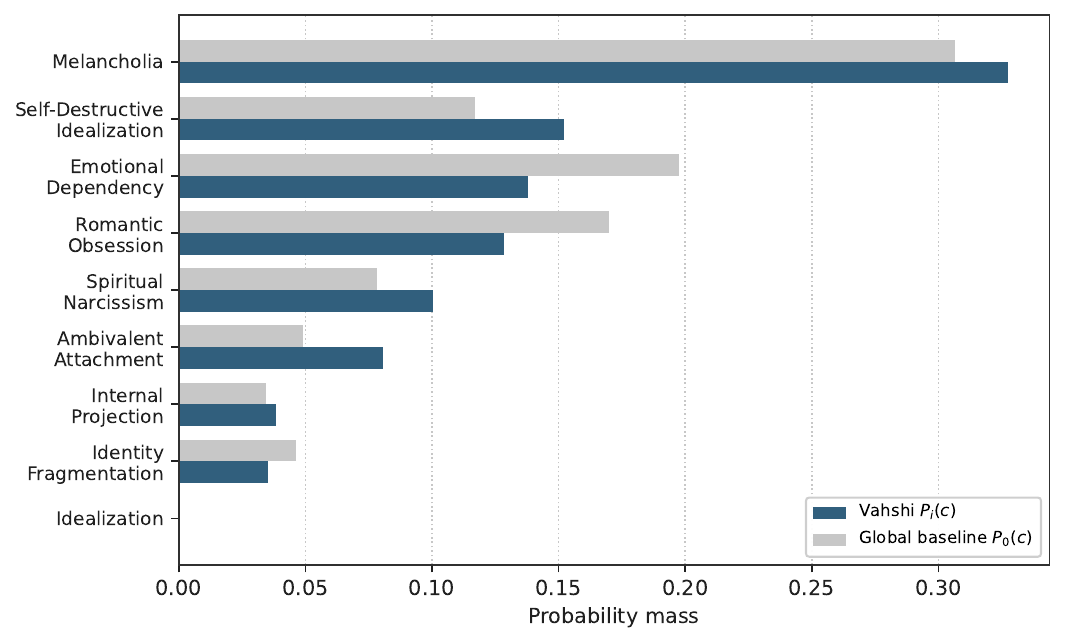}
\caption{Vahshi: concept distribution $P_i(c)$ against the global baseline $P_0(c)$.}
\label{fig:vahshi_p1}
\end{figure}

\begin{figure}[t]
\centering
\includegraphics[width=0.92\linewidth]{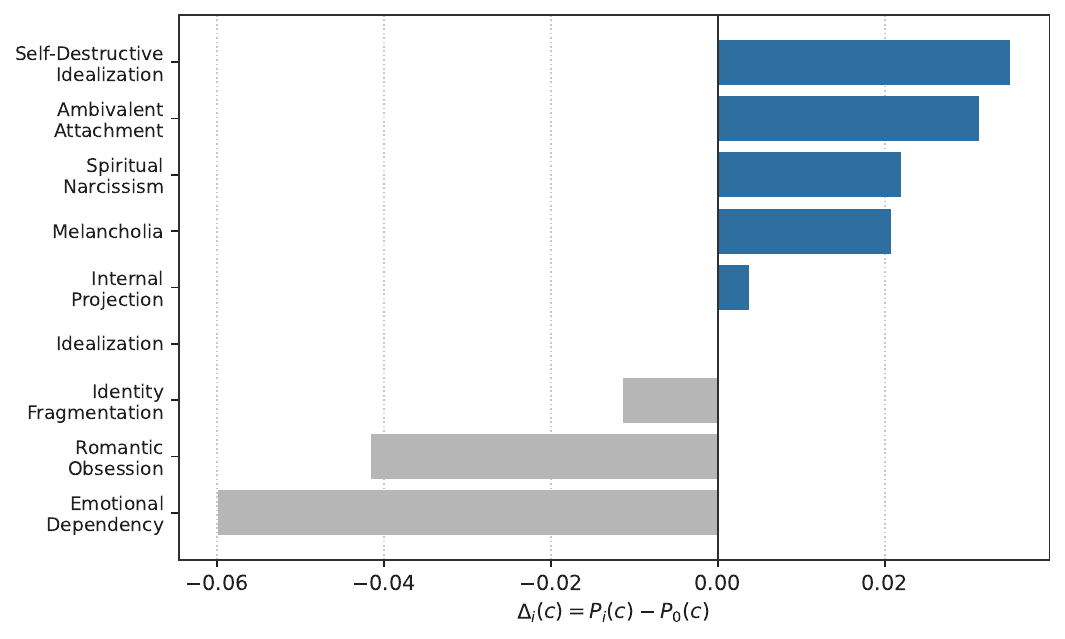}
\caption{Vahshi: lift profile $\Delta_i(c)=P_i(c)-P_0(c)$ with positive and negative deviations.}
\label{fig:vahshi_p2}
\end{figure}

\begin{figure}[t]
\centering
\includegraphics[width=0.72\linewidth]{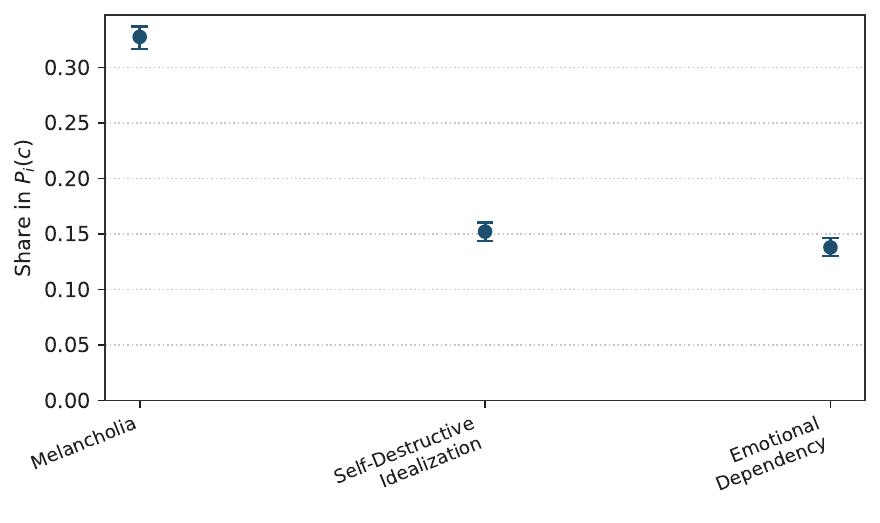}
\caption{Vahshi: bootstrap 95\% intervals for the top-3 concepts by $P_i(c)$.}
\label{fig:vahshi_p3}
\end{figure}

\noindent\textbf{Distant-to-Close Exemplars.}
\begin{table}[t]
\centering
\caption{Vahshi: top verses on the poet's most extreme Eigenmood axis (EM3, positive direction).}
\label{tab:vahshi_axis_exemplars}
\small
\renewcommand{\arraystretch}{1.1}
\begin{tabularx}{0.98\linewidth}{@{}>{\raggedright\arraybackslash}p{0.21\linewidth} S[table-format=-1.3] >{\raggedright\arraybackslash}X@{}}
\toprule
Reference & {Score} & English translation \\
\midrule
EM3 (positive) & 0.505 & \textit{I remember the day your hem was in my hand: you struck with the dagger, and I kissed your feet.} (Translation ours.) \\
\addlinespace[0.35em]
EM3 (positive) & 0.499 & \textit{If you know how many birds circle your snare, cast your trap in the plain, but first cast it over our hearts.} (Translation ours.) \\
\addlinespace[0.35em]
EM3 (positive) & 0.499 & \textit{To strike down thousands unaware, the flower-dagger is hidden beneath the hem.} (Translation ours.) \\
\addlinespace[0.35em]
\bottomrule
\end{tabularx}
\end{table}

In this axis-conditioned retrieval, the selected verses track Vahshi's dominant displacement along EM3: the retrieved lines concentrate lexical and imagistic material that is directionally coherent with the axis loadings, and therefore operationalize the poet's high-level position in the spectral space at verse scale.

\begin{table}[t]
\centering
\caption{Vahshi: top verses for the leading concept (Melancholia).}
\label{tab:vahshi_label_exemplars}
\small
\renewcommand{\arraystretch}{1.1}
\begin{tabularx}{0.98\linewidth}{@{}>{\raggedright\arraybackslash}p{0.21\linewidth} S[table-format=1.3] >{\raggedright\arraybackslash}X@{}}
\toprule
Reference & {Score} & English translation \\
\midrule
Melancholia & 0.900 & \textit{Do not ask after my wounded heart; leave me for a moment with my sorrow and affliction.} (Translation ours.) \\
\addlinespace[0.35em]
Melancholia & 0.900 & \textit{Restless, disordered in mind, constricted in heart, and grief-stricken.} (Translation ours.) \\
\addlinespace[0.35em]
Melancholia & 0.850 & \textit{I once held a treasure of patience in my heart; the army of sorrow plundered that whole treasury.} (Translation ours.) \\
\addlinespace[0.35em]
\bottomrule
\end{tabularx}
\end{table}

The concept-focused retrieval refines this reading: verses with the highest confidence for Melancholia instantiate recurring rhetorical operations that anchor the aggregate profile. Read together with the axis retrieval, they provide a distant-to-close bridge from distributional signature to inspectable poetic evidence.

\noindent\textbf{Cautions.} These profiles model operational constructs in textual discourse and do not support clinical inference about historical persons. Because abstention filters uncertain cases, selection effects remain analytically relevant; profile interpretation therefore combines confidence-weighted aggregates with explicit awareness of model non-commitment.

\FloatBarrier

\subsection{Poet 5: Parvin}

\noindent\textbf{Historical and Literary Context.} The observed configuration aligns with scholarship on Parvin Etesami highlighting didactic-dialogic structures and moral argumentation. \cite{iranica_parvin}

\noindent\textbf{Poet Summary.} Parvin contributes 5,573 verses to the corpus, with abstention rate 0.384 and divergence from baseline $D_{JS}=0.0459$. The uncertainty-aware profile is concentrated in Melancholia, Self-Destructive Idealization, and Spiritual Narcissism.

\begin{table}[t]
\centering
\caption{Core profile metrics for \textsc{Parvin}.}
\label{tab:poet_profile_parvin_metrics}
\footnotesize
\renewcommand{\arraystretch}{1.1}
\begin{tabular}{l S[table-format=5.0] S[table-format=1.3] S[table-format=1.4]}
\toprule
Poet & {Verses} & {Abstain} & {$D_{JS}$} \\
\midrule
Parvin & 5573 & 0.384 & 0.0459 \\
\bottomrule
\end{tabular}
\end{table}

\begin{table}[t]
\centering
\caption{Eigenmood coordinates for \textsc{Parvin}.}
\label{tab:poet_profile_parvin_embedding}
\footnotesize
\renewcommand{\arraystretch}{1.1}
\begin{tabular}{l S[table-format=+1.3] S[table-format=+1.3] S[table-format=+1.3]}
\toprule
Poet & {EM1} & {EM2} & {EM3} \\
\midrule
Parvin & 0.039 & -0.027 & -0.056 \\
\bottomrule
\end{tabular}
\end{table}

\begin{table}[t]
\centering
\caption{Top concept shares for \textsc{Parvin}. Concept names are abbreviated in the header; full interpretation appears in the text.}
\label{tab:poet_profile_parvin_concepts}
\small
\renewcommand{\arraystretch}{1.1}
\begin{tabularx}{0.96\linewidth}{@{}l >{\raggedright\arraybackslash}X S[table-format=1.3]@{}}
\toprule
Rank & Concept & {$P_i(c)$} \\
\midrule
1 & Melancholia & 0.370 \\
2 & Self-Destructive Idealization & 0.190 \\
3 & Spiritual Narcissism & 0.116 \\
\bottomrule
\end{tabularx}
\end{table}

\noindent\textbf{Concept Distribution Interpretation.} We interpret $P_i(c)$ as a confidence-weighted, abstention-aware distribution over operational constructs rather than literal measurements of psychological states. Relative to the global baseline $P_0(c)$, Parvin is over-represented in Self-Destructive Idealization ($\Delta=+0.073$); Melancholia ($\Delta=+0.063$), and under-represented in Romantic Obsession ($\Delta=-0.126$); Emotional Dependency ($\Delta=-0.112$). This suggests a poet-specific rhetorical-affective emphasis that is stable at aggregate scale while remaining compatible with multiple literary readings.

\begin{figure}[t]
\centering
\includegraphics[width=0.92\linewidth]{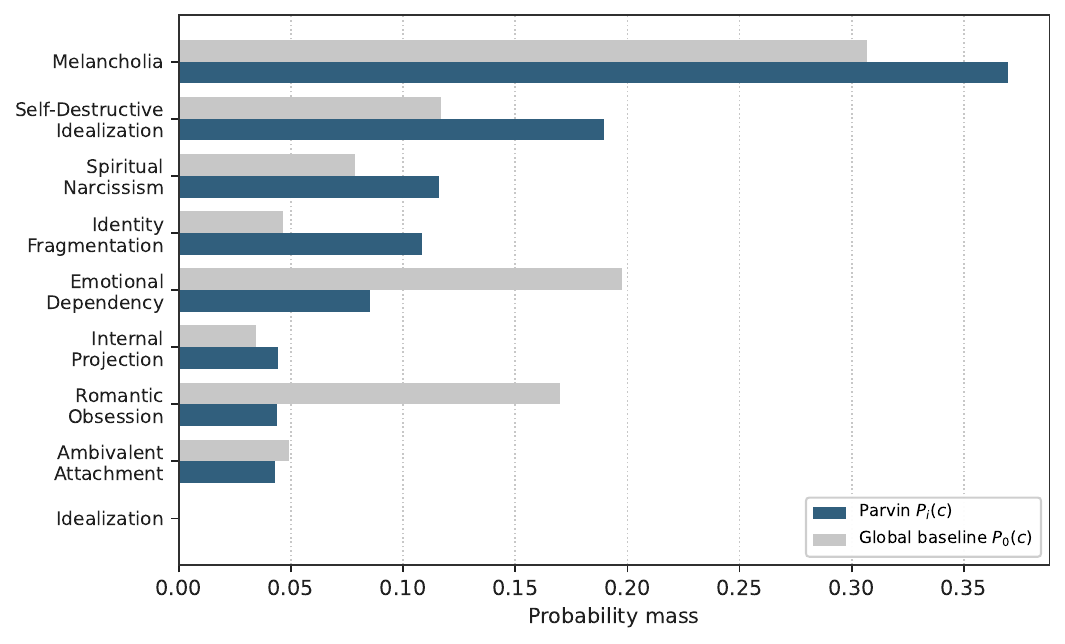}
\caption{Parvin: concept distribution $P_i(c)$ against the global baseline $P_0(c)$.}
\label{fig:parvin_p1}
\end{figure}

\begin{figure}[t]
\centering
\includegraphics[width=0.92\linewidth]{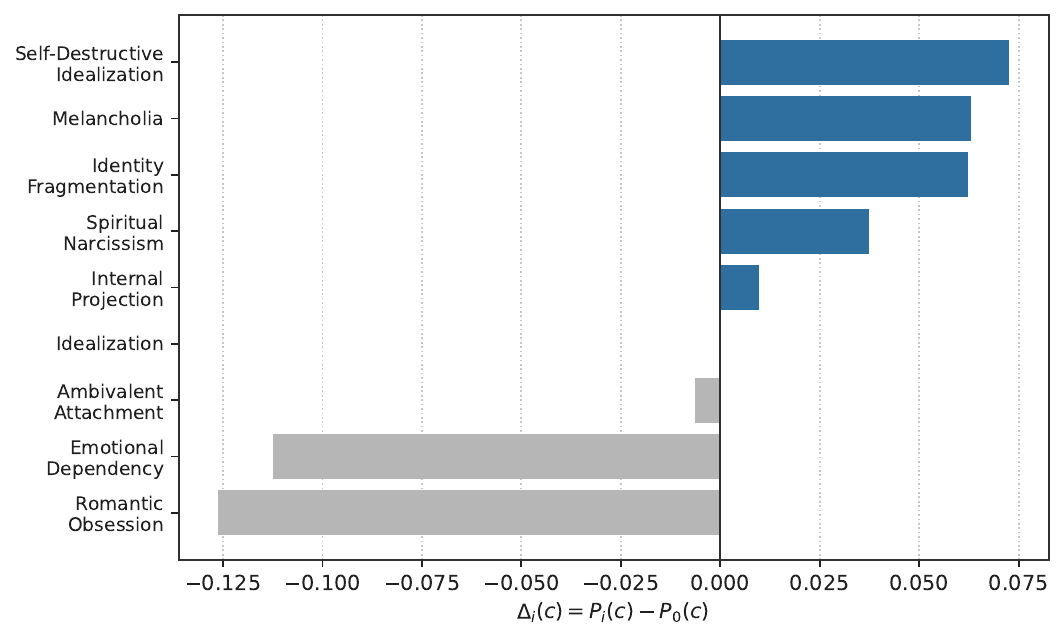}
\caption{Parvin: lift profile $\Delta_i(c)=P_i(c)-P_0(c)$ with positive and negative deviations.}
\label{fig:parvin_p2}
\end{figure}

\begin{figure}[t]
\centering
\includegraphics[width=0.72\linewidth]{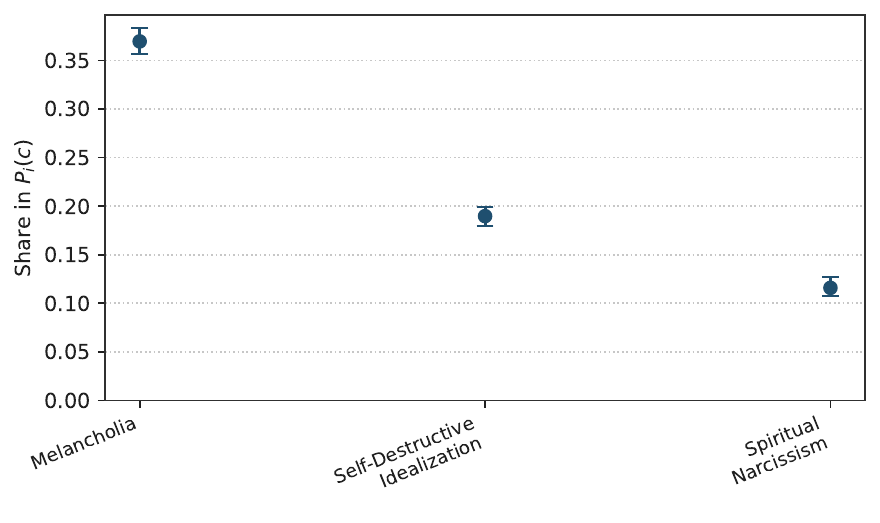}
\caption{Parvin: bootstrap 95\% intervals for the top-3 concepts by $P_i(c)$.}
\label{fig:parvin_p3}
\end{figure}

\noindent\textbf{Distant-to-Close Exemplars.}
\begin{table}[t]
\centering
\caption{Parvin: top verses on the poet's most extreme Eigenmood axis (EM3, negative direction).}
\label{tab:parvin_axis_exemplars}
\small
\renewcommand{\arraystretch}{1.1}
\begin{tabularx}{0.98\linewidth}{@{}>{\raggedright\arraybackslash}p{0.21\linewidth} S[table-format=-1.3] >{\raggedright\arraybackslash}X@{}}
\toprule
Reference & {Score} & English translation \\
\midrule
EM3 (negative) & -0.601 & \textit{The soul's radiant palace stands firm; what endurance and permanence can the body's hut possess?} (Translation ours.) \\
\addlinespace[0.35em]
EM3 (negative) & -0.601 & \textit{The head, without the lamp of reason, is trapped in darkness; the body without spirit is scattered like dust.} (Translation ours.) \\
\addlinespace[0.35em]
EM3 (negative) & -0.601 & \textit{A manifest tale can never be hidden from the eye of reason.} (Translation ours.) \\
\addlinespace[0.35em]
\bottomrule
\end{tabularx}
\end{table}

In this axis-conditioned retrieval, the selected verses track Parvin's dominant displacement along EM3: the retrieved lines concentrate lexical and imagistic material that is directionally coherent with the axis loadings, and therefore operationalize the poet's high-level position in the spectral space at verse scale.

\begin{table}[t]
\centering
\caption{Parvin: top verses for the leading concept (Melancholia).}
\label{tab:parvin_label_exemplars}
\small
\renewcommand{\arraystretch}{1.1}
\begin{tabularx}{0.98\linewidth}{@{}>{\raggedright\arraybackslash}p{0.21\linewidth} S[table-format=1.3] >{\raggedright\arraybackslash}X@{}}
\toprule
Reference & {Score} & English translation \\
\midrule
Melancholia & 0.900 & \textit{Sit in a heart that has been cleansed of grief; then gladden me, O comforter of sorrowing hearts.} (Translation ours.) \\
\addlinespace[0.35em]
Melancholia & 0.900 & \textit{Why must one live so desolate, dead at the morning of life?} (Translation ours.) \\
\addlinespace[0.35em]
Melancholia & 0.900 & \textit{What shall I sing but grief's melody? What shall I say to moonlight and dew?} (Translation ours.) \\
\addlinespace[0.35em]
\bottomrule
\end{tabularx}
\end{table}

The concept-focused retrieval refines this reading: verses with the highest confidence for Melancholia instantiate recurring rhetorical operations that anchor the aggregate profile. Read together with the axis retrieval, they provide a distant-to-close bridge from distributional signature to inspectable poetic evidence.

\noindent\textbf{Cautions.} These profiles model operational constructs in textual discourse and do not support clinical inference about historical persons. Because abstention filters uncertain cases, selection effects remain analytically relevant; profile interpretation therefore combines confidence-weighted aggregates with explicit awareness of model non-commitment.

\FloatBarrier

\subsection{Poet 6: Hafez}

\noindent\textbf{Historical and Literary Context.} This pattern is consistent with scholarship on Hafez emphasizing polyvalent wine/tavern imagery and critique of hypocrisy across lyric registers. \cite{iranica_hafez,britannica_hafez}

\noindent\textbf{Poet Summary.} Hafez contributes 5,221 verses to the corpus, with abstention rate 0.182 and divergence from baseline $D_{JS}=0.0035$. The uncertainty-aware profile is concentrated in Melancholia, Romantic Obsession, and Emotional Dependency.

\begin{table}[t]
\centering
\caption{Core profile metrics for \textsc{Hafez}.}
\label{tab:poet_profile_hafez_metrics}
\footnotesize
\renewcommand{\arraystretch}{1.1}
\begin{tabular}{l S[table-format=5.0] S[table-format=1.3] S[table-format=1.4]}
\toprule
Poet & {Verses} & {Abstain} & {$D_{JS}$} \\
\midrule
Hafez & 5221 & 0.182 & 0.0035 \\
\bottomrule
\end{tabular}
\end{table}

\begin{table}[t]
\centering
\caption{Eigenmood coordinates for \textsc{Hafez}.}
\label{tab:poet_profile_hafez_embedding}
\footnotesize
\renewcommand{\arraystretch}{1.1}
\begin{tabular}{l S[table-format=+1.3] S[table-format=+1.3] S[table-format=+1.3]}
\toprule
Poet & {EM1} & {EM2} & {EM3} \\
\midrule
Hafez & 0.019 & 0.003 & -0.027 \\
\bottomrule
\end{tabular}
\end{table}

\begin{table}[t]
\centering
\caption{Top concept shares for \textsc{Hafez}. Concept names are abbreviated in the header; full interpretation appears in the text.}
\label{tab:poet_profile_hafez_concepts}
\small
\renewcommand{\arraystretch}{1.1}
\begin{tabularx}{0.96\linewidth}{@{}l >{\raggedright\arraybackslash}X S[table-format=1.3]@{}}
\toprule
Rank & Concept & {$P_i(c)$} \\
\midrule
1 & Melancholia & 0.294 \\
2 & Romantic Obsession & 0.188 \\
3 & Emotional Dependency & 0.160 \\
\bottomrule
\end{tabularx}
\end{table}

\noindent\textbf{Concept Distribution Interpretation.} We interpret $P_i(c)$ as a confidence-weighted, abstention-aware distribution over operational constructs rather than literal measurements of psychological states. Relative to the global baseline $P_0(c)$, Hafez is over-represented in Romantic Obsession ($\Delta=+0.018$); Identity Fragmentation ($\Delta=+0.016$), and under-represented in Emotional Dependency ($\Delta=-0.038$); Ambivalent Attachment ($\Delta=-0.016$). This suggests a poet-specific rhetorical-affective emphasis that is stable at aggregate scale while remaining compatible with multiple literary readings.

\begin{figure}[t]
\centering
\includegraphics[width=0.92\linewidth]{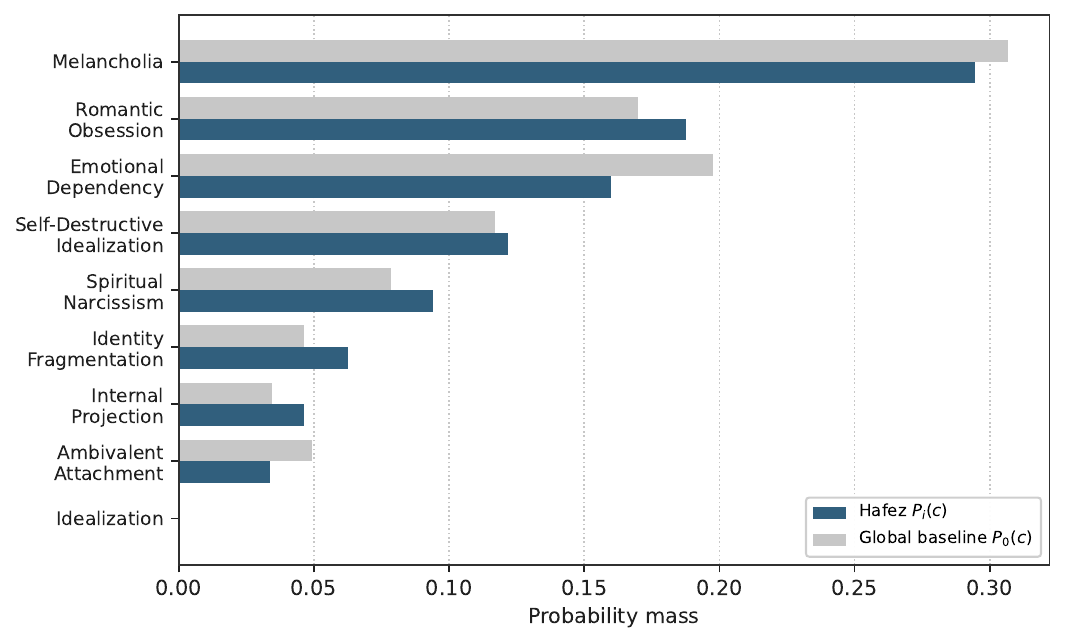}
\caption{Hafez: concept distribution $P_i(c)$ against the global baseline $P_0(c)$.}
\label{fig:hafez_p1}
\end{figure}

\begin{figure}[t]
\centering
\includegraphics[width=0.92\linewidth]{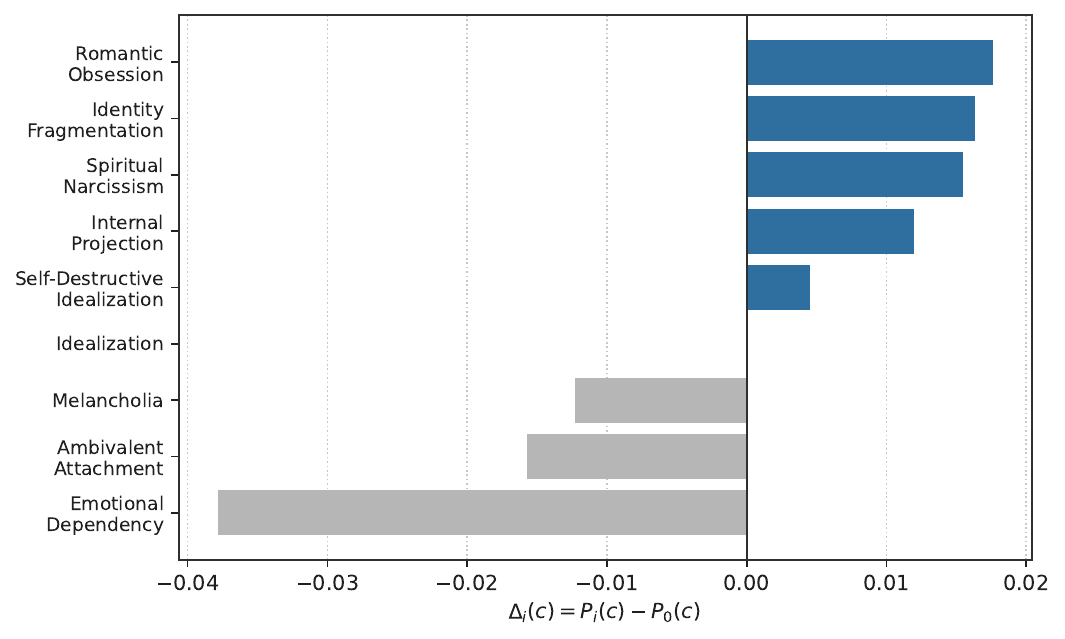}
\caption{Hafez: lift profile $\Delta_i(c)=P_i(c)-P_0(c)$ with positive and negative deviations.}
\label{fig:hafez_p2}
\end{figure}

\begin{figure}[t]
\centering
\includegraphics[width=0.72\linewidth]{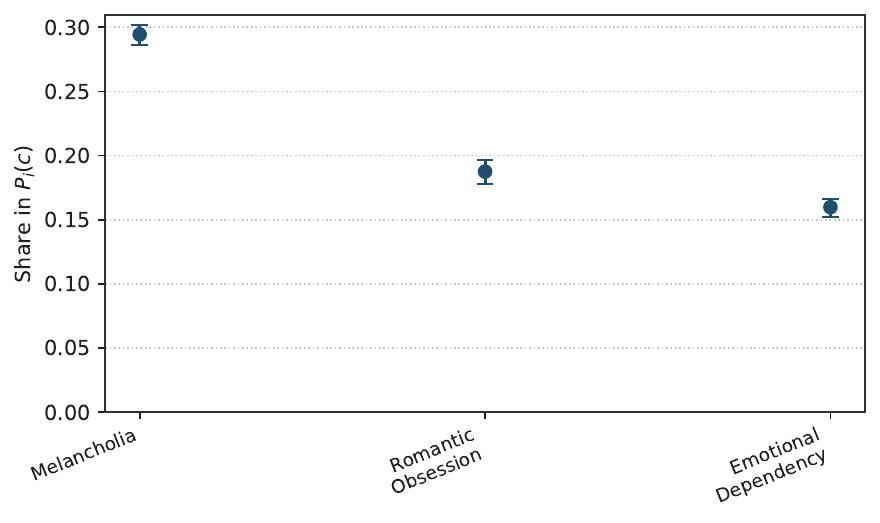}
\caption{Hafez: bootstrap 95\% intervals for the top-3 concepts by $P_i(c)$.}
\label{fig:hafez_p3}
\end{figure}

\noindent\textbf{Distant-to-Close Exemplars.}
\begin{table}[t]
\centering
\caption{Hafez: top verses on the poet's most extreme Eigenmood axis (EM3, negative direction).}
\label{tab:hafez_axis_exemplars}
\small
\renewcommand{\arraystretch}{1.1}
\begin{tabularx}{0.98\linewidth}{@{}>{\raggedright\arraybackslash}p{0.21\linewidth} S[table-format=-1.3] >{\raggedright\arraybackslash}X@{}}
\toprule
Reference & {Score} & English translation \\
\midrule
EM3 (negative) & -0.759 & \textit{Within my wounded heart I do not know who cries out; I am silent, while he is uproar and lament.} (Translation ours.) \\
\addlinespace[0.35em]
EM3 (negative) & -0.759 & \textit{Do not say that, absent heart, I forgot you by mistake.} (Translation ours.) \\
\addlinespace[0.35em]
EM3 (negative) & -0.724 & \textit{My heart was a treasury of secrets; fate locked it and handed its key to the stealer of hearts.} (Translation ours.) \\
\addlinespace[0.35em]
\bottomrule
\end{tabularx}
\end{table}

In this axis-conditioned retrieval, the selected verses track Hafez's dominant displacement along EM3: the retrieved lines concentrate lexical and imagistic material that is directionally coherent with the axis loadings, and therefore operationalize the poet's high-level position in the spectral space at verse scale.

\begin{table}[t]
\centering
\caption{Hafez: top verses for the leading concept (Melancholia).}
\label{tab:hafez_label_exemplars}
\small
\renewcommand{\arraystretch}{1.1}
\begin{tabularx}{0.98\linewidth}{@{}>{\raggedright\arraybackslash}p{0.21\linewidth} S[table-format=1.3] >{\raggedright\arraybackslash}X@{}}
\toprule
Reference & {Score} & English translation \\
\midrule
Melancholia & 0.900 & \textit{My palate turned to poison from grief's bitterness; remember the cry of those who drink joy.} (Translation ours.) \\
\addlinespace[0.35em]
Melancholia & 0.900 & \textit{I go to the tavern weeping and seeking redress; perhaps there I may be released from sorrow's hand.} (Translation ours.) \\
\addlinespace[0.35em]
Melancholia & 0.900 & \textit{I have endured a love-pain beyond asking; I have tasted the poison of separation beyond asking.} (Translation ours.) \\
\addlinespace[0.35em]
\bottomrule
\end{tabularx}
\end{table}

The concept-focused retrieval refines this reading: verses with the highest confidence for Melancholia instantiate recurring rhetorical operations that anchor the aggregate profile. Read together with the axis retrieval, they provide a distant-to-close bridge from distributional signature to inspectable poetic evidence.

\noindent\textbf{Cautions.} These profiles model operational constructs in textual discourse and do not support clinical inference about historical persons. Because abstention filters uncertain cases, selection effects remain analytically relevant; profile interpretation therefore combines confidence-weighted aggregates with explicit awareness of model non-commitment.

\FloatBarrier

\subsection{Poet 7: Eraghi}

\noindent\textbf{Historical and Literary Context.} The observed balance between inwardized affect and devotional language aligns with scholarship on Fakhr-al-Din Eraqi and his Sufi lyric-mystical poetics. \cite{iranica_eraqi}

\noindent\textbf{Poet Summary.} Eraghi contributes 4,668 verses to the corpus, with abstention rate 0.081 and divergence from baseline $D_{JS}=0.0040$. The uncertainty-aware profile is concentrated in Melancholia, Emotional Dependency, and Romantic Obsession.

\begin{table}[t]
\centering
\caption{Core profile metrics for \textsc{Eraghi}.}
\label{tab:poet_profile_eraghi_metrics}
\footnotesize
\renewcommand{\arraystretch}{1.1}
\begin{tabular}{l S[table-format=5.0] S[table-format=1.3] S[table-format=1.4]}
\toprule
Poet & {Verses} & {Abstain} & {$D_{JS}$} \\
\midrule
Eraghi & 4668 & 0.081 & 0.0040 \\
\bottomrule
\end{tabular}
\end{table}

\begin{table}[t]
\centering
\caption{Eigenmood coordinates for \textsc{Eraghi}.}
\label{tab:poet_profile_eraghi_embedding}
\footnotesize
\renewcommand{\arraystretch}{1.1}
\begin{tabular}{l S[table-format=+1.3] S[table-format=+1.3] S[table-format=+1.3]}
\toprule
Poet & {EM1} & {EM2} & {EM3} \\
\midrule
Eraghi & 0.014 & 0.016 & -0.010 \\
\bottomrule
\end{tabular}
\end{table}

\begin{table}[t]
\centering
\caption{Top concept shares for \textsc{Eraghi}. Concept names are abbreviated in the header; full interpretation appears in the text.}
\label{tab:poet_profile_eraghi_concepts}
\small
\renewcommand{\arraystretch}{1.1}
\begin{tabularx}{0.96\linewidth}{@{}l >{\raggedright\arraybackslash}X S[table-format=1.3]@{}}
\toprule
Rank & Concept & {$P_i(c)$} \\
\midrule
1 & Melancholia & 0.277 \\
2 & Emotional Dependency & 0.241 \\
3 & Romantic Obsession & 0.181 \\
\bottomrule
\end{tabularx}
\end{table}

\noindent\textbf{Concept Distribution Interpretation.} We interpret $P_i(c)$ as a confidence-weighted, abstention-aware distribution over operational constructs rather than literal measurements of psychological states. Relative to the global baseline $P_0(c)$, Eraghi is over-represented in Emotional Dependency ($\Delta=+0.043$); Internal Projection ($\Delta=+0.013$), and under-represented in Melancholia ($\Delta=-0.030$); Self-Destructive Idealization ($\Delta=-0.028$). This suggests a poet-specific rhetorical-affective emphasis that is stable at aggregate scale while remaining compatible with multiple literary readings.

\begin{figure}[t]
\centering
\includegraphics[width=0.92\linewidth]{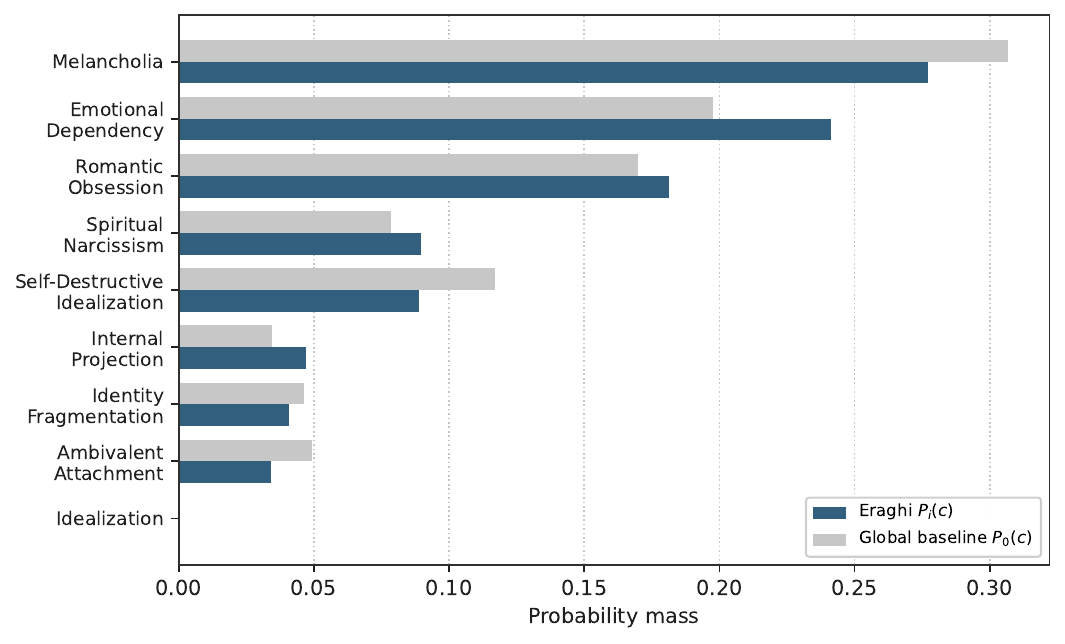}
\caption{Eraghi: concept distribution $P_i(c)$ against the global baseline $P_0(c)$.}
\label{fig:eraghi_p1}
\end{figure}

\begin{figure}[t]
\centering
\includegraphics[width=0.92\linewidth]{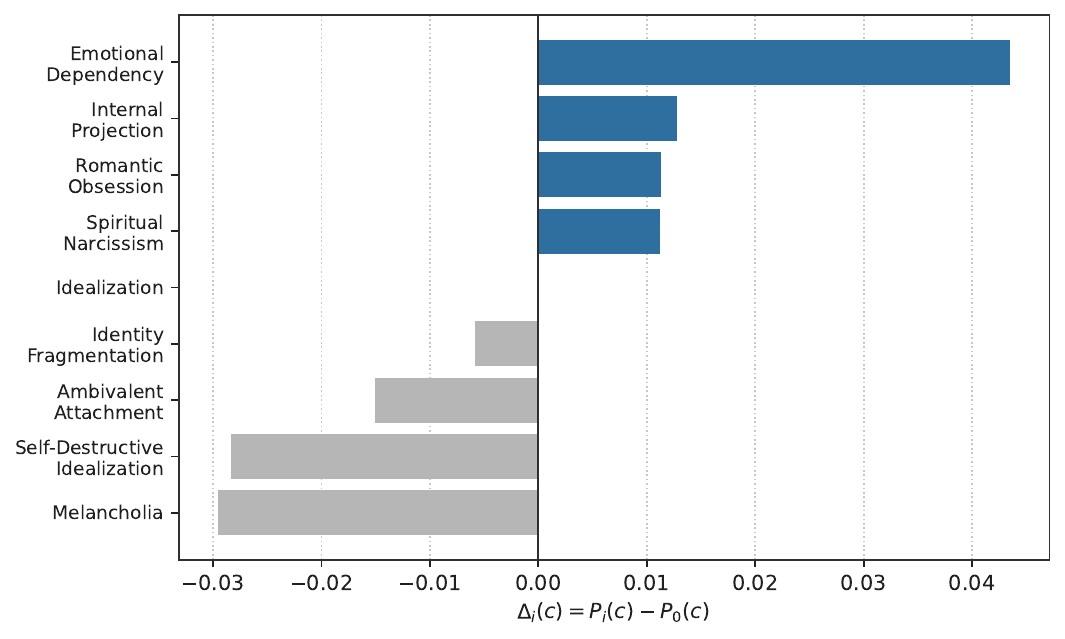}
\caption{Eraghi: lift profile $\Delta_i(c)=P_i(c)-P_0(c)$ with positive and negative deviations.}
\label{fig:eraghi_p2}
\end{figure}

\begin{figure}[t]
\centering
\includegraphics[width=0.72\linewidth]{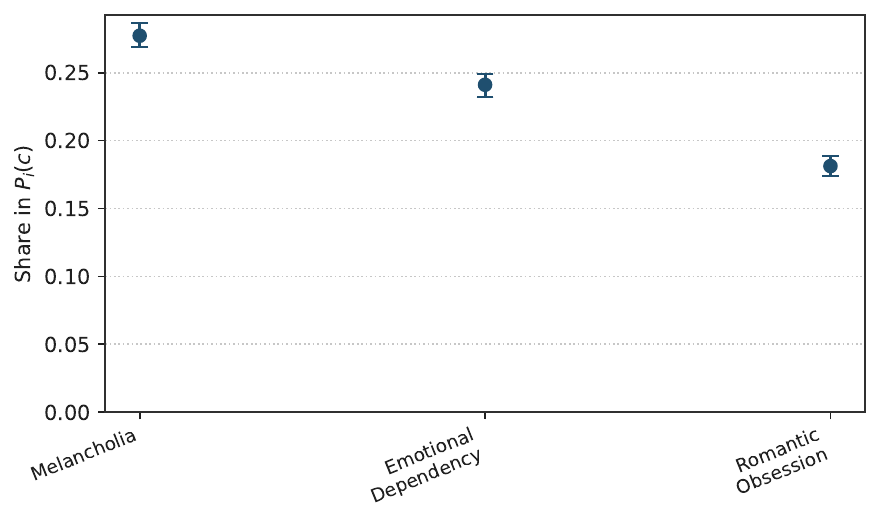}
\caption{Eraghi: bootstrap 95\% intervals for the top-3 concepts by $P_i(c)$.}
\label{fig:eraghi_p3}
\end{figure}

\noindent\textbf{Distant-to-Close Exemplars.}
\begin{table}[t]
\centering
\caption{Eraghi: top verses on the poet's most extreme Eigenmood axis (EM2, positive direction).}
\label{tab:eraghi_axis_exemplars}
\small
\renewcommand{\arraystretch}{1.1}
\begin{tabularx}{0.98\linewidth}{@{}>{\raggedright\arraybackslash}p{0.21\linewidth} S[table-format=-1.3] >{\raggedright\arraybackslash}X@{}}
\toprule
Reference & {Score} & English translation \\
\midrule
EM2 (positive) & 0.796 & \textit{In vain I sought you through the world; you were already within Iraqi's soul.} (Translation ours.) \\
\addlinespace[0.35em]
EM2 (positive) & 0.754 & \textit{If you look to left and right, behind and before, know that before, behind, left, right, and above are all He.} (Translation ours.) \\
\addlinespace[0.35em]
EM2 (positive) & 0.712 & \textit{He set speech upon our tongues and planted seeking within us.} (Translation ours.) \\
\addlinespace[0.35em]
\bottomrule
\end{tabularx}
\end{table}

In this axis-conditioned retrieval, the selected verses track Eraghi's dominant displacement along EM2: the retrieved lines concentrate lexical and imagistic material that is directionally coherent with the axis loadings, and therefore operationalize the poet's high-level position in the spectral space at verse scale.

\begin{table}[t]
\centering
\caption{Eraghi: top verses for the leading concept (Melancholia).}
\label{tab:eraghi_label_exemplars}
\small
\renewcommand{\arraystretch}{1.1}
\begin{tabularx}{0.98\linewidth}{@{}>{\raggedright\arraybackslash}p{0.21\linewidth} S[table-format=1.3] >{\raggedright\arraybackslash}X@{}}
\toprule
Reference & {Score} & English translation \\
\midrule
Melancholia & 0.950 & \textit{At every moment I weep; I weep from the grief of time.} (Translation ours.) \\
\addlinespace[0.35em]
Melancholia & 0.950 & \textit{So long as Iraqi remains in the pain of parting, speak not of cure and do not try to heal me.} (Translation ours.) \\
\addlinespace[0.35em]
Melancholia & 0.950 & \textit{I plucked no flower from the garden of joy; from grief, a hundred thorns were lodged in my soul.} (Translation ours.) \\
\addlinespace[0.35em]
\bottomrule
\end{tabularx}
\end{table}

The concept-focused retrieval refines this reading: verses with the highest confidence for Melancholia instantiate recurring rhetorical operations that anchor the aggregate profile. Read together with the axis retrieval, they provide a distant-to-close bridge from distributional signature to inspectable poetic evidence.

\noindent\textbf{Cautions.} These profiles model operational constructs in textual discourse and do not support clinical inference about historical persons. Because abstention filters uncertain cases, selection effects remain analytically relevant; profile interpretation therefore combines confidence-weighted aggregates with explicit awareness of model non-commitment.

\FloatBarrier

\subsection{Poet 8: Shahriar}

\noindent\textbf{Historical and Literary Context.} The concentration in lament-oriented constructs is consistent with scholarship on Shahriar's modern reworking of classical lyric resources. \cite{iranica_shahriar}

\noindent\textbf{Poet Summary.} Shahriar contributes 1,970 verses to the corpus, with abstention rate 0.141 and divergence from baseline $D_{JS}=0.0030$. The uncertainty-aware profile is concentrated in Melancholia, Romantic Obsession, and Emotional Dependency.

\begin{table}[t]
\centering
\caption{Core profile metrics for \textsc{Shahriar}.}
\label{tab:poet_profile_shahriar_metrics}
\footnotesize
\renewcommand{\arraystretch}{1.1}
\begin{tabular}{l S[table-format=5.0] S[table-format=1.3] S[table-format=1.4]}
\toprule
Poet & {Verses} & {Abstain} & {$D_{JS}$} \\
\midrule
Shahriar & 1970 & 0.141 & 0.0030 \\
\bottomrule
\end{tabular}
\end{table}

\begin{table}[t]
\centering
\caption{Eigenmood coordinates for \textsc{Shahriar}.}
\label{tab:poet_profile_shahriar_embedding}
\footnotesize
\renewcommand{\arraystretch}{1.1}
\begin{tabular}{l S[table-format=+1.3] S[table-format=+1.3] S[table-format=+1.3]}
\toprule
Poet & {EM1} & {EM2} & {EM3} \\
\midrule
Shahriar & -0.007 & 0.010 & -0.012 \\
\bottomrule
\end{tabular}
\end{table}

\begin{table}[t]
\centering
\caption{Top concept shares for \textsc{Shahriar}. Concept names are abbreviated in the header; full interpretation appears in the text.}
\label{tab:poet_profile_shahriar_concepts}
\small
\renewcommand{\arraystretch}{1.1}
\begin{tabularx}{0.96\linewidth}{@{}l >{\raggedright\arraybackslash}X S[table-format=1.3]@{}}
\toprule
Rank & Concept & {$P_i(c)$} \\
\midrule
1 & Melancholia & 0.348 \\
2 & Romantic Obsession & 0.175 \\
3 & Emotional Dependency & 0.151 \\
\bottomrule
\end{tabularx}
\end{table}

\noindent\textbf{Concept Distribution Interpretation.} We interpret $P_i(c)$ as a confidence-weighted, abstention-aware distribution over operational constructs rather than literal measurements of psychological states. Relative to the global baseline $P_0(c)$, Shahriar is over-represented in Melancholia ($\Delta=+0.041$); Internal Projection ($\Delta=+0.013$), and under-represented in Emotional Dependency ($\Delta=-0.046$); Spiritual Narcissism ($\Delta=-0.007$). This suggests a poet-specific rhetorical-affective emphasis that is stable at aggregate scale while remaining compatible with multiple literary readings.

\begin{figure}[t]
\centering
\includegraphics[width=0.92\linewidth]{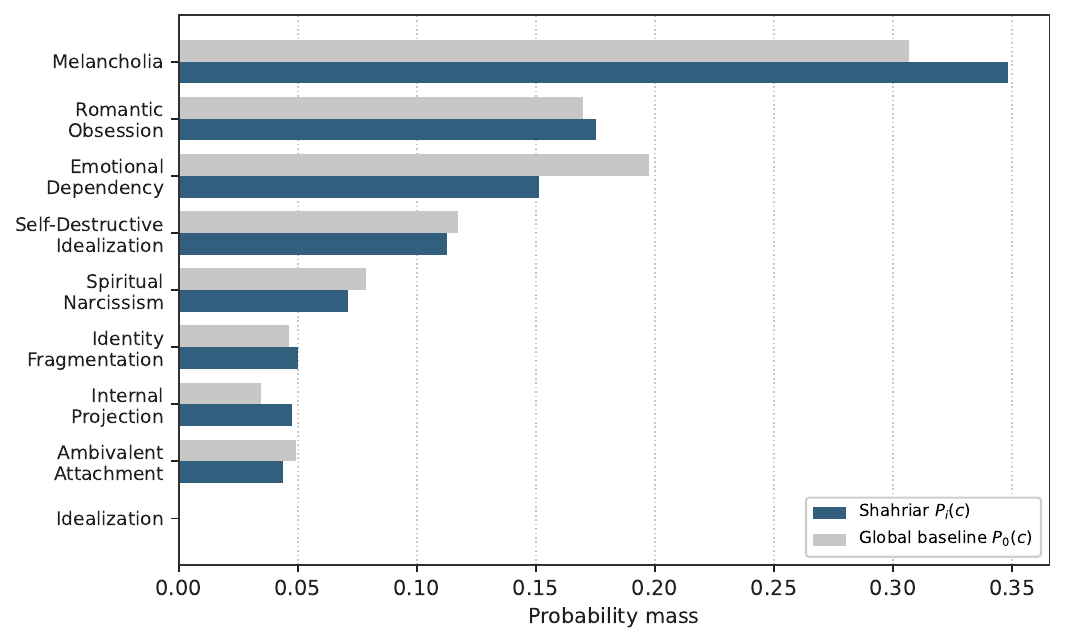}
\caption{Shahriar: concept distribution $P_i(c)$ against the global baseline $P_0(c)$.}
\label{fig:shahriar_p1}
\end{figure}

\begin{figure}[t]
\centering
\includegraphics[width=0.92\linewidth]{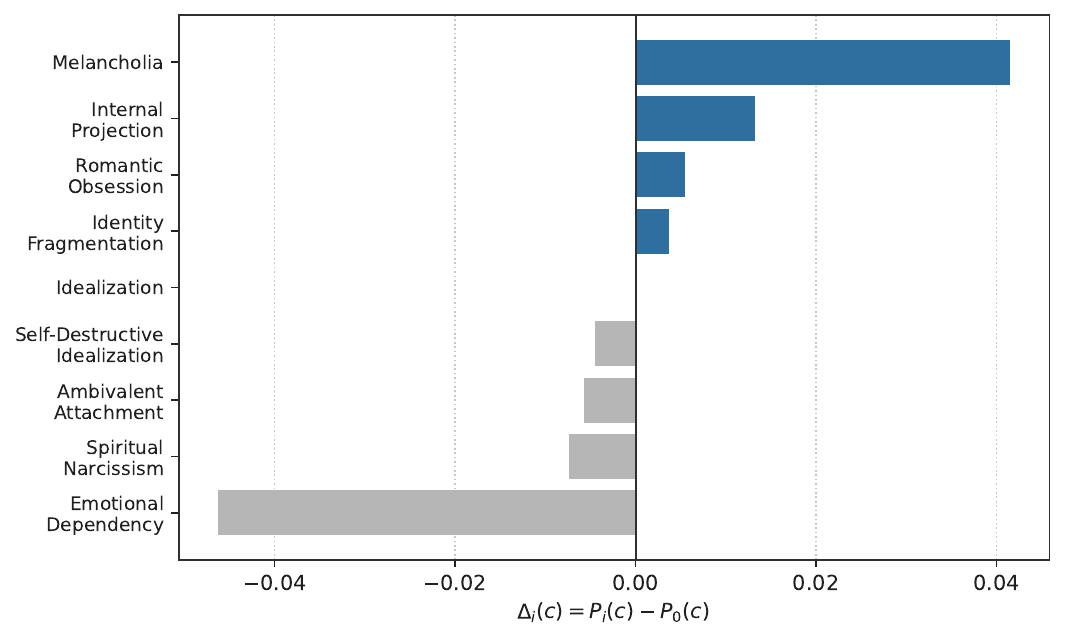}
\caption{Shahriar: lift profile $\Delta_i(c)=P_i(c)-P_0(c)$ with positive and negative deviations.}
\label{fig:shahriar_p2}
\end{figure}

\begin{figure}[t]
\centering
\includegraphics[width=0.72\linewidth]{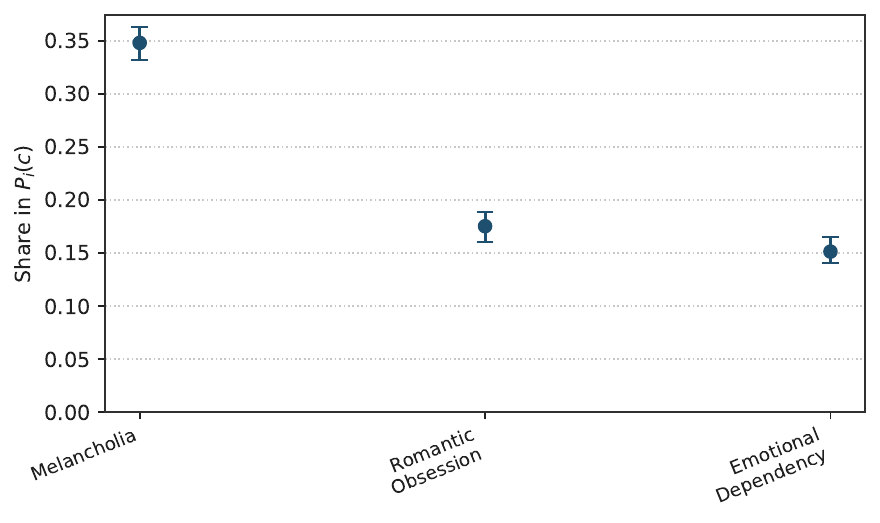}
\caption{Shahriar: bootstrap 95\% intervals for the top-3 concepts by $P_i(c)$.}
\label{fig:shahriar_p3}
\end{figure}

\noindent\textbf{Distant-to-Close Exemplars.}
\begin{table}[t]
\centering
\caption{Shahriar: top verses on the poet's most extreme Eigenmood axis (EM3, negative direction).}
\label{tab:shahriar_axis_exemplars}
\small
\renewcommand{\arraystretch}{1.1}
\begin{tabularx}{0.98\linewidth}{@{}>{\raggedright\arraybackslash}p{0.21\linewidth} S[table-format=-1.3] >{\raggedright\arraybackslash}X@{}}
\toprule
Reference & {Score} & English translation \\
\midrule
EM3 (negative) & -0.759 & \textit{If your heart would open, speak the truth of companionship: you were never my friend; I was yours.} (Translation ours.) \\
\addlinespace[0.35em]
EM3 (negative) & -0.634 & \textit{Let the lamp of wakefulness cast its light; every heart's eye is asleep in heedlessness, my friend.} (Translation ours.) \\
\addlinespace[0.35em]
EM3 (negative) & -0.601 & \textit{The gates of gnosis are shut before your reason; the key to our coffer is love.} (Translation ours.) \\
\addlinespace[0.35em]
\bottomrule
\end{tabularx}
\end{table}

In this axis-conditioned retrieval, the selected verses track Shahriar's dominant displacement along EM3: the retrieved lines concentrate lexical and imagistic material that is directionally coherent with the axis loadings, and therefore operationalize the poet's high-level position in the spectral space at verse scale.

\begin{table}[t]
\centering
\caption{Shahriar: top verses for the leading concept (Melancholia).}
\label{tab:shahriar_label_exemplars}
\small
\renewcommand{\arraystretch}{1.1}
\begin{tabularx}{0.98\linewidth}{@{}>{\raggedright\arraybackslash}p{0.21\linewidth} S[table-format=1.3] >{\raggedright\arraybackslash}X@{}}
\toprule
Reference & {Score} & English translation \\
\midrule
Melancholia & 0.900 & \textit{What a world, if there is a heart and no heart-soothing; what a life, if grief is present and no consoler.} (Translation ours.) \\
\addlinespace[0.35em]
Melancholia & 0.850 & \textit{The candle of delight bloomed in tears and sighs; cloud and moon rose together.} (Translation ours.) \\
\addlinespace[0.35em]
Melancholia & 0.850 & \textit{No one has read your lot as joy beneath this turning sky; here, rejection and denial are the standing decree.} (Translation ours.) \\
\addlinespace[0.35em]
\bottomrule
\end{tabularx}
\end{table}

The concept-focused retrieval refines this reading: verses with the highest confidence for Melancholia instantiate recurring rhetorical operations that anchor the aggregate profile. Read together with the axis retrieval, they provide a distant-to-close bridge from distributional signature to inspectable poetic evidence.

\noindent\textbf{Cautions.} These profiles model operational constructs in textual discourse and do not support clinical inference about historical persons. Because abstention filters uncertain cases, selection effects remain analytically relevant; profile interpretation therefore combines confidence-weighted aggregates with explicit awareness of model non-commitment.

\FloatBarrier

\subsection{Poet 9: Athir}

\noindent\textbf{Historical and Literary Context.} This profile is consistent with scholarship describing Athir Akhsikati as a twelfth-century court poet in a rhetorically dense qasida tradition. \cite{iranica_athir_akhsikati}

\noindent\textbf{Poet Summary.} Athir contributes 1,603 verses to the corpus, with abstention rate 0.138 and divergence from baseline $D_{JS}=0.0073$. The uncertainty-aware profile is concentrated in Melancholia, Romantic Obsession, and Emotional Dependency.

\begin{table}[t]
\centering
\caption{Core profile metrics for \textsc{Athir}.}
\label{tab:poet_profile_athir_metrics}
\footnotesize
\renewcommand{\arraystretch}{1.1}
\begin{tabular}{l S[table-format=5.0] S[table-format=1.3] S[table-format=1.4]}
\toprule
Poet & {Verses} & {Abstain} & {$D_{JS}$} \\
\midrule
Athir & 1603 & 0.138 & 0.0073 \\
\bottomrule
\end{tabular}
\end{table}

\begin{table}[t]
\centering
\caption{Eigenmood coordinates for \textsc{Athir}.}
\label{tab:poet_profile_athir_embedding}
\footnotesize
\renewcommand{\arraystretch}{1.1}
\begin{tabular}{l S[table-format=+1.3] S[table-format=+1.3] S[table-format=+1.3]}
\toprule
Poet & {EM1} & {EM2} & {EM3} \\
\midrule
Athir & -0.044 & -0.007 & 0.004 \\
\bottomrule
\end{tabular}
\end{table}

\begin{table}[t]
\centering
\caption{Top concept shares for \textsc{Athir}. Concept names are abbreviated in the header; full interpretation appears in the text.}
\label{tab:poet_profile_athir_concepts}
\small
\renewcommand{\arraystretch}{1.1}
\begin{tabularx}{0.96\linewidth}{@{}l >{\raggedright\arraybackslash}X S[table-format=1.3]@{}}
\toprule
Rank & Concept & {$P_i(c)$} \\
\midrule
1 & Melancholia & 0.290 \\
2 & Romantic Obsession & 0.222 \\
3 & Emotional Dependency & 0.167 \\
\bottomrule
\end{tabularx}
\end{table}

\noindent\textbf{Concept Distribution Interpretation.} We interpret $P_i(c)$ as a confidence-weighted, abstention-aware distribution over operational constructs rather than literal measurements of psychological states. Relative to the global baseline $P_0(c)$, Athir is over-represented in Romantic Obsession ($\Delta=+0.052$); Ambivalent Attachment ($\Delta=+0.019$), and under-represented in Spiritual Narcissism ($\Delta=-0.041$); Emotional Dependency ($\Delta=-0.031$). This suggests a poet-specific rhetorical-affective emphasis that is stable at aggregate scale while remaining compatible with multiple literary readings.

\begin{figure}[t]
\centering
\includegraphics[width=0.92\linewidth]{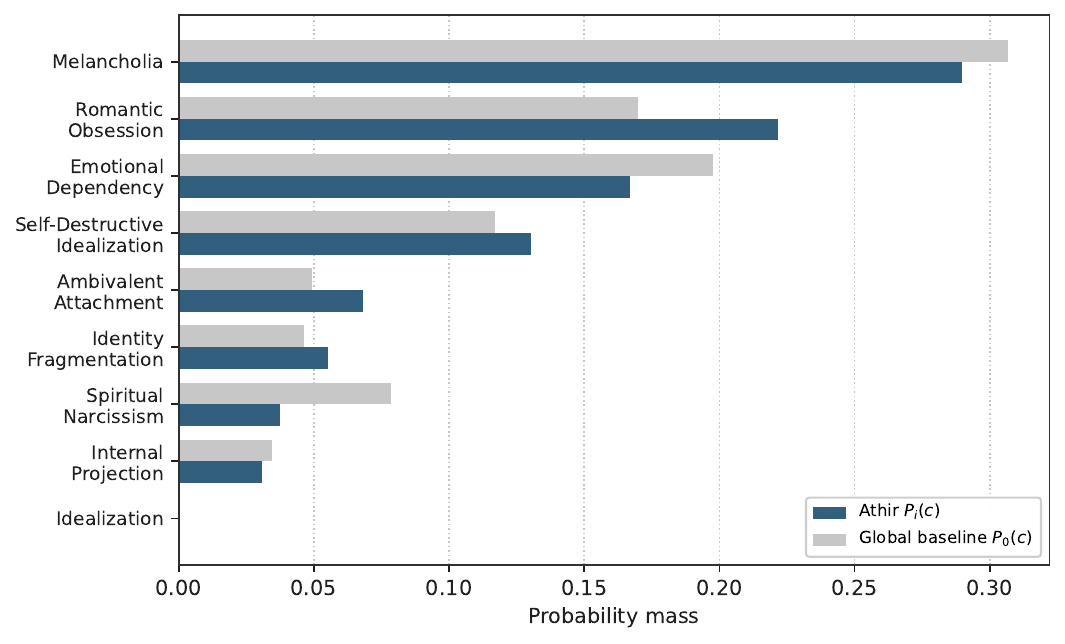}
\caption{Athir: concept distribution $P_i(c)$ against the global baseline $P_0(c)$.}
\label{fig:athir_p1}
\end{figure}

\begin{figure}[t]
\centering
\includegraphics[width=0.92\linewidth]{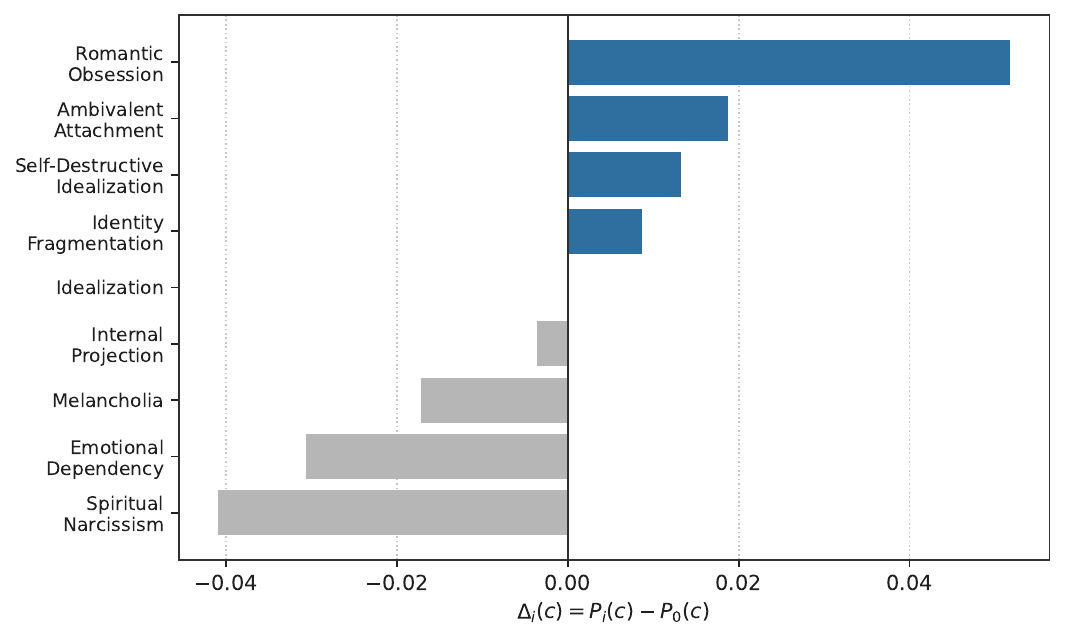}
\caption{Athir: lift profile $\Delta_i(c)=P_i(c)-P_0(c)$ with positive and negative deviations.}
\label{fig:athir_p2}
\end{figure}

\begin{figure}[t]
\centering
\includegraphics[width=0.72\linewidth]{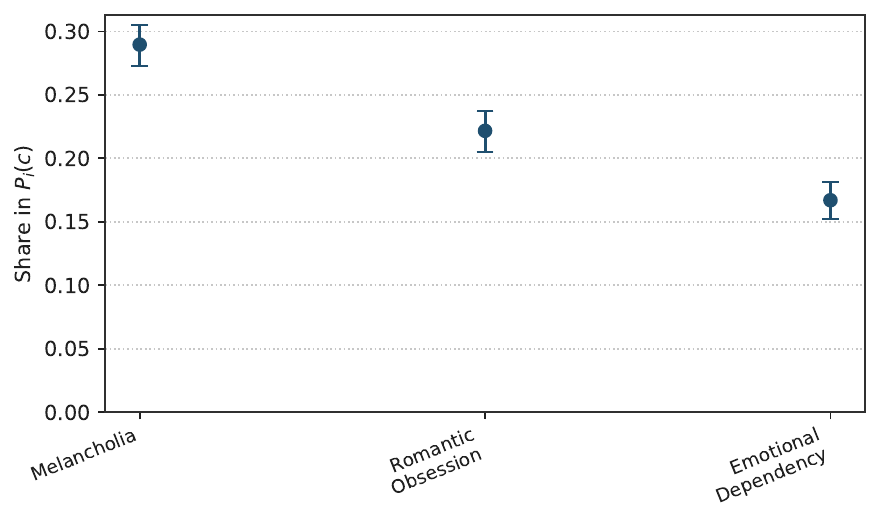}
\caption{Athir: bootstrap 95\% intervals for the top-3 concepts by $P_i(c)$.}
\label{fig:athir_p3}
\end{figure}

\noindent\textbf{Distant-to-Close Exemplars.}
\begin{table}[t]
\centering
\caption{Athir: top verses on the poet's most extreme Eigenmood axis (EM1, negative direction).}
\label{tab:athir_axis_exemplars}
\small
\renewcommand{\arraystretch}{1.1}
\begin{tabularx}{0.98\linewidth}{@{}>{\raggedright\arraybackslash}p{0.21\linewidth} S[table-format=-1.3] >{\raggedright\arraybackslash}X@{}}
\toprule
Reference & {Score} & English translation \\
\midrule
EM1 (negative) & -0.327 & \textit{How can one carry away the heart from my house, when the hand that steals it rises from within?} (Translation ours.) \\
\addlinespace[0.35em]
EM1 (negative) & -0.275 & \textit{I saw that all accounts are zero beside the tally of your pain; I inscribed that tally upon myself.} (Translation ours.) \\
\addlinespace[0.35em]
EM1 (negative) & -0.275 & \textit{Your grief plundered my soul, body, faith, and heart; keep your evil eye away, master thief that you are.} (Translation ours.) \\
\addlinespace[0.35em]
\bottomrule
\end{tabularx}
\end{table}

In this axis-conditioned retrieval, the selected verses track Athir's dominant displacement along EM1: the retrieved lines concentrate lexical and imagistic material that is directionally coherent with the axis loadings, and therefore operationalize the poet's high-level position in the spectral space at verse scale.

\begin{table}[t]
\centering
\caption{Athir: top verses for the leading concept (Melancholia).}
\label{tab:athir_label_exemplars}
\small
\renewcommand{\arraystretch}{1.1}
\begin{tabularx}{0.98\linewidth}{@{}>{\raggedright\arraybackslash}p{0.21\linewidth} S[table-format=1.3] >{\raggedright\arraybackslash}X@{}}
\toprule
Reference & {Score} & English translation \\
\midrule
Melancholia & 0.900 & \textit{Who sits with my heart but grief? Who keeps company with my soul but your pain?} (Translation ours.) \\
\addlinespace[0.35em]
Melancholia & 0.850 & \textit{Little of the night of sorrow remains; we must weep farewell to the day of joy.} (Translation ours.) \\
\addlinespace[0.35em]
Melancholia & 0.850 & \textit{We are left with a life held in grief's hand: much of life is gone, and little patience remains.} (Translation ours.) \\
\addlinespace[0.35em]
\bottomrule
\end{tabularx}
\end{table}

The concept-focused retrieval refines this reading: verses with the highest confidence for Melancholia instantiate recurring rhetorical operations that anchor the aggregate profile. Read together with the axis retrieval, they provide a distant-to-close bridge from distributional signature to inspectable poetic evidence.

\noindent\textbf{Cautions.} These profiles model operational constructs in textual discourse and do not support clinical inference about historical persons. Because abstention filters uncertain cases, selection effects remain analytically relevant; profile interpretation therefore combines confidence-weighted aggregates with explicit awareness of model non-commitment.

\FloatBarrier

\subsection{Poet 10: Khayyam}

\noindent\textbf{Historical and Literary Context.} This result is consistent with scholarship on Omar Khayyam foregrounding impermanence, skepticism, and existential reflection in rubai tradition. \cite{iranica_khayyam,britannica_khayyam}

\noindent\textbf{Poet Summary.} Khayyam contributes 356 verses to the corpus, with abstention rate 0.385 and divergence from baseline $D_{JS}=0.0901$. The uncertainty-aware profile is concentrated in Melancholia, Identity Fragmentation, and Spiritual Narcissism.

\begin{table}[t]
\centering
\caption{Core profile metrics for \textsc{Khayyam}.}
\label{tab:poet_profile_khayyam_metrics}
\footnotesize
\renewcommand{\arraystretch}{1.1}
\begin{tabular}{l S[table-format=5.0] S[table-format=1.3] S[table-format=1.4]}
\toprule
Poet & {Verses} & {Abstain} & {$D_{JS}$} \\
\midrule
Khayyam & 356 & 0.385 & 0.0901 \\
\bottomrule
\end{tabular}
\end{table}

\begin{table}[t]
\centering
\caption{Eigenmood coordinates for \textsc{Khayyam}.}
\label{tab:poet_profile_khayyam_embedding}
\footnotesize
\renewcommand{\arraystretch}{1.1}
\begin{tabular}{l S[table-format=+1.3] S[table-format=+1.3] S[table-format=+1.3]}
\toprule
Poet & {EM1} & {EM2} & {EM3} \\
\midrule
Khayyam & 0.024 & -0.059 & -0.099 \\
\bottomrule
\end{tabular}
\end{table}

\begin{table}[t]
\centering
\caption{Top concept shares for \textsc{Khayyam}. Concept names are abbreviated in the header; full interpretation appears in the text.}
\label{tab:poet_profile_khayyam_concepts}
\small
\renewcommand{\arraystretch}{1.1}
\begin{tabularx}{0.96\linewidth}{@{}l >{\raggedright\arraybackslash}X S[table-format=1.3]@{}}
\toprule
Rank & Concept & {$P_i(c)$} \\
\midrule
1 & Melancholia & 0.538 \\
2 & Identity Fragmentation & 0.147 \\
3 & Spiritual Narcissism & 0.101 \\
\bottomrule
\end{tabularx}
\end{table}

\noindent\textbf{Concept Distribution Interpretation.} We interpret $P_i(c)$ as a confidence-weighted, abstention-aware distribution over operational constructs rather than literal measurements of psychological states. Relative to the global baseline $P_0(c)$, Khayyam is over-represented in Melancholia ($\Delta=+0.232$); Identity Fragmentation ($\Delta=+0.101$), and under-represented in Emotional Dependency ($\Delta=-0.169$); Romantic Obsession ($\Delta=-0.116$). This suggests a poet-specific rhetorical-affective emphasis that is stable at aggregate scale while remaining compatible with multiple literary readings.

\begin{figure}[t]
\centering
\includegraphics[width=0.92\linewidth]{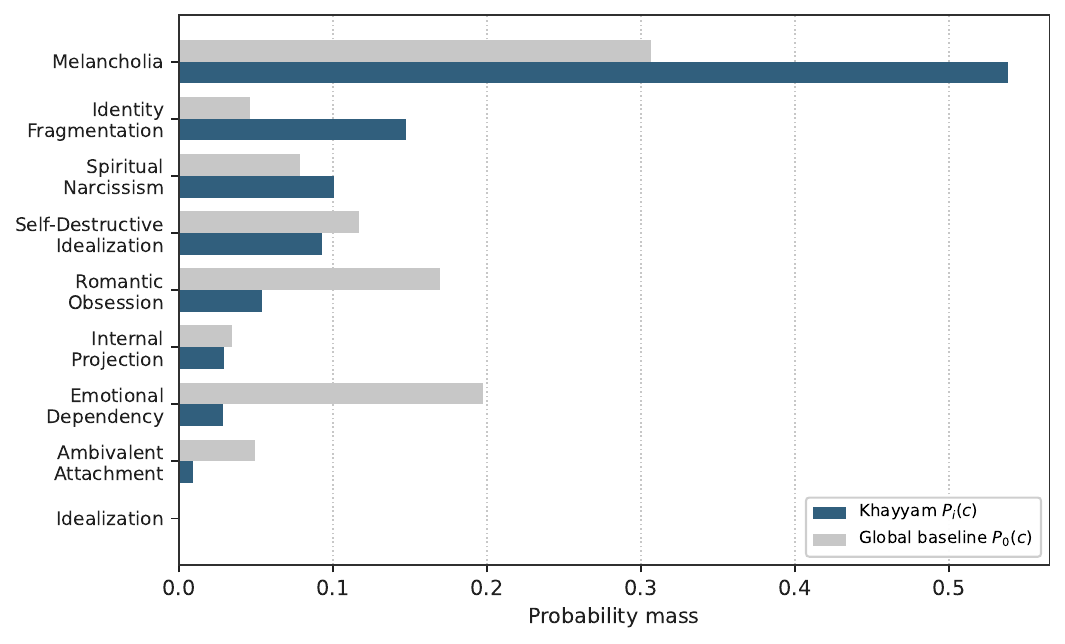}
\caption{Khayyam: concept distribution $P_i(c)$ against the global baseline $P_0(c)$.}
\label{fig:khayyam_p1}
\end{figure}

\begin{figure}[t]
\centering
\includegraphics[width=0.92\linewidth]{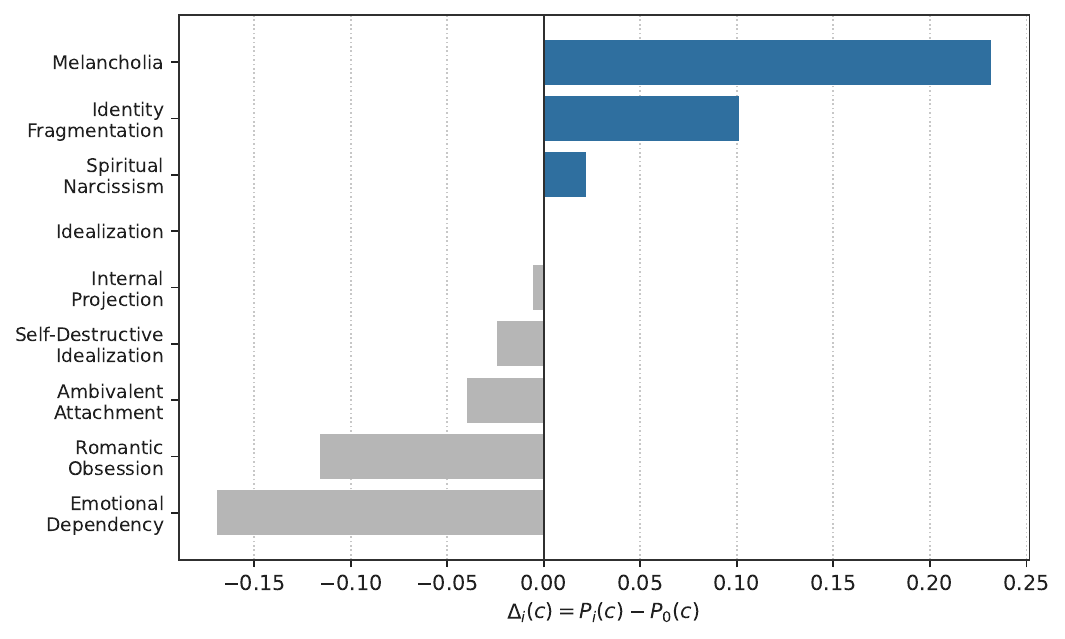}
\caption{Khayyam: lift profile $\Delta_i(c)=P_i(c)-P_0(c)$ with positive and negative deviations.}
\label{fig:khayyam_p2}
\end{figure}

\begin{figure}[t]
\centering
\includegraphics[width=0.72\linewidth]{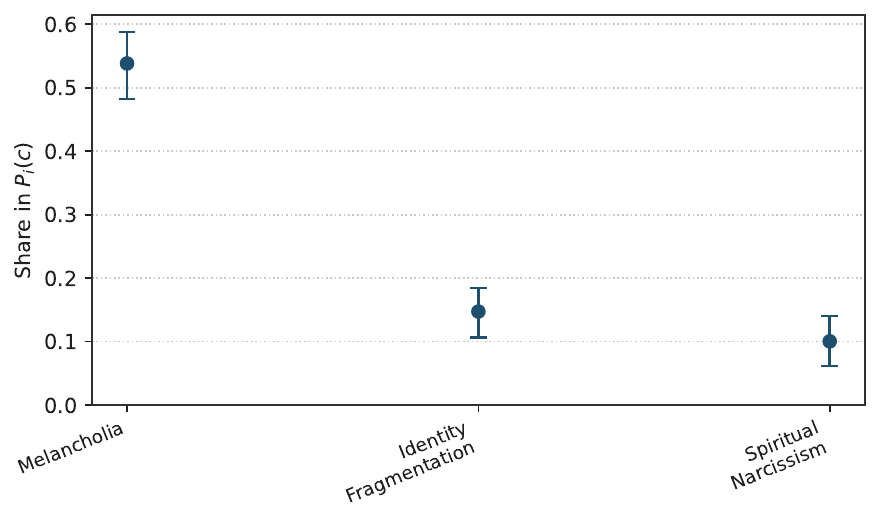}
\caption{Khayyam: bootstrap 95\% intervals for the top-3 concepts by $P_i(c)$.}
\label{fig:khayyam_p3}
\end{figure}

\noindent\textbf{Distant-to-Close Exemplars.}
\begin{table}[t]
\centering
\caption{Khayyam: top verses on the poet's most extreme Eigenmood axis (EM3, negative direction).}
\label{tab:khayyam_axis_exemplars}
\small
\renewcommand{\arraystretch}{1.1}
\begin{tabularx}{0.98\linewidth}{@{}>{\raggedright\arraybackslash}p{0.21\linewidth} S[table-format=-1.3] >{\raggedright\arraybackslash}X@{}}
\toprule
Reference & {Score} & English translation \\
\midrule
EM3 (negative) & -0.724 & \textit{The shell hides that drop in which the sea-heart's secret resides.} (Translation ours.) \\
\addlinespace[0.35em]
EM3 (negative) & -0.637 & \textit{We are the source of joy and also the mine of sorrow.} (Translation ours.) \\
\addlinespace[0.35em]
EM3 (negative) & -0.601 & \textit{We are lowliness and loftiness, fullness and lack; a tarnished mirror and Jamshid's cup.} (Translation ours.) \\
\addlinespace[0.35em]
\bottomrule
\end{tabularx}
\end{table}

In this axis-conditioned retrieval, the selected verses track Khayyam's dominant displacement along EM3: the retrieved lines concentrate lexical and imagistic material that is directionally coherent with the axis loadings, and therefore operationalize the poet's high-level position in the spectral space at verse scale.

\begin{table}[t]
\centering
\caption{Khayyam: top verses for the leading concept (Melancholia).}
\label{tab:khayyam_label_exemplars}
\small
\renewcommand{\arraystretch}{1.1}
\begin{tabularx}{0.98\linewidth}{@{}>{\raggedright\arraybackslash}p{0.21\linewidth} S[table-format=1.3] >{\raggedright\arraybackslash}X@{}}
\toprule
Reference & {Score} & English translation \\
\midrule
Melancholia & 0.850 & \textit{My life has long been dark and my work unstraight; hardship has increased and ease has grown scarce.} (Translation ours.) \\
\addlinespace[0.35em]
Melancholia & 0.850 & \textit{That crystal cup that laughs with wine is a tear concealing the heart's blood.} (Translation ours.) \\
\addlinespace[0.35em]
Melancholia & 0.850 & \textit{Drink wine, for life is one death pursues; better it pass in sleep or in intoxication.} (Translation ours.) \\
\addlinespace[0.35em]
\bottomrule
\end{tabularx}
\end{table}

The concept-focused retrieval refines this reading: verses with the highest confidence for Melancholia instantiate recurring rhetorical operations that anchor the aggregate profile. Read together with the axis retrieval, they provide a distant-to-close bridge from distributional signature to inspectable poetic evidence.

\noindent\textbf{Cautions.} These profiles model operational constructs in textual discourse and do not support clinical inference about historical persons. Because abstention filters uncertain cases, selection effects remain analytically relevant; profile interpretation therefore combines confidence-weighted aggregates with explicit awareness of model non-commitment.

\FloatBarrier

\end{appendices}

\bibliography{sn-bibliography}

\end{document}